\documentclass{article}

\usepackage{rotating}
\usepackage{epic}
\usepackage{eepic}
\usepackage{color}
\usepackage{amsmath}
\usepackage{amssymb}
\usepackage{latexsym}
\usepackage{palatino}
\usepackage{proof}
\usepackage{array}
\usepackage{calc}
\usepackage{mathdots}
\usepackage[english]{babel}
\usepackage[agsm]{harvard}

\newtheorem{theorem}{Theorem}
\newtheorem{definition}[theorem]{Definition}
\newtheorem{example}[theorem]{Example}
\newtheorem{lemma}[theorem]{Lemma}

\newcommand{\qed}{\hfill $\Box$ \vspace{3mm}}

\newlength{\bre}
\newcolumntype{C}{@{}>{\rule{0mm}{\bre}}p{\bre}<{}@{}}

 \newcounter{myovalx}
 \newcounter{myovaly}
 
 \newcounter{myovalxx}
 \newcounter{myovalyy}

 \newcounter{dotfactor}
\newcommand{\ltagz}{\ensuremath{\text{LTAG}_0}}
\newcommand{\ltag}{\mbox{LTAG}}
\newcommand{\nldr}{\textbf{NL}\Diamond_{\calr}}

\newcommand{\editout}[1]{}

\newcommand{\filledcircleindex}[4]{%
         \filledcircle{#1}{#2}{#3}%
         \circleindex{#1}{#2}{\scriptstyle #4}}
\newcommand{\filledcircle}[3]{%
         \put(#1,#2){\circle*{#3}}}%
\newcommand{\circleindex}[3]{%
         \put(#1,#2){\makebox(0,0){\ensuremath{\color{white}\boldsymbol{#3}}}}}

\newcommand{\p}[3]{#2 \circ_{#1} #3}

\newcommand{\zip}[2]{\langle #2 \rangle^{#1}}   
\newcommand{\unzip}[2]{\lfloor #2 \rfloor^{#1}} 
\newcommand{\unpack}[2]{\lceil #2 \rceil^{#1}}

\newcommand{\weg}[1]{}

\newcommand{\one}{\boldsymbol{1}}
\newcommand{\zero}{\boldsymbol{0}}

\newcommand{\bo}{[}
\newcommand{\bc}{]}

\newcommand{\parbottomc}[4]{%
\mbox{%
  \begin{picture}(4,4)
  \put(0,3.4641016){\makebox(0,0){\ensuremath{#3}} }  
  \put(4,3.4641016){\makebox(0,0){\ensuremath{#4}} }  
  \put(2,0){\makebox(0,0){\ensuremath{#2}} }
  \filledcircleindex{2}{2.3094011}{0.8}{#1}
  \drawline(2,0.8)(2,1.9094011)
  \drawline(0.6,3.1176915)(1.6535989,2.5094011)
  \drawline(3.4,3.1176915)(2.34641016,2.5094011)%
  \drawline(2,0.8)(1.9,1)
  \drawline(2,0.8)(2.1,1)
  \end{picture}}}
\newcommand{\parunitb}[2]{%
\mbox{%
  \begin{picture}(4,4)
  \put(2,3.4641016){\makebox(0,0){\ensuremath{#2}} }
  \filledcircleindex{2}{1.7320508}{.8}{#1}
  \drawline(2,2.1320508)(2,2.6641016)
  \drawline(2,2.6641016)(1.9,2.4641016)
  \drawline(2,2.6641016)(2.1,2.4641016)
  \end{picture}}}
\newcommand{\parunita}[2]{%
\mbox{%
  \begin{picture}(4,4)
  \put(2,0){\makebox(0,0){\ensuremath{#2}} }
  \filledcircleindex{2}{1.7320508}{.8}{#1}
  \drawline(2,0.8)(2,1.3320508)
  \drawline(2,0.8)(1.9,1)
  \drawline(2,0.8)(2.1,1)
  \end{picture}}}

\newcommand{\parleftc}[4]{%
\mbox{%
  \begin{picture}(4,4)
  \put(0,3.4641016){\makebox(0,0){\ensuremath{#2}} }  
  \put(4,3.4641016){\makebox(0,0){\ensuremath{#4}} }  
  \put(2,0){\makebox(0,0){\ensuremath{#3}} }
  \filledcircleindex{2}{2.3094011}{0.8}{#1}
  \drawline(2,0.8)(2,1.9094011)
  \drawline(0.6,3.1176915)(1.6535989,2.5094011)
  \drawline(3.4,3.1176915)(2.34641016,2.5094011)
  \drawline(0.6,3.1176915)(0.81754265,3.1099565)
  \drawline(0.6,3.1176915)(0.712320508,2.9370133)
  \end{picture}}}
\newcommand{\parleftcb}[3]{%
\mbox{%
  \begin{picture}(4,4)
  \put(0,3.4641016){\makebox(0,0){\ensuremath{#2}} }  
  \put(2,0){\makebox(0,0){\ensuremath{#3}} }
  \filledcircleindex{2}{2.3094011}{0.8}{#1}
  \drawline(2,0.8)(2,1.9094011)
  \drawline(0.6,3.1176915)(1.6535989,2.5094011)
  \drawline(0.6,3.1176915)(0.81754265,3.1099565)
  \drawline(0.6,3.1176915)(0.712320508,2.9370133)
  \end{picture}}}
\newcommand{\parrightc}[4]{%
\mbox{%
  \begin{picture}(4,4)
  \put(0,3.4641016){\makebox(0,0){\ensuremath{#3}} }  
  \put(4,3.4641016){\makebox(0,0){\ensuremath{#2}} }  
  \put(2,0){\makebox(0,0){\ensuremath{#4}} }
  \filledcircleindex{2}{2.3094011}{0.8}{#1}
  \drawline(2,0.8)(2,1.9094011)
  \drawline(0.6,3.1176915)(1.6535989,2.5094011)
  \drawline(3.4,3.1176915)(2.34641016,2.5094011)
  \drawline(3.4,3.1176915)(3.1824573,3.1099565)
  \drawline(3.4,3.1176915)(3.2876795,2.9370133)
  \end{picture}}}
\newcommand{\parrightcb}[3]{%
\mbox{%
  \begin{picture}(4,4)
  \put(4,3.4641016){\makebox(0,0){\ensuremath{#2}} }  
  \put(2,0){\makebox(0,0){\ensuremath{#3}} }
  \filledcircleindex{2}{2.3094011}{0.8}{#1}
  \drawline(2,0.8)(2,1.9094011)
  \drawline(3.4,3.1176915)(2.34641016,2.5094011)
  \drawline(3.4,3.1176915)(3.1824573,3.1099565)
  \drawline(3.4,3.1176915)(3.2876795,2.9370133)
  \end{picture}}}
\newcommand{\negparleftc}[3]{%
\mbox{%
  \begin{picture}(4,4)
  \put(0,3.4641016){\makebox(0,0){\ensuremath{#2}} }  
  \put(4,3.4641016){\makebox(0,0){\ensuremath{#3}} }  
  \filledcircleindex{2}{2.3094011}{0.8}{#1}
  \drawline(3.4,3.1176915)(2.34641016,2.5094011)
  \drawline(0.6,3.1176915)(1.6535989,2.5094011)
  \drawline(0.6,3.1176915)(0.81754265,3.1099565)
  \drawline(0.6,3.1176915)(0.712320508,2.9370133)
  \end{picture}}}
\newcommand{\negparleftcb}[2]{%
\mbox{%
  \begin{picture}(4,4)
  \put(0,3.4641016){\makebox(0,0){\ensuremath{#2}} }  
  \filledcircleindex{2}{2.3094011}{0.8}{#1}
  \drawline(0.6,3.1176915)(1.6535989,2.5094011)
  \drawline(0.6,3.1176915)(0.81754265,3.1099565)
  \drawline(0.6,3.1176915)(0.712320508,2.9370133)
  \end{picture}}}
\newcommand{\negparrightc}[3]{%
\mbox{%
  \begin{picture}(4,4)
  \put(0,3.4641016){\makebox(0,0){\ensuremath{#3}} }  
  \put(4,3.4641016){\makebox(0,0){\ensuremath{#2}} }  
  \filledcircleindex{2}{2.3094011}{0.8}{#1}
  \drawline(0.6,3.1176915)(1.6535989,2.5094011)
  \drawline(3.4,3.1176915)(2.34641016,2.5094011)
  \drawline(3.4,3.1176915)(3.1824573,3.1099565)
  \drawline(3.4,3.1176915)(3.2876795,2.9370133)
  \end{picture}}}
\newcommand{\negparrightcb}[2]{%
\mbox{%
  \begin{picture}(4,4)
  \put(4,3.4641016){\makebox(0,0){\ensuremath{#2}} }  
  \filledcircleindex{2}{2.3094011}{0.8}{#1}
  \drawline(3.4,3.1176915)(2.34641016,2.5094011)
  \drawline(3.4,3.1176915)(3.1824573,3.1099565)
  \drawline(3.4,3.1176915)(3.2876795,2.9370133)
  \end{picture}}}

\newcommand{\mparbottomc}[4]{%
\mbox{%
  \begin{picture}(4,4)
  \put(0,0){\makebox(0,0){\ensuremath{#3}} }  
  \put(4,0){\makebox(0,0){\ensuremath{#4}} }  
  \put(2,3.4641016){\makebox(0,0){\ensuremath{#2}} }
  \filledcircleindex{2}{1.1547005}{0.8}{#1}
  \drawline(2,2.6641016)(2,1.5547005)
  \drawline(0.6,0.34641012)(1.6535989,0.95470052)
  \drawline(3.4,0.34641012)(2.34641016,0.95470052)%
  \drawline(2,2.6641016)(1.9,2.4641016)
  \drawline(2,2.6641016)(2.1,2.4641016)
  \end{picture}}}
\newcommand{\mparleftc}[4]{%
\mbox{%
  \begin{picture}(4,4)
  \put(0,0){\makebox(0,0){\ensuremath{#2}} }  
  \put(4,0){\makebox(0,0){\ensuremath{#4}} }  
  \put(2,3.4641016){\makebox(0,0){\ensuremath{#3}} }
  \filledcircleindex{2}{1.1547005}{0.8}{#1}
  \drawline(2,2.6641016)(2,1.5547005)
  \drawline(0.6,0.34641012)(1.6535989,0.95470052)
  \drawline(3.4,0.34641012)(2.34641016,0.95470052)
  \drawline(0.6,0.34641012)(0.81754265,0.35414512)
  \drawline(0.6,0.34641012)(0.712320508,0.52708832)
  \end{picture}}}
\newcommand{\mparleftcb}[3]{%
\mbox{%
  \begin{picture}(4,4)
  \put(0,0){\makebox(0,0){\ensuremath{#2}} }  
  \put(2,3.4641016){\makebox(0,0){\ensuremath{#3}} }
  \filledcircleindex{2}{1.1547005}{0.8}{#1}
  \drawline(2,2.6641016)(2,1.5547005)
  \drawline(0.6,0.34641012)(1.6535989,0.95470052)
  \drawline(0.6,0.34641012)(0.81754265,0.35414512)
  \drawline(0.6,0.34641012)(0.712320508,0.52708832)
  \end{picture}}}
\newcommand{\mnegparleftc}[3]{%
\mbox{%
  \begin{picture}(4,4)
  \put(0,0){\makebox(0,0){\ensuremath{#2}} }  
  \put(4,0){\makebox(0,0){\ensuremath{#3}} }  
  \filledcircleindex{2}{1.1547005}{0.8}{#1}
  \drawline(0.6,0.34641012)(1.6535989,0.95470052)
  \drawline(3.4,0.34641012)(2.34641016,0.95470052)
  \drawline(0.6,0.34641012)(0.81754265,0.35414512)
  \drawline(0.6,0.34641012)(0.712320508,0.52708832)
  \end{picture}}}
\newcommand{\mnegparleftcb}[2]{%
\mbox{%
  \begin{picture}(4,4)
  \put(0,0){\makebox(0,0){\ensuremath{#2}} }  
  \filledcircleindex{2}{1.1547005}{0.8}{#1}
  \drawline(0.6,0.34641012)(1.6535989,0.95470052)
  \drawline(0.6,0.34641012)(0.81754265,0.35414512)
  \drawline(0.6,0.34641012)(0.712320508,0.52708832)
  \end{picture}}}
\newcommand{\mparrightc}[4]{%
\mbox{%
  \begin{picture}(4,4)
  \put(0,0){\makebox(0,0){\ensuremath{#3}} }  
  \put(4,0){\makebox(0,0){\ensuremath{#2}} }  
  \put(2,3.4641016){\makebox(0,0){\ensuremath{#4}} }
  \filledcircleindex{2}{1.1547005}{0.8}{#1}
  \drawline(2,2.6641016)(2,1.5547005)
  \drawline(0.6,0.34641012)(1.6535989,0.95470052)
  \drawline(3.4,0.34641012)(2.34641016,0.95470052)
  \drawline(3.4,0.34641012)(3.1824573,0.35414512)
  \drawline(3.4,0.34641012)(3.2876795,0.52708832)
  \end{picture}}}
\newcommand{\mnegparrightc}[3]{%
\mbox{%
  \begin{picture}(4,4)
  \put(0,0){\makebox(0,0){\ensuremath{#3}} }  
  \put(4,0){\makebox(0,0){\ensuremath{#2}} }  
  \filledcircleindex{2}{1.1547005}{0.8}{#1}
  \drawline(0.6,0.34641012)(1.6535989,0.95470052)
  \drawline(3.4,0.34641012)(2.34641016,0.95470052)
  \drawline(3.4,0.34641012)(3.1824573,0.35414512)
  \drawline(3.4,0.34641012)(3.2876795,0.52708832)
  \end{picture}}}
\newcommand{\mnegparrightcb}[2]{%
\mbox{%
  \begin{picture}(4,4)
  \put(4,0){\makebox(0,0){\ensuremath{#2}} }  
  \filledcircleindex{2}{1.1547005}{0.8}{#1}
  \drawline(3.4,0.34641012)(2.34641016,0.95470052)
  \drawline(3.4,0.34641012)(3.1824573,0.35414512)
  \drawline(3.4,0.34641012)(3.2876795,0.52708832)
  \end{picture}}}
\newcommand{\mparrightcb}[3]{%
\mbox{%
  \begin{picture}(4,4)
  \put(4,0){\makebox(0,0){\ensuremath{#2}} }  
  \put(2,3.4641016){\makebox(0,0){\ensuremath{#3}} }
  \filledcircleindex{2}{1.1547005}{0.8}{#1}
  \drawline(2,2.6641016)(2,1.5547005)
  \drawline(3.4,0.34641012)(2.34641016,0.95470052)
  \drawline(3.4,0.34641012)(3.1824573,0.35414512)
  \drawline(3.4,0.34641012)(3.2876795,0.52708832)
  \end{picture}}}
\newcommand{\partopcu}[3]{%
\mbox{%
  \begin{picture}(0,4)
  \put(0,3.4641016){\makebox(0,0){\ensuremath{#3}} }
  \put(0,0){\makebox(0,0){\ensuremath{#2}} }
  \filledcircleindex{0}{1.7320508}{0.8}{#1}
  \drawline(0,0.8)(0,1.3320508)
  \drawline(0,2.1320508)(0,2.6641016)
  \drawline(0,2.6641016)(-0.1,2.4641016)
  \drawline(0,2.6641016)(0.1,2.4641016)
  \end{picture}}}

\newcommand{\parbottomcu}[3]{%
\mbox{%
  \begin{picture}(0,4)
  \put(0,3.4641016){\makebox(0,0){\ensuremath{#3}} }
  \put(0,0){\makebox(0,0){\ensuremath{#2}} }
  \filledcircleindex{0}{1.7320508}{0.8}{#1}
  \drawline(0,0.8)(0,1.3320508)
  \drawline(0,2.1320508)(0,2.6641016)
  \drawline(0,0.8)(-0.1,1)
  \drawline(0,0.8)(0.1,1)
  \end{picture}}}

\newcommand{\ttlparc}[4]{%
\mbox{%
  \begin{picture}(4,4)
  \put(0,0){\makebox(0,0){$ #2 $} }  
  \put(4,0){\makebox(0,0){$ #4 $} }  
  \put(2,0){\makebox(0,0){$ #3 $} }
  \filledcircleindex{2}{1.1547005}{.8}{#1}
  \drawline(2,0.34641012)(2,0.7547005)
  \drawline(0.6,0.34641012)(1.6535989,0.95470052)
  \drawline(3.4,0.34641012)(2.34641016,0.95470052)
  \drawline(0.6,0.34641012)(0.81754265,0.35414512)
  \drawline(0.6,0.34641012)(0.712320508,0.52708832)
  \end{picture}}}
\newcommand{\ttmparc}[4]{%
\mbox{%
  \begin{picture}(4,4)
  \put(0,0){\makebox(0,0){$ #2 $} }  
  \put(4,0){\makebox(0,0){$ #4 $} }  
  \put(2,0){\makebox(0,0){$ #3 $} }
  \filledcircleindex{2}{1.1547005}{.8}{#1}
  \drawline(2,0.34641012)(2,0.7547005)
  \drawline(0.6,0.34641012)(1.6535989,0.95470052)
  \drawline(3.4,0.34641012)(2.34641016,0.95470052)
  \drawline(2,0.34641012)(1.9,0.54641012)
  \drawline(2,0.34641012)(2.1,0.54641012)
  \end{picture}}}
\newcommand{\ttrparc}[4]{%
\mbox{%
  \begin{picture}(4,4)
  \put(0,0){\makebox(0,0){$ #2 $} }  
  \put(4,0){\makebox(0,0){$ #4 $} }  
  \put(2,0){\makebox(0,0){$ #3 $} }
  \filledcircleindex{2}{1.1547005}{.8}{#1}
  \drawline(2,0.34641012)(2,0.7547005)
  \drawline(0.6,0.34641012)(1.6535989,0.95470052)
  \drawline(3.4,0.34641012)(2.34641016,0.95470052)
  \drawline(3.4,0.34641012)(3.1824573,0.35414512)
  \drawline(3.4,0.34641012)(3.2876795,0.52708832)
  \end{picture}}}

\newcommand{\tmlparc}[4]{%
\mbox{%
  \begin{picture}(4,4)
  \put(0,3.4641016){\makebox(0,0){$ #2 $}}  
  \put(4,3.4641016){\makebox(0,0){$ #4 $}}  
  \put(2,3.4641016){\makebox(0,0){$ #3 $}}
  \filledcircleindex{2}{2.30940111}{.8}{#1}
  \drawline(2,3.1176915)(2,2.7094011)
  \drawline(0.6,3.1176915)(1.6535989,2.5094011)
  \drawline(3.4,3.1176915)(2.34641016,2.5094011)
  \drawline(0.6,3.1176915)(0.81754265,3.1099565)
  \drawline(0.6,3.1176915)(0.712320508,2.9370133)
  \end{picture}}}
\newcommand{\tmmparc}[4]{%
\mbox{%
  \begin{picture}(4,4)
  \put(0,3.4641016){\makebox(0,0){$ #2 $}}  
  \put(4,3.4641016){\makebox(0,0){$ #4 $}}  
  \put(2,3.4641016){\makebox(0,0){$ #3 $}}
  \filledcircleindex{2}{2.30940111}{.8}{#1}
  \drawline(2,3.1176915)(2,2.7094011)
  \drawline(0.6,3.1176915)(1.6535989,2.5094011)
  \drawline(3.4,3.1176915)(2.34641016,2.5094011)
  \drawline(2,3.1176915)(1.9,2.9176915)
  \drawline(2,3.1176915)(2.1,2.9176915)
  \end{picture}}}
\newcommand{\tmrparc}[4]{%
\mbox{%
  \begin{picture}(4,4)
  \put(0,3.4641016){\makebox(0,0){$ #2 $}}  
  \put(4,3.4641016){\makebox(0,0){$ #4 $}}  
  \put(2,3.4641016){\makebox(0,0){$ #3 $}}
  \filledcircleindex{2}{2.30940111}{.8}{#1}
  \drawline(2,3.1176915)(2,2.7094011)
  \drawline(0.6,3.1176915)(1.6535989,2.5094011)
  \drawline(3.4,3.1176915)(2.34641016,2.5094011)
  \drawline(3.4,3.1176915)(3.1824573,3.1099565)
  \drawline(3.4,3.1176915)(3.2876795,2.9370133)
  \end{picture}}}

\newcommand{\tensorc}[4]{%
\mbox{%
  \begin{picture}(4,4)
  \put(0,0){\makebox(0,0){$ #3 $} }  
  \put(4,0){\makebox(0,0){$ #4 $} }  
  \put(2,3.4641016){\makebox(0,0){$ #2 $} }
  \put(2,1.1547005){\circle{.8}}
  \put(2,1.1547005){\makebox(0,0){\ensuremath{\scriptstyle #1}}}
  \drawline(2,2.6641016)(2,1.5547005)
  \drawline(0.6,0.34641012)(1.6535989,0.95470052)
  \drawline(3.4,0.34641012)(2.34641016,0.95470052)
  \end{picture}}}
\newcommand{\tensorcl}[4]{%
\mbox{%
  \begin{picture}(4,4)
  \put(0,0){\makebox(0,0){$ #3 $} }  
  \put(2,3.4641016){\makebox(0,0){$ #2 $} }
  \put(2,1.1547005){\circle{.8}}
  \put(2,1.1547005){\makebox(0,0){\ensuremath{\scriptstyle #1}}}
  \drawline(2,2.6641016)(2,1.5547005)
  \drawline(0.6,0.34641012)(1.6535989,0.95470052)
  \end{picture}}}
\newcommand{\tensorcr}[4]{%
\mbox{%
  \begin{picture}(4,4)
  \put(4,0){\makebox(0,0){$ #3 $} }  
  \put(2,3.4641016){\makebox(0,0){$ #2 $} }
  \put(2,1.1547005){\circle{.8}}
  \put(2,1.1547005){\makebox(0,0){\ensuremath{\scriptstyle #1}}}
  \drawline(2,2.6641016)(2,1.5547005)
  \drawline(3.4,0.34641012)(2.34641016,0.95470052)
  \end{picture}}}
\newcommand{\mtensorc}[4]{%
\mbox{%
  \begin{picture}(4,4)
  \put(0,3.4641016){\makebox(0,0){$ #3 $}}  
  \put(4,3.4641016){\makebox(0,0){$ #4 $}}  
  \put(2,0){\makebox(0,0){$ #2 $}}
  \put(2,2.3094011){\circle{.8}}
  \put(2,2.30940111){\makebox(0,0){\ensuremath{\scriptstyle #1}}}
  \drawline(2,0.8)(2,1.9094011)
  \drawline(0.6,3.1176915)(1.6535989,2.5094011)
  \drawline(3.4,3.1176915)(2.34641016,2.5094011)
  \end{picture}}}
\newcommand{\mtensorcright}[3]{%
\mbox{%
  \begin{picture}(4,4)
  \put(4,3.4641016){\makebox(0,0){$ #3 $}}  
  \put(2,0){\makebox(0,0){$ #2 $}}
  \put(2,2.3094011){\circle{.8}}
  \put(2,2.30940111){\makebox(0,0){\ensuremath{\scriptstyle #1}}}
  \drawline(2,0.8)(2,1.9094011)
  \drawline(3.4,3.1176915)(2.34641016,2.5094011)
  \end{picture}}}
\newcommand{\mtensorcleft}[3]{%
\mbox{%
  \begin{picture}(4,4)
  \put(0,3.4641016){\makebox(0,0){$ #3 $}}  
  \put(2,0){\makebox(0,0){$ #2 $}}
  \put(2,2.3094011){\circle{.8}}
  \put(2,2.30940111){\makebox(0,0){\ensuremath{\scriptstyle #1}}}
  \drawline(2,0.8)(2,1.9094011)
  \drawline(0.6,3.1176915)(1.6535989,2.5094011)
  \end{picture}}}
\newcommand{\ttensorc}[4]{%
\mbox{%
  \begin{picture}(4,4)
  \put(0,0){\makebox(0,0){$ #2 $} }  
  \put(4,0){\makebox(0,0){$ #4 $} }  
  \put(2,0){\makebox(0,0){$ #3 $} }
  \put(2,1.1547005){\circle{.8}}
  \put(2,1.1547005){\makebox(0,0){\ensuremath{\scriptstyle #1}}}
  \drawline(2,0.34641012)(2,0.7547005)
  \drawline(0.6,0.34641012)(1.6535989,0.95470052)
  \drawline(3.4,0.34641012)(2.34641016,0.95470052)
  \end{picture}}}
\newcommand{\tmtensorc}[4]{%
\mbox{%
  \begin{picture}(4,4)
  \put(0,3.4641016){\makebox(0,0){$ #2 $}}  
  \put(4,3.4641016){\makebox(0,0){$ #4 $}}  
  \put(2,3.4641016){\makebox(0,0){$ #3 $}}
  \put(2,2.3094011){\circle{.8}}
  \put(2,2.30940111){\makebox(0,0){\ensuremath{\scriptstyle #1}}}
  \drawline(2,3.1176915)(2,2.7094011)
  \drawline(0.6,3.1176915)(1.6535989,2.5094011)
  \drawline(3.4,3.1176915)(2.34641016,2.5094011)
  \end{picture}}}

\newcommand{\negtensora}[3]{%
\mbox{%
  \begin{picture}(4,4)
  \put(0,0){\makebox(0,0){$ #2 $} }  
  \put(4,0){\makebox(0,0){$ #3 $} }  
  \put(2,1.1547005){\circle{.8}}
  \put(2,1.1547005){\makebox(0,0){\ensuremath{\scriptstyle #1}}}
  \drawline(0.6,0.34641012)(1.6535989,0.95470052)
  \drawline(3.4,0.34641012)(2.34641016,0.95470052)
  \end{picture}}}
\newcommand{\negtensorb}[3]{%
\mbox{%
  \begin{picture}(4,4)
  \put(0,3.4641016){\makebox(0,0){$ #2 $}}  
  \put(4,3.4641016){\makebox(0,0){$ #3 $}}
  \put(2,2.3094011){\circle{.8}}
  \put(2,2.30940111){\makebox(0,0){\ensuremath{\scriptstyle #1}}}
  \drawline(0.6,3.1176915)(1.6535989,2.5094011)
  \drawline(3.4,3.1176915)(2.34641016,2.5094011)
  \end{picture}}}
\newcommand{\tensorcu}[3]{%
\mbox{%
  \begin{picture}(4,4)
  \put(2,3.4641016){\makebox(0,0){\ensuremath{#3}} }
  \put(2,0){\makebox(0,0){\ensuremath{#2}} }
  \put(2,1.7320508){\circle{.8}}
  \put(2,1.7320508){\makebox(0,0){\ensuremath{\scriptstyle #1}}}
  \drawline(2,0.8)(2,1.3320508)
  \drawline(2,2.1320508)(2,2.6641016)
  \end{picture}}}
\newcommand{\tensorunita}[2]{%
\mbox{%
  \begin{picture}(4,4)
  \put(2,3.4641016){\makebox(0,0){\ensuremath{#2}} }
  \put(2,1.7320508){\circle{.8}}
  \put(2,1.7320508){\makebox(0,0){\ensuremath{\scriptstyle #1}}}
  \drawline(2,2.1320508)(2,2.6641016)
  \end{picture}}}
\newcommand{\tensorunitb}[2]{%
\mbox{%
  \begin{picture}(4,4)
  \put(2,0){\makebox(0,0){\ensuremath{#2}} }
  \put(2,1.7320508){\circle{.8}}
  \put(2,1.7320508){\makebox(0,0){\ensuremath{\scriptstyle #1}}}
  \drawline(2,0.8)(2,1.3320508)
  \end{picture}}}

\newcommand{\node}[2]{\ensuremath{\overset{#1}{\underset{#2}{\makebox[1ex][c]{\rule{0ex}{1.2ex}\smash{\raisebox{.6ex}{\ensuremath{\centerdot}}}}}}}}
\newcommand{\nnode}[2]{\begin{picture}(0,0)
\put(0,0.12){\makebox(0,0){\ensuremath{\centerdot}}}%
\put(0,0.4){\makebox(0,0)[b]{\ensuremath{\scriptstyle #1}}}%
\put(0,-0.6){\makebox(0,0)[t]{\ensuremath{\scriptstyle #2}}}%
\end{picture}}

\newcommand{\lbbranch}[3]{%
 \begin{picture}(4,3)
  \put(2,3){\makebox(0,0){#1}}
  \put(0,0){\makebox(0,0){#2}}
  \put(4,0){\makebox(0,0){#3}}
  \drawline(0.6,0.6)(1.7,2.4)
  \drawline(3.4,0.6)(2.3,2.4)
 \end{picture}
}
\newcommand{\lubranch}[2]{%
 \begin{picture}(2,3)
  \put(1,3){\makebox(0,0){#1}}
  \put(1,0){\makebox(0,0){#2}}
  \drawline(1,0.6)(1,2.4)
 \end{picture}
}

\newcommand{\terminal}[1]{%
\makebox(0,0){\normalsize\raisebox{-1.1em}{\fbox{\rule{0pt}{0.66em}$\smash{\textup{#1}}_{\rule{0pt}{0.5ex}}$}}}
}

\newcommand{\sk}[1]{%
\vspace{#1\baselineskip}
}

\newcommand{\lolli}{\mbox{$\mathbin{- \!\circ}$}}
\newcommand{\blolli}{\mbox{$\mathbin{\circ\! -}$}}

\newcommand{\pr}{ \mathbin{\smash{\raisebox{1.6ex}{\begin{turn}{-180}{\ensuremath{\&}}\end{turn}}}}}
\newcommand{\prscript}{ \mathbin{\smash{\raisebox{0.8ex}{\begin{turn}{-180}{\ensuremath{\scriptstyle{\&}}}\end{turn}}}}}

\newcommand{\ralabel}[2]{\makebox[2.5em]{$\mathbin{\stackrel{\text{$#1$}}{\rightarrow}_{\text{$#2$}}}$}}

\newcommand{\nodes}[2]{\overset{\text{$#1$}}{\underset{\text{$#2$}}{\node{}{}\;\node{}{}\;\node{}{}}}}

\newcommand{\delete}[1]{}

\newcommand{\ovaalb}{\makebox[0pt]{\begin{picture}(0,0)(-.5,-.5)
\thicklines
\put(0,0){\oval(2.5,1.25)[b]} 
\end{picture}}}

\newcommand{\ovaalt}{\makebox[0pt]{\begin{picture}(0,0)(-.5,-.5)
\thicklines
\put(0,0){\oval(2.5,1.25)[t]} 
\end{picture}}}

\newcommand{\hier}[1]{\makebox[0pt]{\begin{picture}(0,0)(-.5,-.5)
\put(0,0){\makebox(0,0)[b]{\raisebox{-.8ex}[0pt][0pt]{$ #1 $}}}
\end{picture}}
}

\newcommand{\calr}{\mathcal{R}}
\newcommand{\cals}{\mathcal{S}}

\newcommand{\grishina}{\text{Gr1'}}
\newcommand{\grishinb}{\text{Gr1}}
\newcommand{\grishinc}{\text{Gr3}}
\newcommand{\grishind}{\text{Gr3'}}
\newcommand{\grishine}{\text{Gr2'}}
\newcommand{\grishinf}{\text{Gr2}}
\newcommand{\grishing}{\text{Gr4}}
\newcommand{\grishinh}{\text{Gr4'}}
\newcommand{\grishinna}{\text{1'}}
\newcommand{\grishinnb}{\text{1}}
\newcommand{\grishinnc}{\text{3}}
\newcommand{\grishinnd}{\text{3'}}
\newcommand{\grishinne}{\text{2'}}
\newcommand{\grishinnf}{\text{2}}
\newcommand{\grishinng}{\text{4}}
\newcommand{\grishinnh}{\text{4'}}

\begin{document}

\title{Proof Nets for Display Logic}
\author{Richard Moot}
%\date{LaBRI CNRS \& INRIA Futurs \\ Bordeaux University}
\maketitle

%\tableofcontents
\section{Introduction}

\citeasnoun{mp} have introduced proof nets for the multimodal Lambek
calculus $\nldr$. Since then, numerous other connectives have been
proposed to deal with different linguistic phenomena, a par --- or
co-tensor, as some authors prefer to call it --- together with the
corresponding co-implication
\cite{LambekSub,moortgat07sym,bernardi07conti}, Galois and dual-Galois
connectives \cite{galois}.

We can incorporate these extensions (as well as a few others) into the
proof net calculus by simply dropping the restriction that sequents
are trees with a unique root node and obtain what are, in effect,
proof nets for display logic \cite{gore98sub}. The notion of
contraction generalizes to these new connectives without
complications.

 Like for the Lambek calculus, proof nets
for display logic have the advantage of collapsing proofs which differ
only for trivial reasons. The display rules in particular are
compiled away in the proof net representation.

\section{Proof Nets}

%% \begin{itemize}
%%  \item Studia Logica proof nets
%%  \item Definition of link revisited
%%  \item New connectives and their contractions
%%  \item Sequent calculus and links with other systems (display logic/ABM Galois)
%%  \item Special section on 0-ary (if this requires additional restrictions, eg. to prevent disconnected proof structures to become connected)
%% \end{itemize}

Proof nets are an optimal representation for proofs of linear logic
introduced by \citeasnoun{girard}.

\subsection{Links and Proof Structures}

\begin{definition}\label{def:link}
A \emph{link}, as defined by \citeasnoun{mp}\footnote{Some of the
  details are slightly different: the rule name $\nu$ has been
  suppressed since we need only the mode part of it and the
  subsequences $p$ and $q$ have been replaced by the main formula
  argument $m$}, is a tuple $\langle
\tau, P, Q, m \rangle$ where

\begin{itemize}
\item $\tau$, the type of the link, is either $\otimes$ or $\pr$,
\item $P$ is a list of premisses $A_1, \ldots, A_n$,
\item $Q$ is a list of conclusions $B_1, \ldots, B_m$,
\item $m$, the main formula of the link, is either $\epsilon$ or a member of $P \cup Q$.
\end{itemize}

If $m = \epsilon$ then we will call the link \emph{neutral}, if it is
a member of $P$ we will call the link a \emph{left link} and if it is
a member of $Q$ we will call it a \emph{right link}.
\end{definition}

We draw links as shown below, with the premisses from left to right
above the link and the conclusions below it.

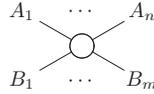
\begin{center}
\scalebox{0.8}{%
\begin{picture}(4,5)(0,1)
\put(0,2){\negtensora{}{B_1}{B_m}}
\put(0,0.8452994){\negtensorb{}{A_1}{A_n}}
\put(2,2){\makebox(0,0){$\cdots$}}
\put(2,4.309401){\makebox(0,0){$\cdots$}}
\end{picture}}
\end{center}

Visually, we distinguish between tensor links --- which we
draw with a white circle at the interior --- and par links --- which
are drawn with a black circle. Finally, unless $m = \epsilon$ we
denote the main formula of the link by drawing an arrow from the
center of the link to this formula. In this case, we will refer to the other formulas as the \emph{active formulas} of the link.

This definition of link allows us to create quite a
number of links in addition to the ones given in that article. The
links there were all possible unary and binary links given the
assumption of a unique conclusion for every tensor link. Once we drop
this constraint, different types of link become possible.

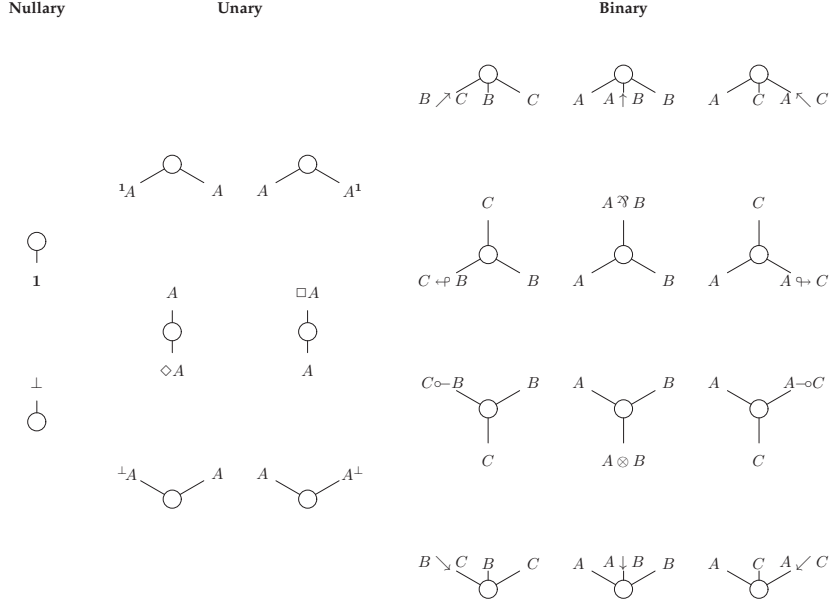
\begin{figure}
 \begin{center}
\scalebox{0.6}{
  \begin{picture}(32,26)
  %\drawline(0,0)(0,23.5)(22,23.5)(22,0)(0,0)
   \put(24,26){\makebox(0,0){\textbf{Binary}}}

   \put(16,-2){\tmtensorc{}{B\searrow C}{B}{C}}
   \put(22,-2){\tmtensorc{}{A}{A\downarrow B}{B}}
   \put(28,-2){\tmtensorc{}{A}{C}{A\swarrow C}}
   \put(22,6){\mtensorc{}{A\otimes B}{A}{B}}
   \put(16,6){\mtensorc{}{C}{C\blolli B}{B}}
   \put(28,6){\mtensorc{}{C}{A}{A\lolli C}}
   \put(22,14){\tensorc{}{A\pr B}{A}{B}}
   \put(16,14){\tensorc{}{C}{C\looparrowleft B}{B}}
   \put(28,14){\tensorc{}{C}{A}{A\looparrowright C}}
   \put(16,22){\ttensorc{}{B\nearrow C}{B}{C}}
   \put(22,22){\ttensorc{}{A}{A\uparrow B}{B}}
   \put(28,22){\ttensorc{}{A}{C}{A\nwarrow C}}

   \put(7,26){\makebox(0,0){\textbf{Unary}}}

   \put(8,10){\tensorcu{}{A}{\Box A}}
   \put(2,10){\tensorcu{}{\Diamond A}{A}}
   \put(2,2){\negtensorb{}{\rule{0pt}{1ex}^{\perp}\! A}{A}}
   \put(8,2){\negtensorb{}{A}{A^{\perp}}}
   \put(2,18){\negtensora{}{\rule{0pt}{1ex}^{\one}\! A}{A}}
   \put(8,18){\negtensora{}{A}{A^{\one}}}

   \put(-2,26){\makebox(0,0){\textbf{Nullary}}}

   \put(-4,6){\tensorunita{}{\perp}}
   \put(-4,14){\tensorunitb{}{\one}}
  \end{picture}}
 \end{center}

\caption{All tensor links of arity 2 or less}
\label{fig:tensors}
\end{figure}

Figure~\ref{fig:tensors} gives an overview of the 9 different forms of
tensor links of arity 2 or less (2 nullary, 3 unary and 4 binary),
together with the logical connectives associated with their different
ports for a total of 2 nullary, 6 unary and 12 binary
connectives.

Note that --- as displayed in the figure --- none of the tensor links
have a main formula according to Definition~\ref{def:link}. However,
in case we need to find the main and active formulas of a link, we can
do so by simply inspecting the formulas assigned to the different ports.

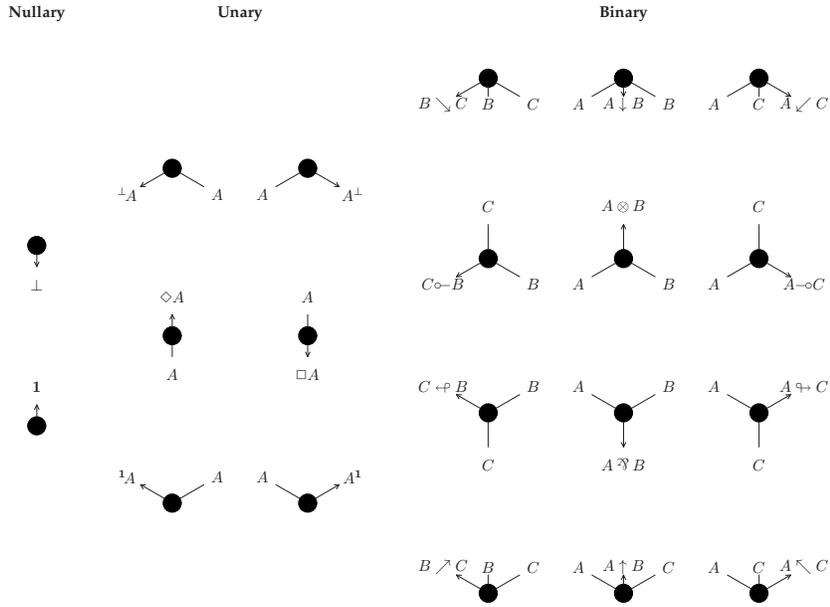
\begin{figure}
 \begin{center}
\scalebox{0.6}{
  \begin{picture}(32,26)
  %\drawline(0,0)(0,23.5)(22,23.5)(22,0)(0,0)
   \put(24,26){\makebox(0,0){\textbf{Binary}}}

   \put(16,-2){\tmlparc{}{B\nearrow C}{B}{C}}
   \put(22,-2){\tmmparc{}{A}{A\uparrow B}{C}}
   \put(28,-2){\tmrparc{}{A}{C}{A\nwarrow C}}
   \put(22,14){\mparbottomc{}{A\otimes B}{A}{B}}
   \put(16,14){\mparleftc{}{C\blolli B}{C}{B}}
   \put(28,14){\mparrightc{}{A\lolli C}{A}{C}}
   \put(22,6){\parbottomc{}{A\pr B}{A}{B}}
   \put(16,6){\parleftc{}{C\looparrowleft B}{C}{B}}
   \put(28,6){\parrightc{}{A\looparrowright C}{A}{C}}
   \put(16,22){\ttlparc{}{B\searrow C}{B}{C}}
   \put(22,22){\ttmparc{}{A}{A\downarrow B}{B}}
   \put(28,22){\ttrparc{}{A}{C}{A\swarrow C}}

   \put(7,26){\makebox(0,0){\textbf{Unary}}}

   \put(10,10){\parbottomcu{}{\Box A}{A}}
   \put(4,10){\partopcu{}{A}{\Diamond A}}
   \put(2,2){\negparleftc{}{\rule{0pt}{1ex}^{\one}\! A}{A}}
   \put(8,2){\negparrightc{}{A^{\one}}{A}}
   \put(2,18){\mnegparleftc{}{\rule{0pt}{1ex}^{\perp}\! A}{A}}
   \put(8,18){\mnegparrightc{}{A^{\perp}}{A}}

   \put(-2,26){\makebox(0,0){\textbf{Nullary}}}

   \put(-4,14){\parunita{}{\perp}}
   \put(-4,6){\parunitb{}{\one}}
  \end{picture}}
 \end{center}

\caption{All par links of arity 2 or less}
\label{fig:pars}
\end{figure}

Corresponding to each tensor link is a par link which is a
`mirror image' of the corresponding tensor link as shown in
Figure~\ref{fig:pars}. 

% For arity $n$ there will generally be $(n+1)(n+2)$
%connectives.

%% \begin{figure}
%%  \begin{center}
%%   \begin{picture}(22,26)
%%   %\drawline(0,0)(0,23.5)(22,23.5)(22,0)(0,0)
%%    \put(18,26){\makebox(0,0){\textbf{Binary}}}

%%    \put(16,-2){\tmtensorc{}{\nnode{B\searrow C}{}}{\nnode{A\downarrow
%%    B}{}}{\nnode{A\swarrow C}{}}}
%%    \put(16,14){\tensorc{}{\nnode{A\pr B}{}}{\nnode{}{C\looparrowleft
%%    B}}{\nnode{}{A\looparrowright C}}}
%%    \put(16,6){\mtensorc{}{\nnode{}{A\otimes B}}{\nnode{C\blolli B}{}}{\nnode{A\lolli C}{}}}
%%    \put(16,22){\ttensorc{}{\nnode{}{B\nearrow C}}{\nnode{}{A\uparrow
%%    B}}{\nnode{}{A\nwarrow C}}}

%%    \put(10,26){\makebox(0,0){\textbf{Unary}}}

%%    \put(8,10){\tensorcu{}{\nnode{}{\Diamond A}}{\nnode{\Box B}{}}}
%%    \put(8,18){\negtensora{}{\nnode{}{A^{\one}}}{\nnode{}{\rule{0pt}{1ex}^{\one}\! A}}}
%%    \put(8,2){\negtensorb{}{\nnode{A^{\perp}}{}}{\nnode{\rule{0pt}{1ex}^{\perp}\! A}{}}}

%%    \put(2,26){\makebox(0,0){\textbf{Nullary}}}

%%    \put(0,6){\tensorunita{}{\nnode{\perp}{}}}
%%    \put(0,14){\tensorunitb{}{\nnode{}{\one}}}
%%   \end{picture}
%%  \end{center}

%% \caption{All neutral tensor links of arity 2 or less}
%% \label{fig:ntensors}
%% \end{figure}

If we want to make more distinctions, we can use modes --- as is usual in
multimodel categorial grammar \cite{M95} --- like we did in \cite{mp}
for $\nldr$ and write the mode in the circle of the link. To somewhat
reduce the (already extensive) vocabulary, we will not talk about
modes in this article, but the current approach can be extended to
incorporate them without problems. Adding them would just amount to
inserting mode information in all tensor and par links and demanding
identity between the two modes to allow a contraction.

\begin{definition} A \emph{proof structure} $\langle S, {\mathcal L}
  \rangle$ is a finite set of formulas $S$ together with a set of
  links ${\mathcal L}$ as shown in Figures~\ref{fig:tensors} and
  \ref{fig:pars} such that.
 \begin{itemize}
  \item every formula of $S$ is at most once the premiss of a link.
  \item every formula of $S$ is at most once the conclusion of a link.
 \end{itemize}

Formulas which are not the conclusion of any link are the
\emph{hypotheses} $H$ of the proof structures, whereas the formulas
which are not the premiss of any link are the it \emph{conclusions}
$C$.
\end{definition}

Readers familiar with proof nets from linear logic will note the
absence or cut and axiom links. We have axiom and cut \emph{formulas}
instead.

\begin{definition} An \emph{axiom formula} is a formula which is not the main
  formula of any link.
A \emph{cut formula} is a formula which is the main formula of two links.
\end{definition}

\begin{figure}
 \begin{center}
\scalebox{0.8}{%
  \begin{picture}(18,13.5)(0,0)
   \put(2,11.4641016){\parunita{}{}}  
   \put(2,8){\mparleftc{}{\perp \blolli A}{\perp}{A}}
   \put(2,2){\mtensorcright{}{}{(\perp \blolli A) \lolli \perp}}
  \qbezier(1.6,7.1176915)(1,6)(3.6535989,4.5094011)
   \put(2,-1.4641016){\tensorunita{}{\perp}}

   \put(12,-1.4641016){\tensorunita{}{}}
   \put(12,11.4641016){\parunita{}{$\nnode{}{}$}}  
   \put(12,8){\mparleftc{}{$\nnode{}{}$}{$\nnode{}{}$}{$\nnode{}{A}$}}
   \put(12,2){\mtensorcright{}{$\nnode{}{}$}{$\nnode{(\perp \blolli A) \lolli \perp}{}$}}
  \qbezier(11.6,7.1176915)(11,6)(13.6535989,4.5094011)
   \put(12,-1.4641016){\tensorunita{}{}}
  \end{picture}}
 \end{center}

 \caption{Proof structure and abstract proof structure}
 \label{grishin1}
\end{figure}
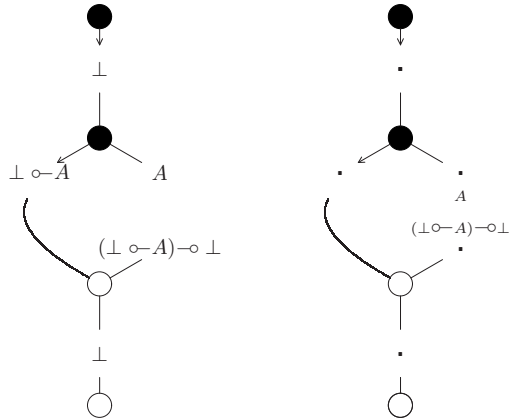

Figure~\ref{grishin1} shows the proof structure for $(\perp \blolli A)
\lolli \perp \vdash A$ on the left. The $A$ formula is the
only axiom in the structure.

%There are four possibilities for axioms, depending on whether the
%formula is a hypothesis of the proof structure, a conclusion of the
%proof structure or both.

There are some differences in the notation of other authors. It is
closest to display logic, with $\pr$ taking the place of $\oplus$ and
${\perp}$ taking the place of $\zero$, much in the spirit of the
connectives from linear logic. The symbols for the two implications
$\looparrowleft$ and $\looparrowright$ have been chosen to remind us
they are the residuals of $\pr$. Table~\ref{tab:logtransl} in Appendix
A gives an
overview of the logical symbols used and the corresponding logical
symbols in various other logics.

\subsection{Abstract Proof Structures}

From a proof structure we obtain an \emph{abstract} proof structure
simply by erasing all formulas on the internal nodes. We only keep the
formulas on the premisses and conlusions of the proof structure,
ie.\ just the leaves. %However, given that a node can be both a premiss
%and a conclusion of a proof structure

\begin{definition} An \emph{abstract proof structure} is a tuple $\langle V,
  {\mathcal L}, p, q \rangle$ such that.

\begin{description}
\item{$V$} is a finite set of vertices,
\item{${\mathcal L}$}  is a set of links such that
 \begin{itemize}
  \item every vertex of $V$ is at most once the premiss of a link,
  \item every vertex of $V$ is at most once the conclusion of a link
 \end{itemize}

 \item{$p$} is a labelling function assigning a formula to the
 hypotheses of the abstract proof structure, that is, to those
 formulas which are not the conclusion of any link,  
\item{$q$} is a labelling function assigning a formula to the
  conclusions of the abstract proof structure, that is, to those
  formulas which are not the premiss of any link.
\end{description}
\end{definition}

We will draw the nodes of abstract proof structures as shown below

\begin{center}
\begin{picture}(0,2)
\put(0,1){\makebox(0,0){\nnode{H}{C}}}
\end{picture}
\end{center}

\noindent where $H$ is the hypothesis assigned to this node and $C$ is
the conclusion assigned to it. Both $H$ and $C$ can be empty.

Figure~\ref{grishin1} shows the abstract proof structure corresponding
to the proof structure of $(\perp \blolli A) \lolli \perp \vdash A$ on
the right.

\begin{definition} A \emph{tensor tree} is an acyclic connected
  abstract proof structure containing only tensor links.
\end{definition}

%\begin{definition} A proof structure $S$ is \emph{correct}, ie.\ it is
%  a \emph{proof net} whenever its underlying abstract proof structure
%  converts to a tensor tree.
%\end{definition}

We say a tensor tree
with hypotheses $A_1, \ldots , A_n$ and conclusions $B_1, \ldots,
B_m$ corresponds to $A_1, \ldots , A_n \vdash B_1, \ldots,
B_m$. However, in order to determine the structure of the sequent to
which a tensor tree corresponds, we first have to do a bit of work.

\subsection{Sequents and Tensor Trees}

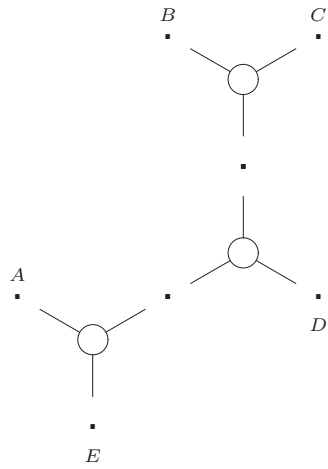
\begin{figure}
 \begin{center}
  \begin{picture}(8,11)
  %\drawline(0,0)(0,10.3)(8,10.3)(8,0)(0,0)
   \put(0,0){\mtensorc{}{\nnode{}{E}}{\nnode{A}{}}{\nnode{}{}}}
   \put(4,3.4641016){\tensorc{}{\nnode{}{}}{\nnode{}{}}{\nnode{}{D}}}
   \put(4,6.9282032){\mtensorc{}{\nnode{}{}}{\nnode{B}{}}{\nnode{C}{}}}
  \end{picture}
 \end{center}

\caption{A tensor tree which is not rooted}
\label{fig:exanew}
\end{figure}

An advantage of the formulation of \citeasnoun{mp} was that, because
of the shape of the two tensor links we considered and because of the
conditions on proof structures, a tensor tree was a rooted tree. The
new types of tensor links do not preserve this
property. Figure~\ref{fig:exanew} shows an example.

Here, we have three premisses ($A$, $B$ and $C$) and two conclusions
($D$ and $E$) but they are grouped in such a way that we cannot turn
them into a sequent $A,B,C \vdash D, E$ straightforwardly.

To solve this problem, we abolish the notion that the premisses of a
sequent are on the left hand side of the turnstile and the conclusions
on the right hand side. We simply split the tensor tree at an
abritrary point and translate the two trees we obtain into
sequents in such a way that we can uniquely recover the original tensor tree.

\begin{figure}
 \begin{center}
\scalebox{0.8}{%
  \begin{picture}(22,26)
  %\drawline(0,0)(0,10.3)(8,10.3)(8,0)(0,0)
   \put(18,26){\makebox(0,0){\textbf{Binary}}}
   \put(16,6){\mtensorc{}{A \circ B}{A}{B}}
   \put(16,10.4641016){\makebox(0,0){$C < B$}}
   \put(20,10.4641016){\makebox(0,0){$A > C$}}
   \put(18,5){\makebox(0,0){$C$}}

   \put(16,14){\tensorc{}{A \circ B}{A}{B}}
   \put(16,13){\makebox(0,0){$C < B$}}
   \put(20,13){\makebox(0,0){$A > C$}}
   \put(18,18.4641016){\makebox(0,0){$C$}}

   \put(16,-2){\tmtensorc{}{\lhd}{\lozenge}{\rhd}}
   \put(16,22){\ttensorc{}{\lhd}{\lozenge}{\rhd}}

   \put(10,26){\makebox(0,0){\textbf{Unary}}}

   \put(8,10){\tensorcu{}{\langle A\rangle}{\langle B\rangle}}
   \put(10,9){\makebox(0,0){$B$}}
   \put(10,14.4641016){\makebox(0,0){$A$}}

   \put(8,18){\negtensora{}{\lfloor A\rfloor}{\lceil B\rceil}}
   \put(8,17){\makebox(0,0){$B$}}
   \put(12,17){\makebox(0,0){$A$}}

   \put(8,2){\negtensorb{}{\lfloor A\rfloor}{\lceil B\rceil}}
   \put(8,6.4641016){\makebox(0,0){$B$}}
   \put(12,6.4641016){\makebox(0,0){$A$}}

   \put(2,26){\makebox(0,0){\textbf{Nullary}}}

   \put(0,6){\tensorunita{}{\epsilon}}
   \put(0,14){\tensorunitb{}{\epsilon}}

  \end{picture}}
 \end{center}

\caption{Information Flow}
\label{fig:flow}
\end{figure}
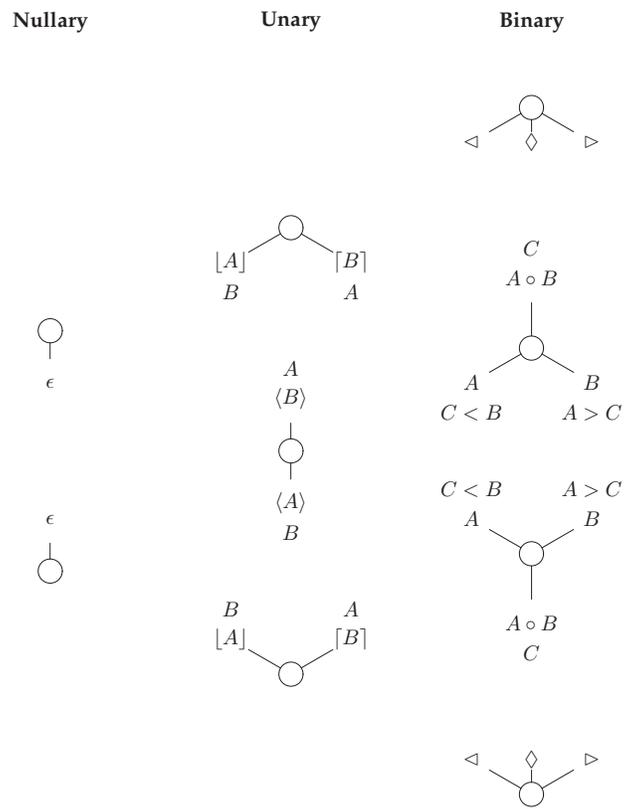

Figure~\ref{fig:flow} lists the structural connectives we need: 1 nullary, 3
unary and 6 binary. The structural connectives are essentially
borrowed from display logic.

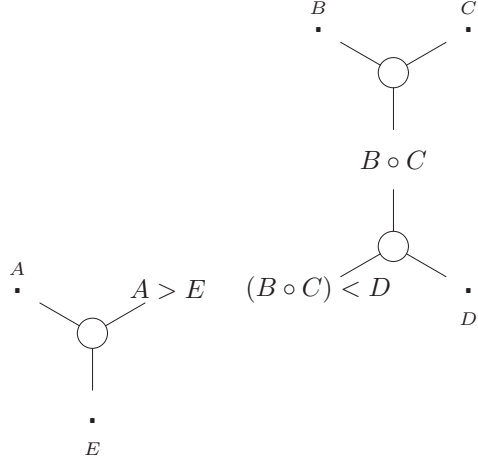
\begin{figure}
 \begin{center}
  \begin{picture}(12,10.3)
  %\drawline(0,0)(0,10.3)(8,10.3)(8,0)(0,0)
   \put(0,0){\mtensorc{}{\nnode{}{E}}{\nnode{A}{}}{A > E}}
   \put(8,3.4641016){\tensorc{}{}{(B \circ C) < D}{\nnode{}{D}}}
   \put(8,6.9282032){\mtensorc{}{B \circ C}{\nnode{B}{}}{\nnode{C}{}}}
  \end{picture}
 \end{center}

\caption{Computing the flow}
\label{fig:exaflow}
\end{figure}

\begin{table}

\begin{center}
\begin{tabular}{cc}
\infer[\bo dgc\bc]{\overline{\unzip{}{\Delta}\vdash \Gamma}}{\unpack{}{\Gamma}\vdash \Delta} &
\infer[\bo gc\bc]{\overline{\Delta\vdash \unzip{}{\Gamma}}}{\Gamma\vdash \unzip{}{\Delta}} \\[2mm]
\multicolumn{2}{c}{\infer[\bo rc\bc]{\overline{\Gamma\vdash \zip{}{\Delta}}}{\zip{}{\Gamma}\vdash \Delta}} \\[2mm]
\infer[\bo rc\bc]{\overline{\Gamma_2 \vdash \Gamma_1>\Delta}}{\infer[\bo rc\bc]{\overline{\p{}{\Gamma_1}{\Gamma_2}\vdash \Delta}}{\Gamma_1 \vdash \Delta<\Gamma_2}} &

\infer[\bo drc\bc]{\overline{\Delta_1>\Gamma \vdash \Delta_2}}{\infer[\bo drc\bc]{\overline{\Gamma\vdash \p{}{\Delta_1}{\Delta_2}}}{\Gamma<\Delta_2\vdash \Delta_1}} \\[2mm]

\infer[\bo dgc\bc]{\overline{\Gamma_1\triangleright \Delta\vdash \Gamma_2}}{\infer[\bo dgc\bc]{\overline{\Gamma_1 \lozenge \Gamma_2 \vdash \Delta}}{\Delta \triangleright \Gamma_2\vdash \Gamma_1}} &
\infer[\bo gc\bc]{\overline{\Delta_2 \vdash \Delta_1\triangleright \Gamma}}{\infer[\bo gc\bc]{\overline{\Gamma \vdash \Delta_1 \lozenge \Delta_2}}{\Delta_1 \vdash \Gamma \triangleleft \Delta_2}} \\[-2mm]
\end{tabular}
\end{center}

\caption{Sequent Rules --- Display Rules}
\label{tab:seqdis}
\end{table}

\begin{definition}\label{def:tttoseq} Let $T$ be a tensor tree and
  $x$ be a node on this  tensor tree, the sequent $T(x)$ is defined as
  follows. We split $T$ at $x$ to obtain a tree $T_h^x$ which has
  $x$ as a hypothesis and a tree $T_c^x$ which has $x$ as a
  conclusion. Without changing the shape of either of the trees, we
  will consider the two instances $x$ as the root of their respective
  trees and all other hypotheses and conclusions as its leaves. Moving
  from these leaves towards $x$ we use the flow of
  Figure~\ref{fig:flow} to compute a term $S_c$ for $T_c^x$ and a term
  $S_h$ for $T_h^x$. The final sequent $T(x)$ is $S_c \vdash S_h$.
\end{definition}

 Note that the tree upwards of the
split point becomes the antecedent, while the tree down from it becomes
the succedent. Figure~\ref{fig:exaflow} shows an example of computing
the flow corresponding a split vertex.
%to the second sequent above.

We see that, depending on our choice of the
`split point' of the tensor tree, Figure~\ref{fig:exanew} corresponds
to one of the following sequents, of which we computed the second in
Figure~\ref{fig:exaflow}.

\sk{.5}
\begin{center}
\begin{tabular}{c}
$ A \vdash E < ((B\circ C) < D)$ \\
$(B \circ C) < D \vdash A > E$ \\
$ A \circ ((B \circ C) < D) \vdash E$ \\
$ B \circ C \vdash (A > E) \circ D$ \\
$ B \vdash ((A > E) \circ D) < C$ \\
$ C \vdash B > ((A > E) \circ D)$ \\
$  (A > E) > (B \circ C) \vdash D$ \\
\end{tabular}
\end{center}
\sk{.5}

 There is exactly one possible
sequent for each vertex in the graph; this is no coincidence as it
corresponds to a `display property' for each vertex. Note that all of
these sequents are interderivable thanks to the display rules of
Table~\ref{tab:seqdis}. In the following, to make it easier to refer
to each of the display rules, we will write the two structural
connectives between parentheses. For example, we will write
$\textit{rc}(<;>)$ for a replacement (from top to bottom) for a $<$
structural connective by a $>$ structural connective.

%\begin{definition} \label{def:tttoseq}
%If $\mathcal T$ is a tensor tree and $x$ a node on this tree, then we
%will denote the sequent $s$ corresponding to the split of the tree on
%$x$ by $\mathcal{T}(x)$ 
%\end{definition}

\begin{lemma} \label{lem:display}
Let $\mathcal T$ be a tensor tree and $x$ and $y$ two nodes on this
tree. Take $s = \mathcal{T}(x)$ and $s' = \mathcal{T}(y)$. $s$ and
$s'$ are equivalent up to the display rules.
\end{lemma}

\paragraph{Proof}
Induction on the length $l$ of the unique path between $x$ and $y$. In case
$l = 0$ then $s$ and $s'$ are identical.

Suppose $l > 0$, induction hypothesis ... we essentially replace one
structural connective by another corresponding to either
$\bo\textit{rc}\bc$, $\bo\textit{drc}\bc$, $\bo\textit{gc}\bc$ or $\bo\textit{rc}\bc$  ... \qed

\begin{lemma} \label{lem:princleaf}
If $\mathcal T$ is a tensor tree containing one or more links, then at
least one of the leaves of the corresponding proof structure
(hypotheses and conclusions) is the main formula of its link.
\end{lemma}

\paragraph{Proof}

Assume $\mathcal T$ has $n$ leaves and that $n-1$ of these leaves are
the active formulas of their link. We show that the last leaf $l$ must be
the main formula of its link. Without changing the orientation of any
of the links, we can see $\mathcal T$ as a tree with root $l$ and and
with $n-1$ leaves. Working our way upward from the deepest level $d$
towards the root we show that the nodes at the next level are always
the active formulas of their link. This means that when we arrive at
the last link of which the root in one of its ports, all its other
ports are the active formulas of the link, which means the root node
must be the main formula. \qed

\subsection{Sequent Rules}

In addition to the display rules, which allow us to turn any formula
in the sequent to either the complete left hand side or the complete
right hand side of a sequent, we have a left and right rule for each
of the connectives.

Given the display property we can always assume --- as shown by the rules in
Tables~\ref{tab:seqnul} to \ref{tab:seqbinb} --- that the context is on the other
side of the sequent as the logical connective we would want to
treat. Apart from the binary galois and dual galois connectives, which
I haven't seen elsewhere, these rules are the same up to notational
choices as those of display logic.

\begin{table}

\begin{center}
\begin{tabular}{cc}
\infer[\bo\textit{L}\one\bc]{\one \vdash \Delta}{\epsilon \vdash \Delta} &
\infer[\bo\textit{R}\one\bc]{\epsilon \vdash \one}{} \\[2mm]
\infer[\bo\textit{L}\perp\bc]{\perp \vdash \epsilon}{} &
\infer[\bo\textit{R}\perp\bc]{\Gamma \vdash \perp}{\Gamma \vdash \epsilon} \\[-2mm]
\end{tabular}
\end{center}

\caption{Sequent Rules --- Nullary Connectives}
\label{tab:seqnul}
\end{table}

\begin{table}

\begin{center}
\begin{tabular}{cc}
\infer[\bo\textit{L}.\rule{0pt}{1ex}^{\perp}\bc]{A^{\perp} \vdash \unpack{}{\Delta}}{\Delta \vdash A} &
\infer[\bo\textit{R}.\rule{0pt}{1ex}^{\perp}\bc]{\Gamma \vdash A^{\perp}}{\Gamma\vdash \unpack{}{A}} \\[2mm] 
\infer[\bo\textit{L}\rule{0pt}{1ex}^{\perp}.\bc]{\rule{0pt}{1ex}^{\perp}\!A \vdash \unzip{}{\Delta}}{\Delta \vdash A} &
\infer[\bo\textit{R}\rule{0pt}{1ex}^{\perp}.\bc]{\Gamma \vdash \rule{0pt}{1ex}^{\perp}\!A}{\Gamma\vdash \unzip{}{A}} \\[2mm] 
\infer[\bo\textit{L}.\rule{0pt}{1ex}^{\one}\bc]{A^{\one} \vdash \Delta}{\unpack{}{A} \vdash \Delta} & 
\infer[\bo\textit{R}.\rule{0pt}{1ex}^{\one}\bc]{\unpack{}{\Gamma} \vdash A^{\one}}{A \vdash \Gamma} \\[2mm]
\infer[\bo\textit{L}\rule{0pt}{1ex}^{\one}.\bc]{\rule{0pt}{1ex}^{\one}\!A\vdash \Delta}{\unzip{}{A}\vdash \Delta} &
\infer[\bo\textit{R}\rule{0pt}{1ex}^{\one}.\bc]{\unzip{}{\Gamma}\vdash \rule{0pt}{1ex}^{\one}\!A}{A\vdash \Gamma} \\[2mm]
\infer[\bo\textit{L}\Diamond\bc]{\Diamond A\vdash \Delta}{\zip{}{A}\vdash \Delta} &
\infer[\bo\textit{R}\Diamond\bc]{\zip{}{\Gamma} \vdash \Diamond A}{\Gamma \vdash A} \\[2mm]
\infer[\bo\textit{L}\Box\bc]{\Box A \vdash \zip{}{\Delta}}{A\vdash \Delta} &
\infer[\bo\textit{R}\Box\bc]{\Gamma \vdash \Box A}{\Gamma \vdash \zip{}{A}} \\[-2mm]
\end{tabular}
\end{center}

\caption{Sequent Rules --- Unary Connectives}
\label{tab:sequna}
\end{table}

\begin{table}

\begin{center}
\begin{tabular}{cc}
\infer[\bo\textit{L}\otimes\bc]{A\otimes B\vdash \Delta}{A \circ B \vdash \Delta} &
\infer[\bo\textit{R}\otimes\bc]{\Gamma_1 \circ \Gamma_2 \vdash A\otimes B}{\Gamma_1 \vdash A & \Gamma_2 \vdash B} \\[2mm]
\infer[\bo\textit{L}\lolli\bc]{A\lolli B\vdash \Gamma > \Delta}{\Gamma \vdash A & B\vdash \Delta} &
\infer[\bo\textit{R}\lolli\bc]{\Gamma \vdash A\lolli B}{\Gamma \vdash A>B} \\[2mm]
\infer[\bo\textit{L}\blolli\bc]{A\blolli B\vdash \Delta< \Gamma}{A\vdash \Delta & \Gamma\vdash B} &
\infer[\bo\textit{R}\blolli\bc]{\Gamma \vdash A\blolli B}{\Gamma \vdash A<B} \\[2mm]
\infer[\bo\textit{L}\pr\bc]{A\pr B \vdash \Delta_1 \circ \Delta_2}{A\vdash \Delta_1 & B\vdash \Delta_2} &
\infer[\bo\textit{R}\pr\bc]{\Gamma \vdash A\pr B}{\Gamma \vdash A \circ B} \\[2mm]
\infer[\bo\textit{L}\looparrowright\bc]{A\looparrowright B\vdash  \Delta}{A>B \vdash \Delta} &
\infer[\bo\textit{R}\looparrowright\bc]{\Delta > \Gamma\vdash A\looparrowright B}{A\vdash \Delta & \Gamma\vdash B} \\[2mm]
\infer[\bo\textit{L}\looparrowleft\bc]{A\looparrowleft B\vdash \Delta}{A<B\vdash \Delta} &
\infer[\bo\textit{R}\looparrowleft\bc]{\Gamma< \Delta\vdash A\looparrowleft B}{\Gamma\vdash A & B\vdash \Delta} \\[-2mm]
\end{tabular}
\end{center}

\caption{Sequent Rules --- Binary Connectives}
\label{tab:seqbin}
\end{table}

\begin{table}

\begin{center}
\begin{tabular}{cc}
\infer[\bo\textit{L}\downarrow\bc]{A\downarrow B\vdash \Gamma_1 \lozenge \Gamma_2}{\Gamma_1\vdash A & \Gamma_2\vdash B} &
\infer[\bo\textit{R}\downarrow\bc]{\Gamma \vdash A\downarrow B}{\Gamma \vdash A\lozenge B} \\[2mm]
\infer[\bo\textit{L}\swarrow\bc]{A\swarrow B\vdash \Gamma_1 \triangleleft \Gamma_2}{\Gamma_1 \vdash A & \Gamma_2 \vdash B} &
\infer[\bo\textit{R}\swarrow\bc]{\Gamma \vdash A\swarrow B}{\Gamma \vdash A\triangleleft B} \\[2mm]
\infer[\bo\textit{L}\searrow\bc]{A\searrow B\vdash \Gamma_1 \triangleright \Gamma_2}{\Gamma_1 \vdash A & \Gamma_2 \vdash B} &
\infer[\bo\textit{R}\searrow\bc]{\Gamma \vdash A\searrow B}{\Gamma \vdash A\triangleright B} \\[2mm]
\infer[\bo\textit{L}\uparrow\bc]{A\uparrow B\vdash \Delta}{A\lozenge B\vdash \Delta} &
\infer[\bo\textit{R}\uparrow\bc]{\Delta_1 \lozenge \Delta_2\vdash A\uparrow B}{A\vdash \Delta_1 & B\vdash \Delta_2} \\[2mm]
\infer[\bo\textit{L}\nwarrow\bc]{A\nwarrow B\vdash \Delta}{A\triangleleft B\vdash \Delta} &
\infer[\bo\textit{R}\nwarrow\bc]{\Delta_1 \triangleleft \Delta_2\vdash A\nwarrow B}{A\vdash \Delta_1 & B\vdash \Delta_2} \\[2mm]
\infer[\bo\textit{L}\nearrow\bc]{A\nearrow B\vdash \Delta}{A\triangleright B\vdash \Delta} &
\infer[\bo\textit{R}\nearrow\bc]{\Delta_1 \triangleright \Delta_2\vdash A\nwarrow B}{A\vdash \Delta_1 & B\vdash \Delta_2} \\[-2mm]
\end{tabular}
\end{center}

\caption{Sequent Rules --- Binary Connectives (continued)}
\label{tab:seqbinb}
\end{table}

%% \subsection{The Nullary Connectives}

%% Nullary contractions $\approx$ axiom link for propositional constants.

%% Possible for abstract proof structure to `split up' when looking at
%% the conversion sequence from right-to-left. But this isn't that
%% different from before, where a par link effectively split a tensor
%% tree into two.

%% Define axiom links on {\em abstract} proof structures and include them
%% with the other conversions.

\subsection{Contractions}

A tensor and a par link contract when the tensor link is connected  ---
respecting up/down and left/right --- to the par link at all its ports
except the single main port of the par link and the corresponding port
of the tensor link.

% Note that the nullary connectives satisfy
%this demand trivially.

The redex for all contraction is a single node as follows.

\begin{center}
\begin{picture}(0,2)
\put(0,1){\makebox(0,0){\nnode{H}{C}}}
\end{picture}
\end{center}

Both links and the internal nodes will be removed from the resulting
graph and the two exterior nodes will be merged, inheriting the
hypothesis and conclusion label of the nodes in case either node is a
hypothesis or conclusion of the abstract proof structure.

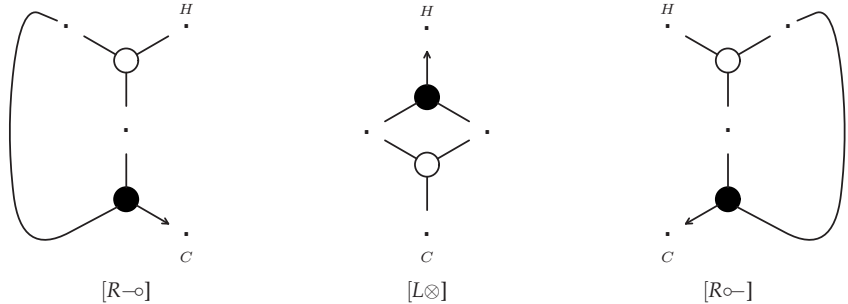
\begin{figure}
\begin{center}
\scalebox{0.8}{%
\begin{picture}(26,10)(0,-1)
\thicklines
   \put(1,0.4){\mparrightcb{}{\nnode{}{C}}{}{}}
   \put(1,3.8641016){\mtensorc{}{\nnode{}{}}{\nnode{}{}}{\nnode{H}{}}}
   \spline(2.7,1.3)(-0.5,-0.4)(-1,3.8)(-0.5,8.0)(0.7,7.5)
%   \spline(0.7,0.1)(-0.5,-0.4)(-1,3.8)(-0.5,8.0)(2.7,6.3)
%   \spline(2.7,2.1)(1,0.4)(0,3.8)(1,7.2)(2.7,5.5)
%   \put(4.5,8){\oval(4,2.5)}
%   \put(4.5,-0.5){\oval(4,2.5)}

   \put(11,0.4){\mtensorc{}{\nnode{}{C}}{}{}}
   \put(11,3.8){\mparbottomc{}{\nnode{H}{}}{\nnode{}{}}{\nnode{}{}}}
%   \put(13,8){\oval(4,2.5)}
%   \put(13,-0.5){\oval(4,2.5)}

   \put(21,0.4){\mparleftcb{}{\nnode{}{C}}{}}
   \put(21,3.8641016){\mtensorc{}{\nnode{}{}}{\nnode{H}{}}{\nnode{}{}}}
   \spline(23.3,1.3)(26.5,-0.4)(27,3.8)(26.5,8.0)(25.3,7.5)
   \put(3,-1.5){\makebox(0,0){\ensuremath{[\textit{R}\lolli}]}}
   \put(13,-1.5){\makebox(0,0){\ensuremath{[\textit{L}\otimes}]}}
   \put(23,-1.5){\makebox(0,0){\ensuremath{[\textit{R}\blolli}]}}
%   \spline(25.3,0.1)(26.5,-0.4)(27,3.8)(26.5,8.0)(23.3,6.3)
%   \spline(23.3,2.1)(25,0.4)(26,3.8)(25,7.2)(23.3,5.5)
%   \put(21,3.8641016){\parleftcb{}{\nnode{H}{}}{}}
%   \put(21,0.4){\tensorc{}{\nnode{}{}}{\nnode{}{C}}{\nnode{}{}}}
%   \spline(25.3,0.1)(26.5,-0.4)(27,3.8)(26.5,8.0)(23.3,6.3)
%   \put(21.5,8){\oval(4,2.5)}
%   \put(21.5,-0.5){\oval(4,2.5)}
%   \put(22,14){\mparbottomc{}{A\otimes B}{A}{B}}
%   \put(16,14){\mparleftc{}{A\blolli B}{A}{B}}
%   \put(28,14){\mparrightc{}{A\lolli B}{A}{B}}
\end{picture}}
\end{center}
\caption{Contractions: binary residuated}
\label{fig:contr:binres}
\end{figure}

Figure~\ref{fig:contr:binres} shows the contractions for the binary
residuated connectives $\lolli$, $\otimes$ and $\blolli$. These are
extactly the contractions proposed for NL.

%% \begin{figure}
%% \scalebox{0.8}{%
%% \begin{picture}(26,12)(0,-2)
%% \thicklines
%%    \put(1,3.0641015){\parrightcb{}{\nnode{H}{}}{}{}}
%%    \put(1,1.2){\mtensorcright{}{}{\nnode{}{C}}}
%%    \spline(2.7,2.1)(1,0.4)(0,3.8)(1,7.2)(2.7,5.5)
%% %   \put(4.5,8){\oval(4,2.5)}
%% %   \put(4.5,-0.5){\oval(4,2.5)}

%%    \put(11,0.7){\mtensorc{}{\nnode{}{C}}{}{}}
%%    \put(11,3.5){\mparbottomc{}{\nnode{H}{}}{}{}}
%% %   \put(13,8){\oval(4,2.5)}
%% %   \put(13,-0.5){\oval(4,2.5)}

%%    \put(21,3.0641016){\parleftcb{}{\nnode{H}{}}{}}
%%    \put(21,){\mtensorcleft{}{}{\nnode{}{C}}}
%%    \spline(23.3,2.1)(25,0.4)(26,3.8)(25,7.2)(23.3,5.5)
%% %   \put(21.5,8){\oval(4,2.5)}
%% %   \put(21.5,-0.5){\oval(4,2.5)}
%% \end{picture}}
%% \caption{Contractions: binary residuated}
%% \end{figure}

\begin{figure}
\begin{center}
\scalebox{0.8}{%
\begin{picture}(26,11)(0,-1)
\thicklines
   \put(1,3.8641015){\parrightcb{}{\nnode{H}{}}{\nnode{}{}}{}}
   \put(1,0.4){\tensorc{}{\nnode{}{}}{\nnode{}{}}{\nnode{}{C}}}
   \spline(0.7,0.1)(-0.5,-0.4)(-1,3.8)(-0.5,8.0)(2.7,6.3)
%   \spline(2.7,1.3)(1,-0.4)(0,3.8)(1,8.0)(2.7,6.3)
%   \put(4.5,8){\oval(4,2.5)}
%   \put(4.5,-0.5){\oval(4,2.5)}

   \put(11,3.8){\tensorc{}{\nnode{H}{}}{\nnode{}{}}{}}
   \put(11,0.4){\parbottomc{}{\nnode{}{C}}{}{\nnode{}{}}}
%   \put(13,8){\oval(4,2.5)}
%   \put(13,-0.5){\oval(4,2.5)}

   \put(21,3.8641016){\parleftcb{}{\nnode{H}{}}{}}
   \put(21,0.4){\tensorc{}{\nnode{}{}}{\nnode{}{C}}{\nnode{}{}}}
   \spline(25.3,0.1)(26.5,-0.4)(27,3.8)(26.5,8.0)(23.3,6.3)
%   \spline(23.3,2.1)(25,0.4)(26,3.8)(25,7.2)(23.3,5.5)
%   \put(21.5,8){\oval(4,2.5)}
%   \put(21.5,-0.5){\oval(4,2.5)}
%   \put(16,6){\parleftc{}{A\looparrowleft B}{A}{B}}
%   \put(28,6){\parrightc{}{A\looparrowright B}{A}{B}}
   \put(3,-1.5){\makebox(0,0){\ensuremath{[\textit{L}\looparrowleft}]}}
   \put(13,-1.5){\makebox(0,0){\ensuremath{[\textit{R}\pr}]}}
   \put(23,-1.5){\makebox(0,0){\ensuremath{[\textit{L}\looparrowright}]}}
\end{picture}}
\end{center}
\caption{Contractions: binary dual residuated}
\label{fig:contr:bindualres}
\end{figure}
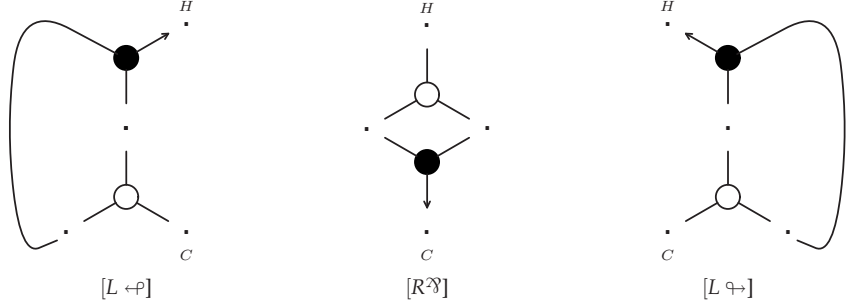

Figure~\ref{fig:contr:bindualres} shows their duals: the contractions
for $\looparrowright$, $\pr$ and $\looparrowleft$. They are obtained
from the residuated conversions by mirroring the figures on the x-axis
and by exchanging hypotheses and conclusions.

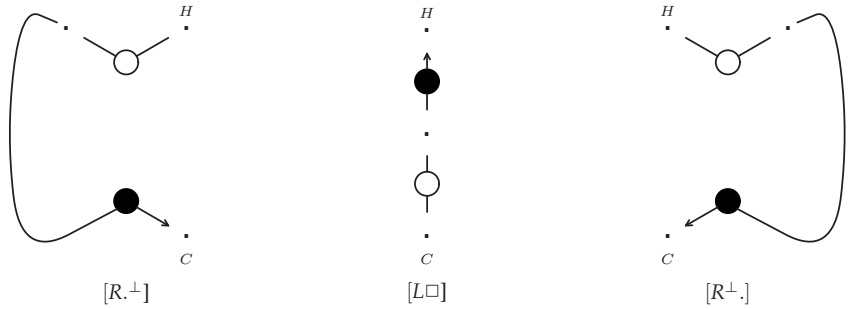
\begin{figure}
\begin{center}
\scalebox{0.8}{%
\begin{picture}(26,10)(0,-1)
\thicklines
   \put(1,0.4){\mnegparrightcb{}{\nnode{}{C}}{}}
   \put(1,3.8641016){\negtensorb{}{\nnode{}{}}{\nnode{H}{}}}
   \spline(2.7,1.3)(-0.5,-0.4)(-1,3.8)(-0.5,8.0)(0.7,7.5)
%   \spline(0.7,0.1)(-0.5,-0.4)(-1,3.8)(-0.5,8.0)(2.7,6.3)
%   \spline(2.7,2.1)(1,0.4)(0,3.8)(1,7.2)(2.7,5.5)
%   \put(4.5,8){\oval(4,2.5)}
%   \put(4.5,-0.5){\oval(4,2.5)}

   \put(11,0.4){\tensorcu{}{\nnode{}{C}}{}}
   \put(13,3.8){\partopcu{}{\nnode{}{}}{\nnode{H}{}}}
%   \put(13,8){\oval(4,2.5)}
%   \put(13,-0.5){\oval(4,2.5)}

   \put(21,0.4){\mnegparleftcb{}{\nnode{}{C}}{}}
   \put(21,3.8641016){\negtensorb{}{\nnode{H}{}}{\nnode{}{}}}
   \spline(23.3,1.3)(26.5,-0.4)(27,3.8)(26.5,8.0)(25.3,7.5)
%   \spline(25.3,0.1)(26.5,-0.4)(27,3.8)(26.5,8.0)(23.3,6.3)
%   \spline(23.3,2.1)(25,0.4)(26,3.8)(25,7.2)(23.3,5.5)
%   \put(21,3.8641016){\parleftcb{}{\nnode{H}{}}{}}
%   \put(21,0.4){\tensorc{}{\nnode{}{}}{\nnode{}{C}}{\nnode{}{}}}
%   \spline(25.3,0.1)(26.5,-0.4)(27,3.8)(26.5,8.0)(23.3,6.3)
%   \put(21.5,8){\oval(4,2.5)}
%   \put(21.5,-0.5){\oval(4,2.5)}
%   \put(22,14){\mparbottomc{}{A\otimes B}{A}{B}}
%   \put(16,14){\mparleftc{}{A\blolli B}{A}{B}}
%   \put(28,14){\mparrightc{}{A\lolli B}{A}{B}}
   \put(3,-1.5){\makebox(0,0){\ensuremath{[\textit{R}.\rule{0pt}{1ex}^{\perp}}]}}
   \put(13,-1.5){\makebox(0,0){\ensuremath{[\textit{L}\Box\rule{0pt}{1ex}}]}}
   \put(23,-1.5){\makebox(0,0){\ensuremath{[\textit{R}\rule{0pt}{1ex}^{\perp}.]}}}
\end{picture}}
\end{center}
\caption{Contractions: unary 1}
\label{fig:contr:un1}
\end{figure}

\begin{figure}
\begin{center}
\scalebox{0.8}{%
\begin{picture}(26,11)(0,-1)
\thicklines
   \put(1,3.8641015){\negparrightcb{}{\nnode{H}{}}}
   \put(1,0.4){\negtensora{}{\nnode{}{}}{\nnode{}{C}}}
   \spline(0.7,0.1)(-0.5,-0.4)(-1,3.8)(-0.5,8.0)(2.7,6.3)
%   \spline(2.7,1.3)(1,-0.4)(0,3.8)(1,8.0)(2.7,6.3)
%   \put(4.5,8){\oval(4,2.5)}
%   \put(4.5,-0.5){\oval(4,2.5)}

   \put(11,3.8){\tensorcu{}{\nnode{}{}}{\nnode{H}{}}}
   \put(13,0.4){\parbottomcu{}{\nnode{}{C}}{}}
%   \put(13,8){\oval(4,2.5)}
%   \put(13,-0.5){\oval(4,2.5)}

   \put(21,3.8641016){\negparleftcb{}{\nnode{H}{}}{}}
   \put(21,0.4){\negtensora{}{\nnode{}{C}}{\nnode{}{}}}
   \spline(25.3,0.1)(26.5,-0.4)(27,3.8)(26.5,8.0)(23.3,6.3)
%   \spline(23.3,2.1)(25,0.4)(26,3.8)(25,7.2)(23.3,5.5)
%   \put(21.5,8){\oval(4,2.5)}
%   \put(21.5,-0.5){\oval(4,2.5)}
%   \put(16,6){\parleftc{}{A\looparrowleft B}{A}{B}}
%   \put(28,6){\parrightc{}{A\looparrowright B}{A}{B}}
   \put(3,-1.5){\makebox(0,0){\ensuremath{[\textit{L}.\rule{0pt}{1ex}^{\one}]}}}
   \put(13,-1.5){\makebox(0,0){\ensuremath{[\textit{R}\Diamond}]}}
   \put(23,-1.5){\makebox(0,0){\ensuremath{[\textit{L}\rule{0pt}{1ex}^{\one}.]}}}
\end{picture}}
\end{center}
\caption{Contractions: unary 2}
\label{fig:contr:un2}
\end{figure}
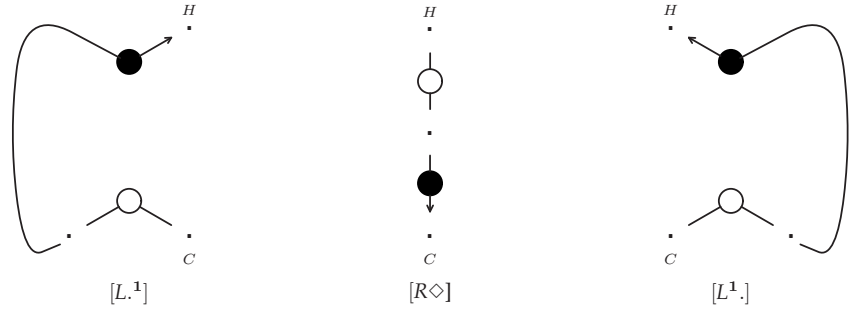

Figures~\ref{fig:contr:un1} and \ref{fig:contr:un2} show the unary
contractions for the Galois, residuated and dual Galois
connectives. They are obtained from Figures~\ref{fig:contr:binres} and
\ref{fig:contr:bindualres} by removing one up-down connection from
each of the redexes.

\begin{figure}
\begin{center}
\scalebox{0.8}{%
\begin{picture}(16,10)(0,-1)
\thicklines
   \put(1,3.8641016){\tensorunita{}{\nnode{H}{}}}
   \put(1,0.4){\parunita{}{\nnode{}{C}}}

   \put(11,0.4){\tensorunitb{}{\nnode{}{C}}}
   \put(11,3.8641016){\parunitb{}{\nnode{H}{}}}

   \put(3,-1.5){\makebox(0,0){\ensuremath{[\textit{R}\perp]}}}
   \put(13,-1.5){\makebox(0,0){\ensuremath{[\textit{L}\one]}}}
\end{picture}}
\end{center}

\caption{Contractions: nullary}
\label{fig:contr:nul}
\end{figure}
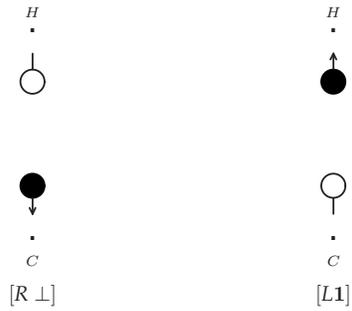

Figure~\ref{fig:contr:nul} shows the nullary contractions. Since the
nullary links have just one port, the condition that all ports except
the main port have to be connected is statisfied trivially, thus the
nullary contractions simply identify different nodes in the graph.

\subsection{Structural Rules}

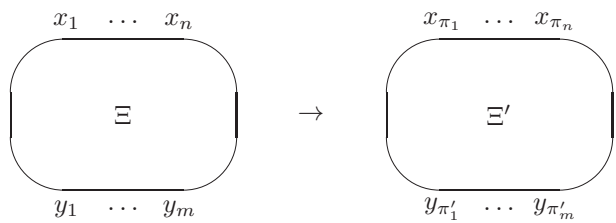
\begin{figure}
\begin{center}
\begin{picture}(18,6)
 \put(2.5,5.5){\makebox(0,0){$x_1$}}
 \put(4,5.5){\makebox(0,0){$\ldots$}}
 \put(5.5,5.5){\makebox(0,0){$x_n$}}
 \put(2.5,0.5){\makebox(0,0){$y_1$}}
 \put(4,0.5){\makebox(0,0){$\ldots$}}
 \put(5.5,0.5){\makebox(0,0){$y_m$}}
 \put(4,3){\oval(6,4)}
 \put(14,3){\oval(6,4)}
 \put(9,3){\makebox(0,0){$\rightarrow$}}
 \put(4,3){\makebox(0,0){$\Xi$}}
 \put(14,3){\makebox(0,0){$\Xi'$}}
 \put(12.5,5.5){\makebox(0,0){$x_{\pi_1}$}}
 \put(14,5.5){\makebox(0,0){$\ldots$}}
 \put(15.5,5.5){\makebox(0,0){$x_{\pi_n}$}}
 \put(12.5,0.5){\makebox(0,0){$y_{\pi'_1}$}}
 \put(14,0.5){\makebox(0,0){$\ldots$}}
 \put(15.5,0.5){\makebox(0,0){$y_{\pi'_m}$}}
\end{picture}
\end{center}
\caption{Schematic Form of a Structural Conversion}
\label{fig:strconv}
\end{figure}

We can extend the proof net as well as the sequent calculus with an
arbitrary number of structural conversions. A structural conversion in
the proof net calculus is simply a rewrite of one tensor tree into
another in such a way that both trees have the same hypotheses and
the same conclusions, though we are allowed to change their
order. Figure~\ref{fig:strconv} shows the schematic form of a
structural conversion. The $x$ vertices are the $n$ hypotheses of the
conversion, the $y$ vertices the $m$ conclusions and $\pi$ (resp.\ $\pi'$)
is a permutation of the hypotheses (resp.\ the conclusions).

This restriction means the contraction and weakening rules are not
allowed: we operate essentially in a fragment of multiplicative linear
logic \cite{girard}.

% Every structural conversion of this form will correspond to $n+m$
% structural rules in the sequent calculus.

\begin{figure}
 \begin{center}
\scalebox{0.8}{%
  \begin{picture}(18,9)(0,-2)
   \put(0,2){\tensorc{}{$\nnode{X}{}$}{$\nnode{}{Y}$}{}}
   \put(2,-1.4641016){\tensorunita{}{$\nnode{}{}$}}
   \put(14,2){\tensorc{}{$\nnode{X}{}$}{}{$\nnode{}{Y}$}}
   \put(12,-1.4641016){\tensorunita{}{$\nnode{}{}$}}
   \put(9,2){\makebox(0,0){$\nnode{X}{Y}$}}
   \put(6.5,2){\makebox(0,0){$\leftrightsquigarrow_{\text{Idr}\otimes\epsilon}$}}
   \put(11.5,2){\makebox(0,0){$\leftrightsquigarrow_{\text{Idl}\otimes\epsilon}$}}
  \end{picture}}

\scalebox{0.8}{%
  \begin{picture}(18,6)(0,-2)
   \put(0,-1.4641016){\mtensorc{}{$\nnode{}{Y}$}{$\nnode{X}{}$}{}}
   \put(2,2){\tensorunitb{}{$\nnode{}{}$}}
   \put(14,-1.4641016){\mtensorc{}{$\nnode{}{Y}$}{}{$\nnode{X}{}$}}
   \put(12,2){\tensorunitb{}{$\nnode{}{}$}}
   \put(9,2){\makebox(0,0){$\nnode{X}{Y}$}}
   \put(6.5,2){\makebox(0,0){$\leftrightsquigarrow_{\text{Idr}\prscript\epsilon}$}}
   \put(11.5,2){\makebox(0,0){$\leftrightsquigarrow_{\text{Idl}\prscript\epsilon}$}}
  \end{picture}}
 \end{center}
 \caption{Identity rules for tensor and par}
 \label{fig:identity}
\end{figure}
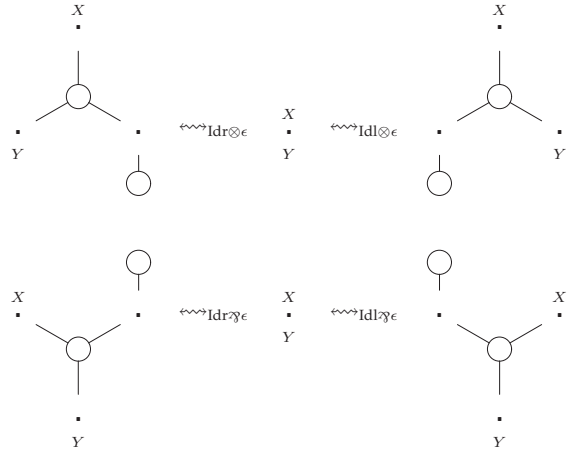

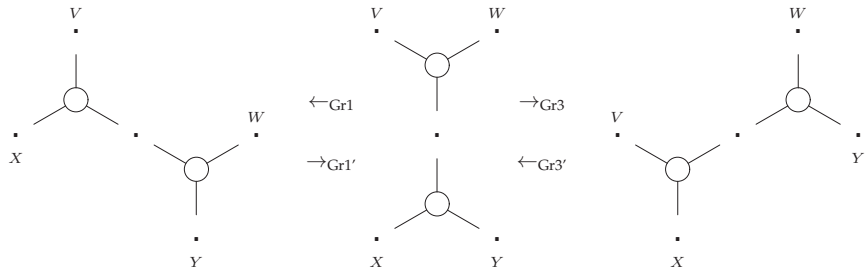
\begin{figure}
 \begin{center}
\scalebox{0.8}{%
  \begin{picture}(18,10.3)(4,3)
  %\drawline(0,0)(0,10.3)(8,10.3)(8,0)(0,0)
%   \put(0,0){\mtensorc{}{\nnode{}{E}}{\nnode{A}{}}{>}}
   \put(0,6.9282032){\tensorc{}{$\nnode{V}{}$}{$\nnode{}{X}$}{\nnode{}{}}}
   \put(4,3.4641016){\mtensorc{}{$\nnode{}{Y}$}{}{$\nnode{W}{}$}}
   \put(10.5,5.92){\makebox(0,0){$\rightarrow_{\grishina}$}}
   \put(10.5,7.92){\makebox(0,0){$\leftarrow_{\grishinb}$}}
   \put(12,3.4641016){\tensorc{}{}{$\nnode{}{X}$}{$\nnode{}{Y}$}}
   \put(12,6.9282032){\mtensorc{}{\nnode{}{}}{$\nnode{V}{}$}{$\nnode{W}{}$}}
   \put(17.5,7.92){\makebox(0,0){$\rightarrow_{\grishinc}$}}
   \put(17.5,5.92){\makebox(0,0){$\leftarrow_{\grishind}$}}
   \put(24,6.9282032){\tensorc{}{$\nnode{W}{}$}{\nnode{}{}}{$\nnode{}{Y}$}}
   \put(20,3.4641016){\mtensorc{}{$\nnode{}{X}$}{$\nnode{V}{}$}{}}
  \end{picture}}
%% \scalebox{0.8}{%
%%   \begin{picture}(18,10.3)(0,3)
%%   %\drawline(0,0)(0,10.3)(8,10.3)(8,0)(0,0)
%% %   \put(0,0){\mtensorc{}{\nnode{}{E}}{\nnode{A}{}}{>}}
%%    \put(4,6.9282032){\tensorc{}{$\nnode{W}{}$}{\nnode{}{}}{$\nnode{}{Y}$}}
%%    \put(0,3.4641016){\mtensorc{}{$\nnode{}{X}$}{$\nnode{V}{}$}{}}
%%    \put(11,7.92){\makebox(0,0){$\rightarrow_{\text{Grn(c)}}$}}
%%    \put(11,5.92){\makebox(0,0){$\leftarrow_{\text{Grn(d)}}$}}
%% %   \put(11,6.92){\makebox(0,0){$\leftrightsquigarrow_{\text{Grn(ii)}}$}}
%%    \put(14,3.4641016){\tensorc{}{}{$\nnode{}{X}$}{$\nnode{}{Y}$}}
%%    \put(14,6.9282032){\mtensorc{}{\nnode{}{}}{$\nnode{V}{}$}{$\nnode{W}{}$}}
%%   \end{picture}}
 \end{center}

\caption{Grishin Rules: Mixed Associativity}
\label{fig:grishin}
\end{figure}

Figures~\ref{fig:identity} and \ref{fig:grishin} shows some well-known
examples of valid structural rules which don't change the order of the
hypotheses and premisses.

\begin{figure}
 \begin{center}
\scalebox{0.8}{%
  \begin{picture}(18,10.3)(4,3)
   \put(1,9){\tensorcl{}{$\nnode{W}{}$}{$\nnode{}{X}$}{}}
   \put(1,3){\mtensorcleft{}{$\nnode{}{Y}$}{$\nnode{V}{}$}{}}
%  \drawline(3.4,0.34641012)(2.34641016,0.95470052)
%   \drawline(4.4,3.34641012)(3.34641016,3.95470052)
   \drawline(3.4,5.45470052)(6,7)
   \drawline(3.4,9.95470052)(6,8.4)
   \put(6,7.7){\makebox(0,0){\nnode{}{}}}
   \put(9.5,5.92){\makebox(0,0){$\rightarrow_{\grishine}$}}
   \put(9.5,7.92){\makebox(0,0){$\leftarrow_{\grishinf}$}}
   \put(12,3.4641016){\tensorc{}{}{$\nnode{}{X}$}{$\nnode{}{Y}$}}
   \put(12,6.9282032){\mtensorc{}{\nnode{}{}}{$\nnode{V}{}$}{$\nnode{W}{}$}}
   \put(18.5,7.92){\makebox(0,0){$\rightarrow_{\grishing}$}}
   \put(18.5,5.92){\makebox(0,0){$\leftarrow_{\grishinh}$}}
   \put(23,9){\tensorcr{}{$\nnode{V}{}$}{$\nnode{}{Y}$}}
   \put(23,3){\mtensorcright{}{$\nnode{}{X}$}{$\nnode{W}{}$}{}}
   \drawline(24.7,5.45470052)(22.1,7)
   \drawline(24.7,9.95470052)(22.1,8.4)
   \put(22.1,7.7){\makebox(0,0){\nnode{}{}}}
  \end{picture}}
 \end{center}

\caption{Grishin Rules: Mixed Commutativity}
\label{fig:grishinmc}
\end{figure}
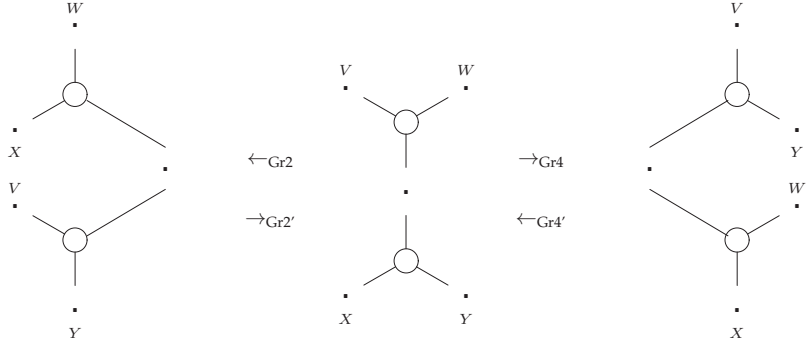

The structural rules of Figure~\ref{fig:grishin} play a role similar
to the mixed associativity rules in multimodal system. The
corresponding mixed commutativity rules are shown in
Figure~\ref{fig:grishinmc}. We will refer to the primed versions of
the structural conversions, ie.\ those moving \emph{towards} the center
structure, as the Grishin class I rules, whereas the non-primed rules,
those moving \emph{away from} the center structure are the Grishin class IV
rules \cite{grishin}.

Given a structural conversion, what is the sequent rule which
corresponds to it? As shown by the following lemma, there are multiple
equivalent possibilities, depending on which of the leaves is displayed.

\begin{definition}
Let $s$ be a structural conversion and $l$ one of its leaves. $s(l)$
will denote the structural rule obtained by computing the flow
according to Definition~\ref{def:tttoseq} with the exception that
every hypothesis leaf $x_i$ will correspond to a structural variable
$\Gamma_i$ and every conclusion leaf $y_i$ to a structural variable
$\Delta_i$.
\end{definition}

For example, depending on whether we use hypothesis $X$ or conclusion
$Y$ to obtain a corresponding structural rule, we obtain either rule
$\bo\textit{Idr}\otimes\epsilon1\bc$ or rule
$\bo\textit{Idr}\otimes\epsilon2\bc$. 

\begin{center}
\begin{tabular}{c@{\qquad\qquad}c}
\mbox{\infer[\bo\textit{Idr}\otimes\epsilon1\bc]{\Gamma \vdash
    \Delta}{\Gamma \circ \epsilon \vdash \Delta}}
&
\mbox{\infer[\bo\textit{Idr}\otimes\epsilon2\bc]{\Gamma \vdash
    \Delta}{\Gamma \vdash \Delta < \epsilon}}
\end{tabular}
\end{center}

Note that the two rules are equivalent.

\begin{center}
\begin{tabular}{c@{\qquad\qquad}c}
\mbox{\infer[\bo\textit{Idr}\otimes\epsilon1\bc]{\Gamma \vdash
    \Delta}{\infer[\bo\textit{rc}\circ<\bc]{\Gamma \circ \epsilon
      \vdash \Delta}{\Gamma \vdash \Delta < \epsilon}}}
&
\mbox{\infer[\bo\textit{Idr}\otimes\epsilon2\bc]{\Gamma \vdash
    \Delta}{\infer[\bo\textit{rc}<\circ\bc]{\Gamma \vdash \Delta <
      \epsilon}{\Gamma \circ \epsilon \vdash \Delta}}}
\end{tabular}
\end{center}

\begin{lemma}
Let $l_1$ and $l_2$ be two distinct leaves of a structural conversion
$s$. Then the structural rules $s(l_1)$ and $s(l_2)$ are
interderivable using only the other rule and the display rules.
\end{lemma}

\paragraph{Proof} Similar to the proof of Lemma~\ref{lem:display} we
follow the unique path from $l_1$ to $l_2$ applying a display rule at
each step to derive $s(l_2)$ from $s(l_1)$. Given that all the display
rules are reversible we can derive $s(l_1)$ from $s(l_2)$ using the
inverse rules. \qed

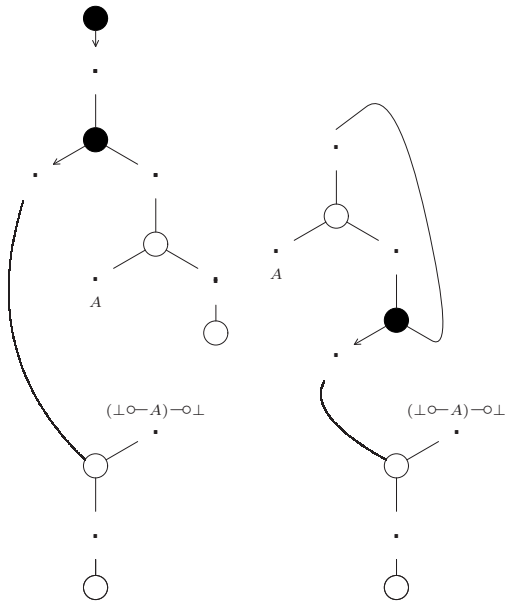
\begin{figure}
 \begin{center}
\scalebox{0.8}{%
  \begin{picture}(18,20)(0,0)
   \put(2,-1.4641016){\tensorunita{}{}}
   \put(2,17.4641016){\parunita{}{$\nnode{}{}$}}  
   \put(2,14){\mparleftc{}{$\nnode{}{}$}{$\nnode{}{}$}{$\nnode{}{}$}}
   \put(2,2){\mtensorcright{}{$\nnode{}{}$}{$\nnode{(\perp \blolli A) \lolli \perp}{}$}}
  \qbezier(1.6,13.1176915)(0,8)(3.6535989,4.5094011)
   \put(2,-1.4641016){\tensorunita{}{}}
   \put(4,10.5358984){\tensorc{}{$\nnode{}{}$}{$\nnode{}{A}$}{$\nnode{}{}$}}
   \put(6,7){\tensorunita{}{\nnode{}{}}}

   \put(12,-1.4641016){\tensorunita{}{}}
   \put(12,8){\mparleftcb{}{$\nnode{}{}$}{$\nnode{}{}$}}
   \spline(14.34641016,8.95470052)(16,8)(14,17)(12,15.5)
   \put(12,2){\mtensorcright{}{$\nnode{}{}$}{$\nnode{(\perp \blolli A) \lolli \perp}{}$}}
  \qbezier(11.6,7.1176915)(11,6)(13.6535989,4.5094011)
   \put(12,-1.4641016){\tensorunita{}{}}
   \put(10,11.4641016){\tensorc{}{$\nnode{}{}$}{$\nnode{}{A}$}{}}
  \end{picture}}
 \end{center}

 \caption{Right identity for $\otimes$, followed by the $\perp$ contraction}
 \label{grishin2}
\end{figure}

\begin{figure}
 \begin{center}
\scalebox{0.8}{%
  \begin{picture}(18,20)(0,0)
   \put(2,-1.4641016){\tensorunita{}{}}
   \put(2,8){\mparleftcb{}{$\nnode{}{}$}{$\nnode{}{}$}}
   \spline(4.34641016,8.95470052)(6,8)(6,20.9641016)(4,18.9641016)
   \put(2,2){\mtensorcright{}{$\nnode{}{}$}{$\nnode{(\perp \blolli A) \lolli \perp}{}$}}
  \qbezier(1.6,7.1176915)(1,6)(3.6535989,4.5094011)
   \put(2,-1.4641016){\tensorunita{}{}}
   \put(0,11.4641016){\tensorc{}{}{$\nnode{}{A}$}{}}
   \put(0,14.9282032){\mtensorc{}{$\nnode{}{}$}{$\nnode{}{}$}{$\nnode{}{}$}}
   \put(-2,18.3923048){\tensorunitb{}{}}

   \put(12,-1.4641016){\tensorunita{}{}}
   \put(12,8){\mparleftcb{}{$\nnode{}{}$}{$\nnode{}{}$}}
   \spline(14.34641016,8.95470052)(16,8)(18,17.5)(16,15.5)
   \put(12,2){\mtensorcright{}{$\nnode{}{}$}{$\nnode{(\perp \blolli A) \lolli \perp}{}$}}
  \qbezier(11.6,7.1176915)(11,6)(13.6535989,4.5094011)
   \put(12,-1.4641016){\tensorunita{}{}}
   \put(8,14.9282032){\tensorc{}{}{$\nnode{}{A}$}{}}
   \put(12,11.4641016){\mtensorc{}{$\nnode{}{}$}{$\nnode{}{}$}{$\nnode{}{}$}}
   \put(8,18.3923048){\tensorunitb{}{$\nnode{}{}$}}
  \end{picture}}
 \end{center}
  \caption{Left identity for par followed by $\grishinb$}
  \label{grishin3}
\end{figure}
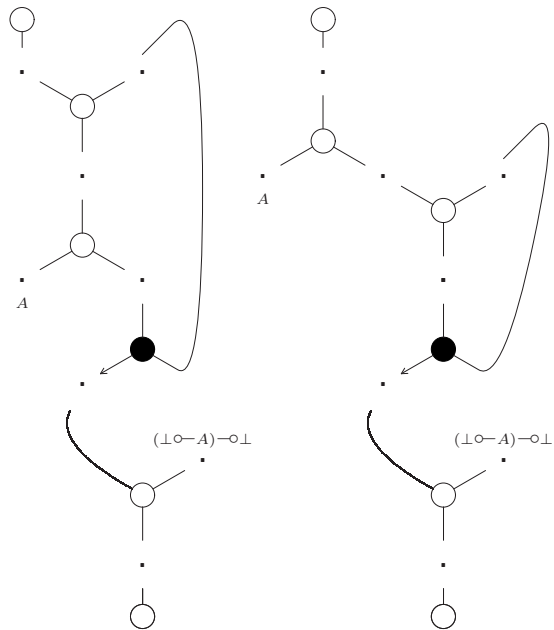

\begin{figure}
 \begin{center}
\scalebox{0.8}{%
  \begin{picture}(18,7)(0,9)
   \put(4,8){\tensorunita{}{$\nnode{}{}$}}
   \put(2,11.4641016){\tensorc{}{$\nnode{(\perp \blolli A) \lolli \perp}{}$}{$\nnode{}{A}$}{}}

   \put(14,11.4641016){\makebox(0,0){$\nnode{(\perp \blolli A) \lolli \perp}{A}$}}
  \end{picture}}
  \caption{Left identity for par and right identity for tensor}
  \label{grishinend}
 \end{center}
\end{figure}
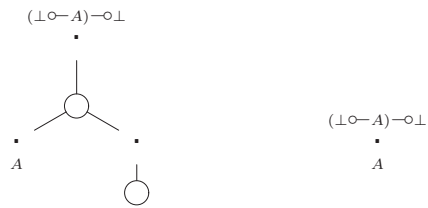

\begin{example}
Using $\grishina$, $\grishinb$ and the right identity for tensor and
left identity for par, we can derive $(\perp \blolli A) \lolli \perp \vdash
A$. Figures~\ref{grishin2} to \ref{grishinend} show how the abstract proof
structure of Figure~\ref{grishin1} can be contracted.
\end{example}

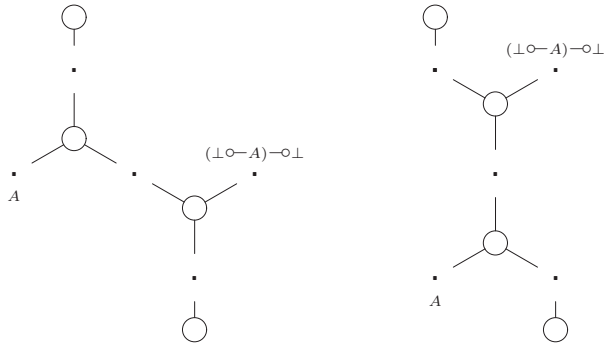
\begin{figure}
 \begin{center}
\scalebox{0.8}{%
  \begin{picture}(18,10)(0,10)
   \put(2,8){\tensorunita{}{}}
   \put(-2,14.9282032){\tensorc{}{}{$\nnode{}{A}$}{}}
   \put(2,11.4641016){\mtensorc{}{$\nnode{}{}$}{$\nnode{}{}$}{$\nnode{(\perp \blolli A) \lolli \perp}{}$}}
   \put(-2,18.3923048){\tensorunitb{}{$\nnode{}{}$}}

   \put(14,8){\tensorunita{}{$\nnode{}{}$}}
   \put(12,11.4641016){\tensorc{}{}{$\nnode{}{A}$}{}}
   \put(12,14.9282032){\mtensorc{}{$\nnode{}{}$}{$\nnode{}{}$}{$\nnode{(\perp \blolli A) \lolli \perp}{}$}}
   \put(10,18.3923048){\tensorunitb{}{$\nnode{}{}$}}
  \end{picture}}
 \end{center}
 \caption{The $\blolli$ contraction followed by $\grishina$}
\end{figure}

\section{Correctness}

We are now in a position to prove the main theorem: that derivability
in the sequent calculus and contractability in the proof net calculus
coincide.

\begin{theorem} A proof structure $\mathcal S$ is correct
  (ie.\ corresponds to a sequent proof of $\Gamma \vdash \Delta$) if
  and only if its abstract proof structure $\mathcal A$ converts to a
  tensor tree of $\Gamma \vdash \Delta$.
\end{theorem}

\paragraph{$\Rightarrow$} Suppose $\pi$ is a sequent calculus proof of
$\Gamma \vdash \Delta$. We construct a proof structure together with a
reduction sequence $\rho$ reducing it to tensor tree  $\Gamma
\vdash \Delta$ by induction on the depth $d$ of $\pi$.

If $d = 1$ then $\pi$ is one of the axioms. We conside each case
separately.

If the conclusion of the sequent is $\perp \vdash \epsilon$ then the
corresponding proof structure and abstract proof structure look as
shown below.

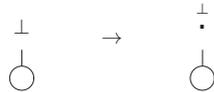
\begin{center}
\scalebox{0.8}{%
\begin{picture}(8,4)
 \put(0,-1){\tensorunita{}{\perp}}
 \put(5,2){\makebox(0,0){$\rightarrow$}}
 \put(6,-1){\tensorunita{}{\nnode{\perp}{}}}
\end{picture}}
\end{center}

Note how this is a proof net of $\perp \vdash \epsilon$ as required.

Similarly, if the conclusion of the axiom is $\epsilon \vdash \one$
then the corresponding proof structure and abstract proof structure
form a proof net of this sequent as shown below.

\begin{center}
\scalebox{0.8}{%
\begin{picture}(8,4)
 \put(0,1){\tensorunitb{}{\one}}
 \put(5,2){\makebox(0,0){$\rightarrow$}}
 \put(6,1){\tensorunitb{}{\nnode{}{\one}}}
\end{picture}}
\end{center}

Finally, if the conclusion of the axiom is $A \vdash A$ for some
formula $A$ then we are in the following situation

\begin{center}
\scalebox{0.8}{%
\begin{picture}(8,4)
 \put(2,2){\makebox(0,0){$A$}}
 \put(5,2){\makebox(0,0){$\rightarrow$}}
 \put(8,2){\makebox(0,0){$\nnode{A}{A}$}}
\end{picture}}
\end{center}

\noindent which is a proof net of $A \vdash A$.

If $d > 1$ then we look at the last rule in the proof. Suppose it is a
$[\textit{L}\pr]$ rule.

$$
\infer[\bo\textit{L}\pr\bc]{A\pr B \vdash \Delta_1
    \circ \Delta_2}{\infer*[\pi_1]{A\vdash \Delta_1}{} & \infer*[\pi_2]{B\vdash \Delta_2}{}}
$$

Given that both $\pi_1$ and $\pi_2$ have a depth smaller than $d$, we
can apply to induction hypothesis to obtain a proof structure $S_1$
with hypothesis $A$ which reduces to a tensor tree of $A \vdash
\Delta_1$ by reduction sequence $\rho_1$ and a proof structure $S_2$
with hypothesis $B$ which reduces to a tensor tree of $B \vdash
\Delta_2$ by reduction sequence $\rho_2$.

We can combine these two proof nets as shown below.

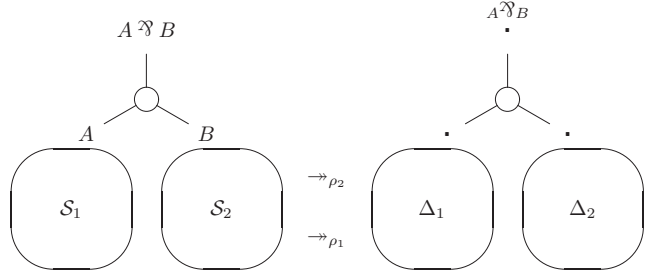
\begin{center}
\scalebox{0.8}{%
\begin{picture}(19,8)
   \put(1,4){\tensorc{}{A\pr B}{A}{B}}
   \put(0.5,1.5){\oval(4,4)}
   \put(5.5,1.5){\oval(4,4)}
   \put(0.5,1.5){\makebox(0,0){$\mathcal S_1$}}
   \put(5.5,1.5){\makebox(0,0){$\mathcal S_2$}}
   \put(9,0.5){\makebox(0,0){$\twoheadrightarrow_{{\rho_1}}$}}
   \put(9,2.5){\makebox(0,0){$\twoheadrightarrow_{{\rho_2}}$}}

   \put(13,4){\tensorc{}{\nnode{A\pr B}{}}{\nnode{}{}}{\nnode{}{}}}
   \put(12.5,1.5){\oval(4,4)}
   \put(17.5,1.5){\oval(4,4)}
   \put(12.5,1.5){\makebox(0,0){$\Delta_1$}}
   \put(17.5,1.5){\makebox(0,0){$\Delta_2$}}
\end{picture}}
\end{center}

Note that, since
$\rho_1$ and $\rho_2$ operate on different parts of the resulting
abstract proof structure, any interleaving of $\rho_1$ and $\rho_2$
will provide a valid reduction sequence $\rho$ producing a proof net
of $A \pr B \vdash \Delta_1 \circ \Delta_2$.

Suppose the last rule is a $[\textit{R}\pr]$ rule.

$$
\infer[\bo\textit{R}\pr\bc]{\Gamma \vdash A\pr
  B}{\infer*[\pi_1]{\Gamma \vdash A \circ B}{}}
$$

Given that $\pi_1$ has a depth of $d-1$ we can apply the induction
hypotesis to give us a proof structure with conclusions $A$ and $B$
which converts to a tensor tree of $\Gamma \vdash A \pr B$ by a
conversion sequence $\rho_1$.

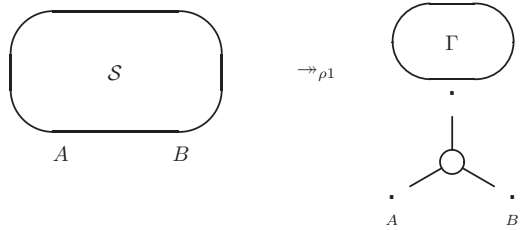
\begin{center}
\scalebox{0.8}{%
\begin{picture}(26,8.0)(-4,2.5)
\thicklines
%   \put(-3,2){\parbottomc{}{A\pr B}{A}{B}}
   \put(-1.2,8){\oval(7,4)}
   \put(-1.2,8){\makebox(0,0){$\mathcal S$}}
   \put(-3,5.3){\makebox(0,0){$A$}}
   \put(1,5.3){\makebox(0,0){$B$}}

%   \put(-1.2,-0.7){\oval(7,4)}
%   \put(-1.2,-0.7){\makebox(0,0){$\mathcal S$}}
%   \put(-1,2){\makebox(0,0){$A\pr B$}}
%   \put(8,0.4){\parbottomc{}{\nnode{}{}}{}{}{}}
   \put(8,3.8641016){\tensorc{}{\nnode{}{}}{\nnode{}{A}}{\nnode{}{B}}}

   \put(10,9){\oval(4,2.5)}
%   \put(10,-1.5){\oval(4,2.5)}
   \put(10,9){\makebox(0,0){$\Gamma$}}
%   \put(10,-1.5){\makebox(0,0){$\Delta$}}
%   \put(10,0.4){\makebox(0,0){\nnode{A\pr B}{}}}

  \put(5.5,8){\makebox(0,0){$\twoheadrightarrow_{\rho 1}$}}
%  \put(5.5,-0.7){\makebox(0,0){$\twoheadrightarrow_{\rho 2}$}}
\end{picture}}
\end{center}

We can add the par link for $A\pr B$ to the proof structure above
after which the reduction sequent $\rho_1$ produces a redex for the
$[\textit{R}\pr]$ contraction. We append this contraction at the end
of $\rho_1$ to produce the final contraction sequence $\rho$,
producing a proof net of $\Gamma \vdash A \pr B$ as required.

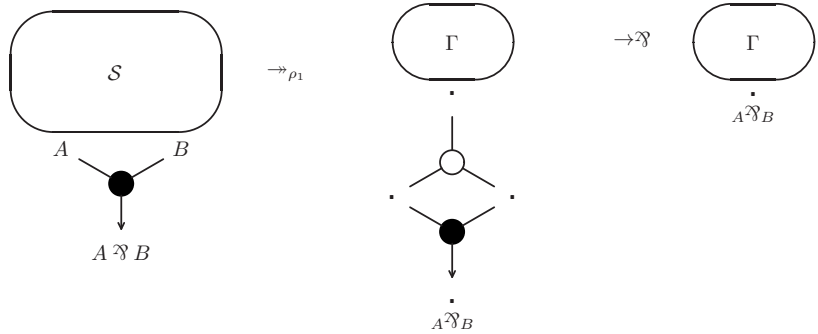
\begin{center}
\scalebox{0.8}{%
\begin{picture}(26,11.5)(-4,-1)
\thicklines
   \put(-3,2){\parbottomc{}{A\pr B}{A}{B}}
   \put(-1.2,8){\oval(7,4)}
   \put(-1.2,8){\makebox(0,0){$\mathcal S$}}

%   \put(-1.2,-0.7){\oval(7,4)}
%   \put(-1.2,-0.7){\makebox(0,0){$\mathcal S_2$}}
   \put(8,0.4){\parbottomc{}{\nnode{}{A\pr B}}{}{}{}}
   \put(8,3.8641016){\tensorc{}{\nnode{}{}}{\nnode{}{}}{\nnode{}{}}}

   \put(10,9){\oval(4,2.5)}
 %  \put(10,-1.5){\oval(4,2.5)}
   \put(10,9){\makebox(0,0){$\Gamma$}}
 %  \put(10,-1.5){\makebox(0,0){$\Delta$}}

   \put(20,9){\oval(4,2.5)}
%   \put(20,2){\oval(4,2.5)}
   \put(20,7.3){\makebox(0,0){\nnode{}{A\pr B}}}
   \put(20,9){\makebox(0,0){$\Gamma$}}
%   \put(20,2){\makebox(0,0){$\Delta$}}
   \put(16,9){\makebox(0,0){$\rightarrow_{\pr}$}}
   \put(4.5,8){\makebox(0,0){$\twoheadrightarrow_{{\rho_1}}$}}
\end{picture}}
\end{center}

The other logical rules are similar and easily verified.

\paragraph{$\Leftarrow$} Suppose $\mathcal A$ corresponding to $\mathcal S$ converts to a
tensor tree $\mathcal T$ by means of a conversion sequence $s$. We proceeding by
induction on the length $l$ of $s$ to constuct a sequent proof of
$\mathcal T$. In what follows, I will often use `a derivation $d$ of
$\Gamma \vdash \Delta$' where it would be more precise but also more
cumbersome to use `a derivation $d$ which can be extended using only
the display rules to a derivation of $\Gamma \vdash \Delta$. I trust
this will not lead to confusion.

If $l=0$ then $\mathcal{A} = \mathcal{T}$. We proceed by
induction on the number of connectors $c$ in $\mathcal{T}$. 

\begin{description}
\item{If $c=0$} then $\mathcal{S}$ and $\mathcal{T}$ are of the form 

\begin{center}
\sk{.5}
$A \ \rightarrow \ \ \nnode{A}{A}$ 
\sk{1}
\end{center}

\noindent which corresponds to the sequent proof

\begin{center}
\sk{.5}
\mbox{\infer[\bo\textit{Ax}\bc]{A \vdash A}{}}
\sk{.5}
\end{center}

\item{If $c>0$} then by Lemma~\ref{lem:princleaf}, we know that $\mathcal{S}$
has a formula which is the main leaf of its link, call it $D$. We proceed by
case analysis. If $D$ is of the form $A\pr B$, then we are in the
following situation.

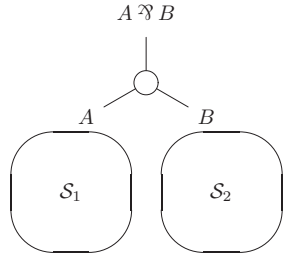
\begin{center}
\scalebox{0.8}{%
\begin{picture}(8,8)
   \put(2,4){\tensorc{}{A\pr B}{A}{B}}
   \put(1.5,1.5){\oval(4,4)}
   \put(6.5,1.5){\oval(4,4)}
   \put(1.5,1.5){\makebox(0,0){$\mathcal S_1$}}
   \put(6.5,1.5){\makebox(0,0){$\mathcal S_2$}}
\end{picture}}
\end{center}

Because the proof structure is a tree, the par link separates the
structure into two parts:  $\mathcal{S}_1$ with hypothesis $A$ and
$\mathcal{S}_2$ with hypothesis $B$. By induction hypothesis, there
are derivations $d_1$ of $A \vdash \Delta_1$ and $d_2$ of $B \vdash
\Delta_2$, which we can combine as follows.

$$
\infer[\bo\textit{L}\pr\bc]{A\pr B \vdash \Delta_1 \circ \Delta_2}{A\vdash \Delta_1 & B\vdash \Delta_2}
$$

The other cases are similar.
\end{description}

Suppose now $l > 0$. We look at the last conversion. 

% NOTE: is it a good idea to add the axiom connection here? The case
% where axiom splits into two proof structures is easy (but it's still
% a bit strange when axiom corresponds to cut) but for non-splitting
% axioms we need some more convoluted reasoning, like: one of the
% leaves must be attached to a par link, otherwise we couldn't
% contract to a tensor tree.
%Suppose the last conversion is an axiom connection. This means we are
%schematically in the following situation.

Suppose the last conversion is a structural conversion, then we are
schematically in the following situation. % (but note that the $\Gamma_i$ and
% $\Delta_i$ need not be disjoint).

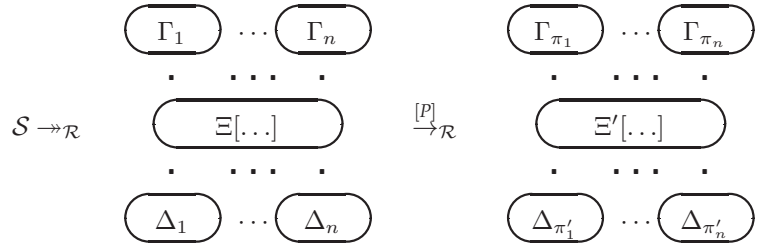
\begin{center}
\begin{tabular}{CCCCCCCCCCCCCC}
&&&&\ovaalb\ovaalt\hier{\Gamma_{1}}&&\hier{\ \ \cdots}
&&\ovaalb\ovaalt\hier{\Gamma_{n}}\\
&&&&\hier{\node{}{}}&&\hier{\ \ \nodes{}{}}&&\hier{\node{}{}}\\
\hier{\cals}&\hier{\twoheadrightarrow_{\calr}}&&&&&\makebox[0pt]{\begin{picture}(0,0)(-.5,-.5)
\thicklines
\put(0,0){\oval(5,1.25)} 
\end{picture}}\hier{\Xi[\ldots]}&&&&&\hier{\ralabel{[\textit P]}{\calr}}\\
&&&&\hier{\node{}{}}&&\hier{\ \ \nodes{}{}}&&\hier{\node{}{}}\\
&&&&\ovaalb\ovaalt\hier{\Delta_{1}}&&\hier{\ \ \cdots}
&&\ovaalb\ovaalt\hier{\Delta_{n}}\\
\end{tabular}
\begin{tabular}{CCCCCCCCCCC}
&&\ovaalb\ovaalt\hier{\Gamma_{\pi_1}}&&\hier{\ \ \cdots}
&&\ovaalb\ovaalt\hier{\Gamma_{\pi_n}}\\
&&\hier{\node{}{}}&&\hier{\ \ \nodes{}{}}&&\hier{\node{}{}}\\
&&&&\makebox[0pt]{\begin{picture}(0,0)(-.5,-.5)
\thicklines
\put(0,0){\oval(5,1.25)} 
\end{picture}}\hier{\Xi'[\ldots]}&&\\
&&\hier{\node{}{}}&&\hier{\ \ \nodes{}{}}&&\hier{\node{}{}}\\
&&\ovaalb\ovaalt\hier{\Delta_{\pi'_1}}&&\hier{\ \ \cdots}
&&\ovaalb\ovaalt\hier{\Delta_{\pi'_n}}\\
\end{tabular}
\end{center}

Given that we have a structural conversion $\bo\textit{P}\bc$ we know
there is at least one structural rule which corresponds to it, where
one of the leaves of both $\Xi$ and $\Xi'$ is displayed. In case this leaf
is a hypothesis of $\Xi$ (assume it's $\Gamma_i$), induction
hypothesis gives us a proof $d$ which we can extend as follows.

\begin{center}
\mbox{\infer[\bo\textit{P}\bc]{\Gamma_i \vdash
    \Xi'}{\infer*[d]{\Gamma_i \vdash \Xi}{}}}
\end{center}

In case the leaf is a conclusion $\Delta_i$ we operate symmetrically,
obtaining.

\begin{center}
\mbox{\infer[\bo\textit{P}\bc]{\Xi' \vdash
    \Delta_i}{\infer*[d]{\Xi \vdash \Delta_i}{}}}
\end{center}

Suppose the last conversion is a contraction. We proceed by case analysis.

\begin{description}
\item{$[\textit{R}.^{\perp}]$} In case the last conversion is a $.^{\perp}$ contraction we are
schematically in the following situation.

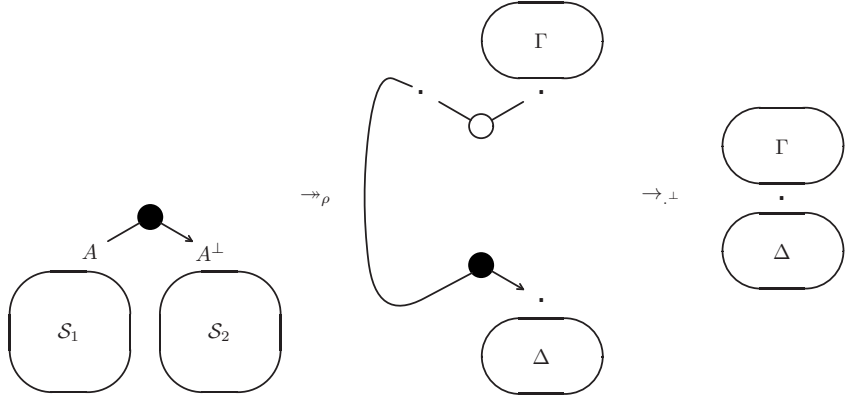
\begin{center}
\scalebox{0.8}{%
\begin{picture}(26,13)(-4,-2.5)
\thicklines
   \put(-3,2){\mnegparrightc{}{A^{\perp}}{A}}
   \put(-3.7,-0.7){\oval(4,4)}
   \put(1.3,-0.7){\oval(4,4)}
   \put(-3.7,-0.7){\makebox(0,0){$\mathcal S_1$}}
   \put(1.3,-0.7){\makebox(0,0){$\mathcal S_2$}}

   \put(8,0.4){\mnegparrightcb{}{\nnode{}{}}{}}
   \put(8,3.8641016){\negtensorb{}{\nnode{}{}}{\nnode{}{}}}
   \spline(9.7,1.3)(6.5,-0.4)(6,3.8)(6.5,8.0)(7.7,7.5)

   \put(12,9){\oval(4,2.5)}
   \put(12,-1.5){\oval(4,2.5)}
   \put(12,9){\makebox(0,0){$\Gamma$}}
   \put(12,-1.5){\makebox(0,0){$\Delta$}}

   \put(20,5.5){\oval(4,2.5)}
   \put(20,2){\oval(4,2.5)}
   \put(20,3.8){\makebox(0,0){\nnode{}{}}}
   \put(20,5.5){\makebox(0,0){$\Gamma$}}
   \put(20,2){\makebox(0,0){$\Delta$}}
   \put(16,3.8){\makebox(0,0){$\rightarrow_{.^\perp}$}}
   \put(4.5,3.8){\makebox(0,0){$\twoheadrightarrow_{\rho}$}}
\end{picture}}
\end{center}

Looking backwards from the endsequent, the par link forms a barrier:
every structural rewrite has to be performed either fully in $\Gamma$
--- where it will finally end up producing ${\mathcal S}_1$ --- or
fully in $\Delta$ --- where it will finally end up producing
${\mathcal S}_2$. From this perspective, every contraction simply
expands a single node and is therefore performed in just one of the
two substructures as well. Therefore, we can separate the conversions
of $\rho$ into those which are fully in ${\mathcal S}_1$ reducing it to $\Gamma
\vdash \lceil A \rceil$ and those which are fully in ${\mathcal S}_2$
reducing it to $A^{\perp} \vdash \Delta$. We will call these two
reduction sequences $\rho_1$ and $\rho_2$ respectively.

Removing the par link from the figure above gives us the
following two proof structures with their corresponding reduction
sequences.

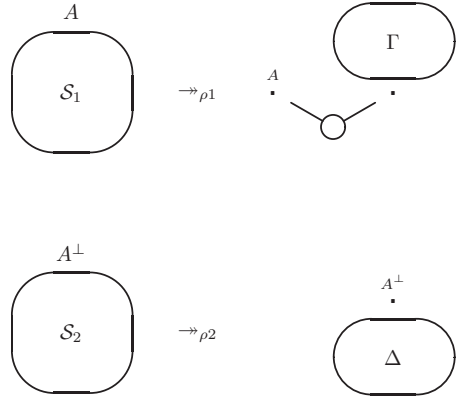
\begin{center}
\scalebox{0.8}{%
\begin{picture}(26,13)(-3,-2.5)
\thicklines
   \put(1.3,7.3){\oval(4,4)}
   \put(1.3,7.3){\makebox(0,0){$\mathcal S_1$}}
   \put(1.3,-0.7){\oval(4,4)}
   \put(1.3,-0.7){\makebox(0,0){$\mathcal S_2$}}
%   \put(-3,2){\mnegparrightc{}{A^{\perp}}{A}}
  \put(1.3,10){\makebox(0,0){$A$}}
  \put(1.3,2){\makebox(0,0){$A^{\perp}$}}
%   \put(-1.2,-0.7){\oval(7,4)}
%   \put(-1.2,-0.7){\makebox(0,0){$\mathcal S$}}

%   \put(8,0.4){\mnegparrightcb{}{\nnode{}{}}{}}
   \put(12,0.4){\makebox(0,0){\nnode{A^{\perp}}{}}}
   \put(8,3.8641016){\negtensorb{}{\nnode{A}{}}{\nnode{}{}}}

   \put(12,9){\oval(4,2.5)}
   \put(12,-1.5){\oval(4,2.5)}
   \put(12,9){\makebox(0,0){$\Gamma$}}
   \put(12,-1.5){\makebox(0,0){$\Delta$}}
  \put(5.5,7.3){\makebox(0,0){$\twoheadrightarrow_{\rho 1}$}}
  \put(5.5,-0.7){\makebox(0,0){$\twoheadrightarrow_{\rho 2}$}}
\end{picture}}
\end{center}

Since the length of $\rho_1 + \rho_2$ is less than the length of
$\rho$ --- the final contraction being removed --- we can apply the
induction hypothesis to give us a proof $d_1$ of $\Gamma \vdash
\lceil A \rceil$ and a proof $d_2$ of $A^{\perp} \vdash \Delta$. We
can combine these two proofs into a proof of $\Gamma \vdash \Delta$ as
follows.

$$
\infer[\bo\textit{Cut}\bc]{\Gamma \vdash \Delta}{
   \infer[\bo\textit{R}.^{\perp}\bc]{\Gamma \vdash A^{\perp}}{
      \infer*[d_1]{\Gamma \vdash \lceil A \rceil}{}
   }
   &
   \infer*[d_2]{A^{\perp} \vdash \Delta}{}
}
$$

\item{$[\textit{R}\rule{0pt}{1ex}^{\perp}\!.]$} Symmetric.

\item{$[\textit{L}\otimes]$} In case the last contraction is a $\otimes$
  contraction, the proof structure and the conversion sequence for the
  corresponding abstract proof structure look as shown below.

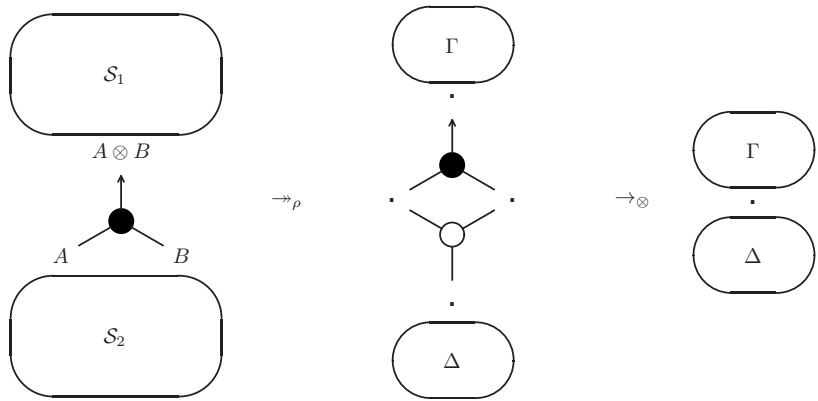
\begin{center}
\scalebox{0.8}{%
\begin{picture}(26,13.5)(-4,-2.5)
\thicklines
   \put(-3,2){\mparbottomc{}{A\otimes B}{A}{B}}
   \put(-1.2,8){\oval(7,4)}
   \put(-1.2,8){\makebox(0,0){$\mathcal S_1$}}

   \put(-1.2,-0.7){\oval(7,4)}
   \put(-1.2,-0.7){\makebox(0,0){$\mathcal S_2$}}
   \put(8,3.8641016){\mparbottomc{}{\nnode{}{}}{}{}{}}
   \put(8,0.4){\mtensorc{}{\nnode{}{}}{\nnode{}{}}{\nnode{}{}}}

   \put(10,9){\oval(4,2.5)}
   \put(10,-1.5){\oval(4,2.5)}
   \put(10,9){\makebox(0,0){$\Gamma$}}
   \put(10,-1.5){\makebox(0,0){$\Delta$}}

   \put(20,5.5){\oval(4,2.5)}
   \put(20,2){\oval(4,2.5)}
   \put(20,3.8){\makebox(0,0){\nnode{}{}}}
   \put(20,5.5){\makebox(0,0){$\Gamma$}}
   \put(20,2){\makebox(0,0){$\Delta$}}
   \put(16,3.8){\makebox(0,0){$\rightarrow_{\otimes}$}}
   \put(4.5,3.8){\makebox(0,0){$\twoheadrightarrow_{\rho}$}}
\end{picture}}
\end{center}

We again eliminate the par link and its contraction and partition the
remaining conversions over two disjoint sequences as shown below.

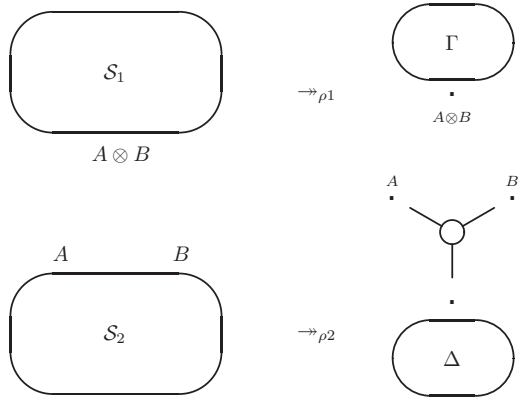
\begin{center}
\scalebox{0.8}{%
\begin{picture}(26,13.5)(-4,-2.5)
\thicklines
%   \put(-3,2){\mparbottomc{}{A\otimes B}{A}{B}}
   \put(-1.2,8){\oval(7,4)}
   \put(-1.2,8){\makebox(0,0){$\mathcal S_1$}}
   \put(-3,2){\makebox(0,0){$A$}}
   \put(1,2){\makebox(0,0){$B$}}

   \put(-1,5.3){\makebox(0,0){$A\otimes B$}}

   \put(-1.2,-0.7){\oval(7,4)}
   \put(-1.2,-0.7){\makebox(0,0){$\mathcal S_2$}}
%   \put(8,3.8641016){\mparbottomc{}{\nnode{}{}}{}{}{}}
   \put(8,0.4){\mtensorc{}{\nnode{}{}}{\nnode{A}{}}{\nnode{B}{}}}

   \put(10,9){\oval(4,2.5)}
   \put(10,-1.5){\oval(4,2.5)}
   \put(10,7.3){\makebox(0,0){\nnode{}{A\otimes B}}}
   \put(10,9){\makebox(0,0){$\Gamma$}}
   \put(10,-1.5){\makebox(0,0){$\Delta$}}

  \put(5.5,7.3){\makebox(0,0){$\twoheadrightarrow_{\rho 1}$}}
  \put(5.5,-0.7){\makebox(0,0){$\twoheadrightarrow_{\rho 2}$}}
 \end{picture}}
\end{center}

Now the induction hypothesis gives us a derivation $d_1$ of $\Gamma
\vdash A \otimes B$ and a derivation $d_2$ of $A \circ B \vdash
\Delta$. We combine these two derivations into a derivation of $\Gamma
\vdash \Delta$ as follows.

$$
\infer[\bo\textit{Cut}\bc]{\Gamma \vdash \Delta}{
   \infer*[d_1]{\Gamma \vdash A\otimes B}{}
 & \infer[\bo\textit{L}\otimes\bc]{A\otimes B \vdash \Delta}{
      \infer*[d_2]{A \circ B \vdash \Delta}{}
   }
}
$$

\item{$[\textit{R}\lolli]$} If the last contraction is a $\lolli$ contraction, the proof structure
and reduction sequence look as follows.

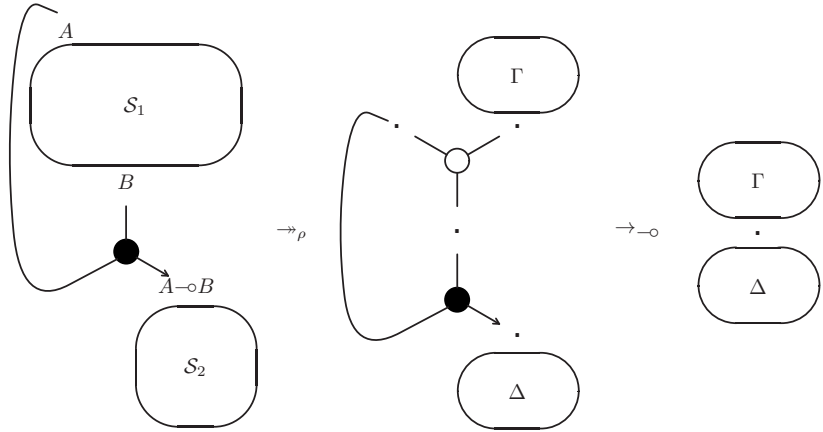
\begin{center}
\scalebox{0.8}{%
\begin{picture}(26,13.5)(-4,-2.5)
\thicklines
   \put(-3,2){\mparrightcb{}{A\lolli B}{B}}
   \put(-0.7,8){\oval(7,4)}
%   \put(-1.2,8){\oval(7,4)[t]}
%   \put(-2.7,4.0){\oval(4,12)[l]}
   \put(-0.7,8){\makebox(0,0){$\mathcal S_1$}}
%   \put(-1.2,-0.7){\makebox(0,0){$\mathcal S_2$}}
   \put(1.3,-0.7){\oval(4,4)}
   \put(1.3,-0.7){\makebox(0,0){$\mathcal S_2$}}
   \spline(-1.3,2.9)(-4.5,1.2)(-5,6.4)(-4.5,11.6)(-3.3,11.1)
   \put(-3,10.5){\makebox(0,0){$A$}}
   \put(8,0.4){\mparrightcb{}{\nnode{}{}}{}}
   \put(8,3.8641016){\mtensorc{}{\nnode{}{}}{\nnode{}{}}{\nnode{}{}}}
   \spline(9.7,1.3)(6.5,-0.4)(6,3.8)(6.5,8.0)(7.7,7.5)

   \put(12,9){\oval(4,2.5)}
   \put(12,-1.5){\oval(4,2.5)}
   \put(12,9){\makebox(0,0){$\Gamma$}}
   \put(12,-1.5){\makebox(0,0){$\Delta$}}

   \put(20,5.5){\oval(4,2.5)}
   \put(20,2){\oval(4,2.5)}
   \put(20,3.8){\makebox(0,0){\nnode{}{}}}
   \put(20,5.5){\makebox(0,0){$\Gamma$}}
   \put(20,2){\makebox(0,0){$\Delta$}}
   \put(16,3.8){\makebox(0,0){$\rightarrow_{\lolli}$}}
   \put(4.5,3.8){\makebox(0,0){$\twoheadrightarrow_{\rho}$}}
\end{picture}}
\end{center}

As before we remove the par link and its contraction and separate the
conversion sequences which are in $\Gamma$ and $\Delta$. The result is
shown below.

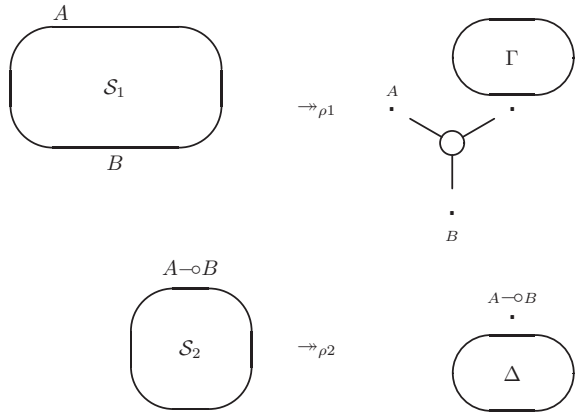
\begin{center}
\scalebox{0.8}{%
\begin{picture}(26,13.5)(-4,-2.5)
\thicklines
   \put(-1.2,8){\oval(7,4)}
   \put(-1.2,8){\makebox(0,0){$\mathcal S_1$}}
   \put(1.3,-0.7){\oval(4,4)}
   \put(1.3,-0.7){\makebox(0,0){$\mathcal S_2$}}
   \put(-3,10.5){\makebox(0,0){$A$}}
   \put(-1.2,5.5){\makebox(0,0){$B$}}
   \put(1.3,2){\makebox(0,0){$A \lolli B$}}
   \put(8,3.8641016){\mtensorc{}{\nnode{}{B}}{\nnode{A}{}}{\nnode{}{}}}

   \put(12,9){\oval(4,2.5)}
   \put(12,-1.5){\oval(4,2.5)}
   \put(12,9){\makebox(0,0){$\Gamma$}}
   \put(12,-1.5){\makebox(0,0){$\Delta$}}
   \put(12,0.4){\makebox(0,0){\nnode{A\lolli B}{}}}

  \put(5.5,7.3){\makebox(0,0){$\twoheadrightarrow_{\rho 1}$}}
  \put(5.5,-0.7){\makebox(0,0){$\twoheadrightarrow_{\rho 2}$}}
\end{picture}}
\end{center}

Induction hypothesis now gives us a derivation $d_1$ from ${\mathcal
  S}_1$ to $\Gamma \vdash A > B$ and a derivation $d_2$ from ${\mathcal
  S}_2$ to $A \lolli B \vdash \Delta$. We can combine these proofs in
the following way.

$$
\infer[\bo\textit{Cut}\bc]{\Gamma \vdash \Delta}{
   \infer[\bo\textit{R}\lolli\bc]{\Gamma \vdash A \lolli B}{
      \infer*[d_1]{\Gamma \vdash A > B}{}
   }
   &
   \infer*[d_2]{A\lolli B \vdash \Delta}{}
}
$$

\item{$[\textit{R}\blolli]$} Symmetric.

\item{$[\textit{R}\pr]$} If the last contraction is a $\pr$ contraction, the proof structure
and reduction sequence look as follows.

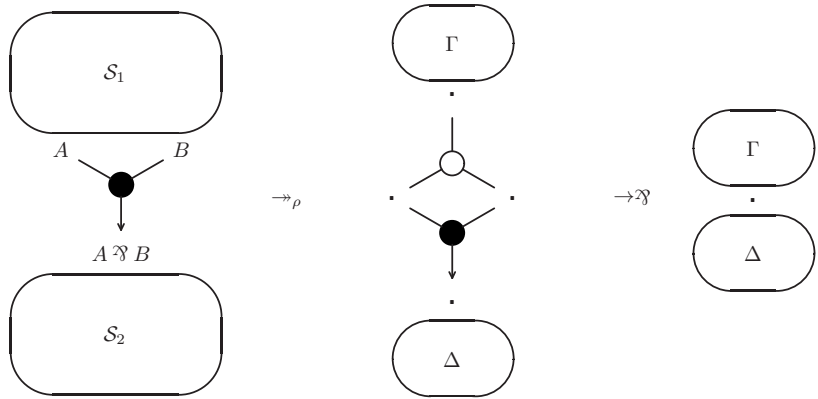
\begin{center}
\scalebox{0.8}{%
\begin{picture}(26,13.5)(-4,-2.5)
\thicklines
   \put(-3,2){\parbottomc{}{A\pr B}{A}{B}}
   \put(-1.2,8){\oval(7,4)}
   \put(-1.2,8){\makebox(0,0){$\mathcal S_1$}}

   \put(-1.2,-0.7){\oval(7,4)}
   \put(-1.2,-0.7){\makebox(0,0){$\mathcal S_2$}}
   \put(8,0.4){\parbottomc{}{\nnode{}{}}{}{}{}}
   \put(8,3.8641016){\tensorc{}{\nnode{}{}}{\nnode{}{}}{\nnode{}{}}}

   \put(10,9){\oval(4,2.5)}
   \put(10,-1.5){\oval(4,2.5)}
   \put(10,9){\makebox(0,0){$\Gamma$}}
   \put(10,-1.5){\makebox(0,0){$\Delta$}}

   \put(20,5.5){\oval(4,2.5)}
   \put(20,2){\oval(4,2.5)}
   \put(20,3.8){\makebox(0,0){\nnode{}{}}}
   \put(20,5.5){\makebox(0,0){$\Gamma$}}
   \put(20,2){\makebox(0,0){$\Delta$}}
   \put(16,3.8){\makebox(0,0){$\rightarrow_{\pr}$}}
   \put(4.5,3.8){\makebox(0,0){$\twoheadrightarrow_{\rho}$}}
\end{picture}}
\end{center}

Removing the par link and splitting the remaining conversions over the
two substructures will give us the situation shown below.

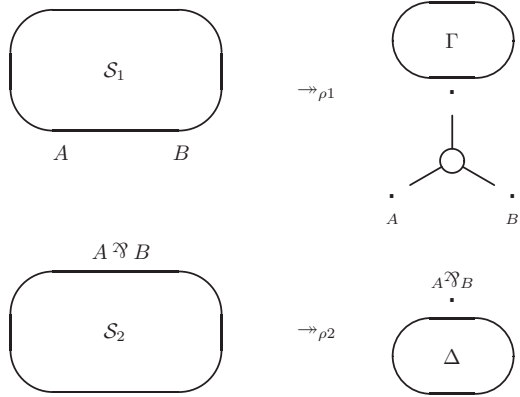
\begin{center}
\scalebox{0.8}{%
\begin{picture}(26,13.5)(-4,-2.5)
\thicklines
%   \put(-3,2){\parbottomc{}{A\pr B}{A}{B}}
   \put(-1.2,8){\oval(7,4)}
   \put(-1.2,8){\makebox(0,0){$\mathcal S_1$}}
   \put(-3,5.3){\makebox(0,0){$A$}}
   \put(1,5.3){\makebox(0,0){$B$}}

   \put(-1.2,-0.7){\oval(7,4)}
   \put(-1.2,-0.7){\makebox(0,0){$\mathcal S_2$}}
   \put(-1,2){\makebox(0,0){$A\pr B$}}
%   \put(8,0.4){\parbottomc{}{\nnode{}{}}{}{}{}}
   \put(8,3.8641016){\tensorc{}{\nnode{}{}}{\nnode{}{A}}{\nnode{}{B}}}

   \put(10,9){\oval(4,2.5)}
   \put(10,-1.5){\oval(4,2.5)}
   \put(10,9){\makebox(0,0){$\Gamma$}}
   \put(10,-1.5){\makebox(0,0){$\Delta$}}
   \put(10,0.4){\makebox(0,0){\nnode{A\pr B}{}}}

  \put(5.5,7.3){\makebox(0,0){$\twoheadrightarrow_{\rho 1}$}}
  \put(5.5,-0.7){\makebox(0,0){$\twoheadrightarrow_{\rho 2}$}}
\end{picture}}
\end{center}

We apply the induction hypothesis to obtain a sequent proof $d_1$ of $\Gamma
\vdash A \circ B$ and a sequent proof $d_2$ of $A\pr B \vdash \Delta$
and combine these two proofs as follows.

$$
\infer[\bo\textit{Cut}\bc]{\Gamma \vdash \Delta}{
   \infer[\bo\textit{R}\pr\bc]{\Gamma \vdash A \pr B}{
      \infer*[d_1]{\Gamma \vdash A \circ B}{}
   }
   &
   \infer*[d_2]{A\pr B \vdash \Delta}{}
}
$$

\item The other cases are similar \qed

\end{description}

\section{Complexity}

In this section I will discuss the computational complexity of the
contraction criterion for several different fragments of the proof net
calculus.

\subsection{Binary Without Structural Conversions}

A first case is to decide the contractibility of a proof structure
containing only binary links and without any structural
conversions. When we look at the redexes of the different
contractions, we see that there is no possibility of overlap: a link
with three ports cannot be linked at two of its ports by two different
links while having each of its nodes be at most once a conclusion and
at most once a premiss of its link, as required by our definition of
proof nets.

So even a naive contraction strategy which traverses the graph in
search of contractible par links and contracts them as soon as they
are found then makes another pass untill it either fails to contract
any par links --- in which case the proof structure is not a proof net
--- or until there are no par links left --- in which case we
\emph{do} have a proof net. This gives us an $O(n^2)$ algorithm, where
$n$ is the number of links in the graph. Without too much effort, we
can improve this to $O(p^2)$ where $p$ is the number of par links in
the graph.

\subsection{Binary With Grishin Interactions}

A more complex case uses only the binary connectives but adds the
Grishin rules of Figures~\ref{fig:grishin} and \ref{fig:grishinmc}.

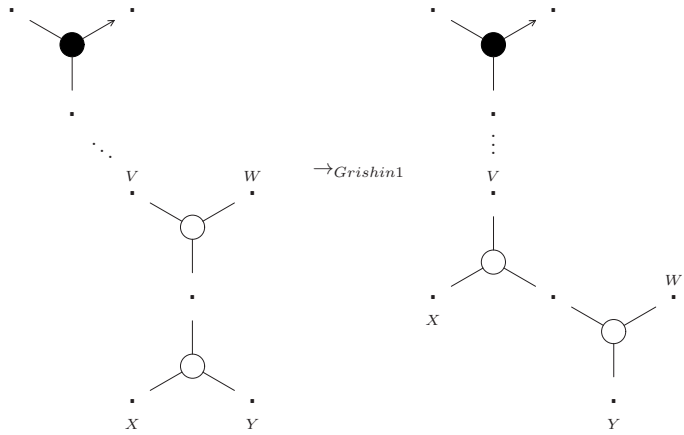
\begin{figure}
\begin{center}
\scalebox{0.8}{%
\begin{picture}(24,14.5)
\put(4,0.4641016){\tensorc{}{}{$\nnode{}{X}$}{$\nnode{}{Y}$}}
\put(4,3.9282032){\mtensorc{}{\nnode{}{}}{$\nnode{V}{}$}{$\nnode{W}{}$}}
\put(3,9){\makebox(0,0){$\ddots$}}
\put(0,10){\parrightc{}{\nnode{}{}}{\nnode{}{}}{\nnode{}{}}}

\put(10,8){$\rightarrow_{Grishin1}$}

\put(14,10){\parrightc{}{\nnode{}{}}{\nnode{}{}}{\nnode{}{}}}
\put(16,9.2){\makebox(0,0){$\vdots$}}
\put(14,3.9282032){\tensorc{}{$\nnode{V}{}$}{$\nnode{}{X}$}{\nnode{}{}}}
\put(18,0.4641016){\mtensorc{}{$\nnode{}{Y}$}{}{$\nnode{W}{}$}}

\end{picture}}
\end{center}
\caption{Moving a contensor link left towards the corresponding par link}
\label{fig:moveleft}
\end{figure}

Look at the $[\textit{L}\looparrowright]$ contraction of
Figure~\ref{fig:contr:bindualres} and suppose the top and bottom part of
the cotensor and par link are not already connected. For a Grishin
interaction to apply, we need the dual residuated cotensor link to be
connected to a residuated tensor link. There are two cases to
consider. If we can reach the par link by the left branch of the
tensor link, we apply the \textit{Grishin 1} rule as shown in
Figure~\ref{fig:moveleft}. The cotensor links moves upward an to the
left, reducing the distance to the par link.

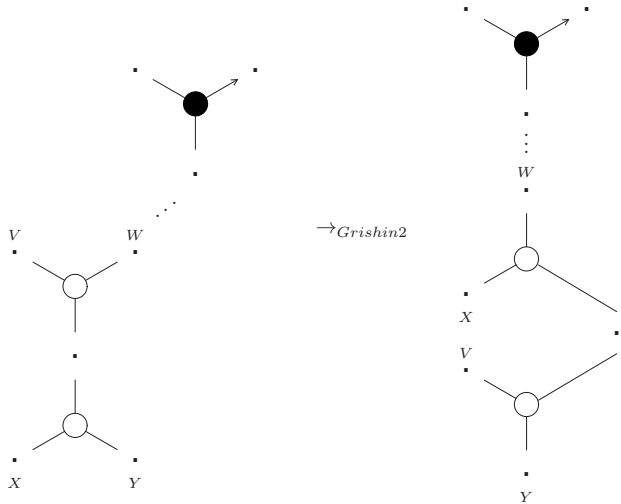
\begin{figure}
\begin{center}
\scalebox{0.8}{%
  \begin{picture}(24,16)
   \put(4,10){\parrightc{}{\nnode{}{}}{\nnode{}{}}{\nnode{}{}}}
   \put(15,6){\tensorcl{}{$\nnode{W}{}$}{$\nnode{}{X}$}{}}
   \put(15,0){\mtensorcleft{}{$\nnode{}{Y}$}{$\nnode{V}{}$}{}}
%  \drawline(3.4,0.34641012)(2.34641016,0.95470052)
%   \drawline(4.4,3.34641012)(3.34641016,3.95470052)
   \drawline(17.4,2.45470052)(20,4)
   \drawline(17.4,6.95470052)(20,5.4)
   \put(20,4.7){\makebox(0,0){\nnode{}{}}}
   \put(15,12){\parrightc{}{\nnode{}{}}{\nnode{}{}}{\nnode{}{}}}
   \put(17,11.2){\makebox(0,0){$\vdots$}}
%   \put(9.5,5.92){\makebox(0,0){$\rightarrow_{\grishine}$}}
%   \put(9.5,7.92){\makebox(0,0){$\leftarrow_{\grishinf}$}}
   \put(0,0.4641016){\tensorc{}{}{$\nnode{}{X}$}{$\nnode{}{Y}$}}
   \put(0,3.9282032){\mtensorc{}{\nnode{}{}}{$\nnode{V}{}$}{$\nnode{W}{}$}}
   \put(5,9){\makebox(0,0){$\ \iddots\ $}}
\put(10,8){$\rightarrow_{Grishin2}$}
  \end{picture}}
\end{center}\
\caption{Moving a contensor link right towards the corresponding par link}
\label{fig:moveright}
\end{figure}

Symmetrically, if the par link can be reached by the right branch of
the tensor link, we apply the \textit{Grishin 2} rule as shown in
Figure~\ref{fig:moveright}. We move the cotensor link up and to the right
and one step closer to the par link it needs to reach for its
contraction.

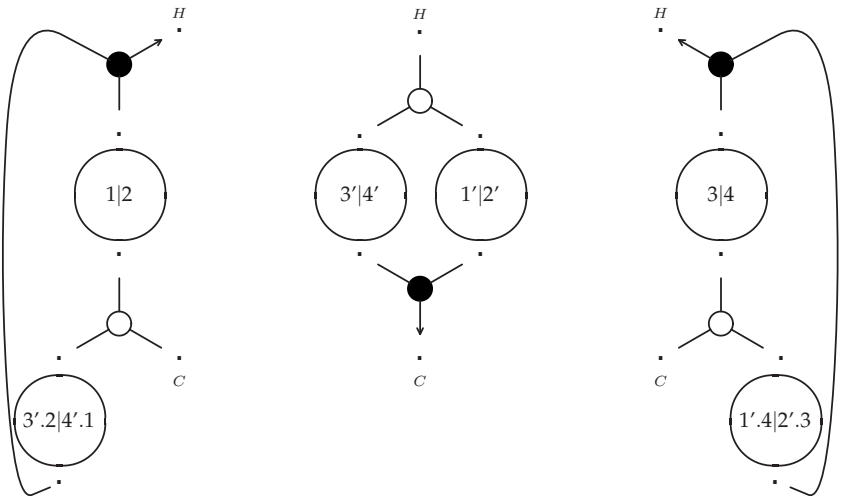
\begin{figure}
\begin{center}
\scalebox{0.8}{%
\begin{picture}(26,17)(0,-3)
\thicklines
   \put(1,7.8641015){\parrightcb{}{\nnode{H}{}}{\nnode{}{}}{}}
   \put(1,0.4){\tensorc{}{\nnode{}{}}{\nnode{}{}}{\nnode{}{C}}}
   \spline(0.7,-3.9)(-0.5,-4.4)(-1,3.8)(-0.5,12.0)(2.7,10.3)
   \put(3,5.8){\oval(3,3)}
   \put(3,5.8){\makebox(0,0){$\grishinnb | \grishinnf$}}
   \put(1,-1.7){\oval(3,3)}
   \put(1,-1.7){\makebox(0,0){$\grishinnd.\grishinnf | \grishinnh.\grishinnb$}}
   \put(1,-3.7){\makebox(0,0){\nnode{}{}}}

   \put(11,7.8){\tensorc{}{\nnode{H}{}}{\nnode{}{}}{\nnode{}{}}}
   \put(11,0.4){\parbottomc{}{\nnode{}{C}}{\nnode{}{}}{\nnode{}{}}}
   \put(11,5.8){\oval(3,3)}
   \put(11,5.8){\makebox(0,0){$\grishinnd | \grishinnh$}}
   \put(15,5.8){\oval(3,3)}
   \put(15,5.8){\makebox(0,0){$\grishinna | \grishinne$}}

   \put(21,7.8641016){\parleftcb{}{\nnode{H}{}}{\nnode{}{}}}
   \put(21,0.4){\tensorc{}{\nnode{}{}}{\nnode{}{C}}{\nnode{}{}}}
   \spline(25.3,-3.9)(26.5,-4.4)(27,3.8)(26.5,12.0)(23.3,10.3)
   \put(23,5.8){\oval(3,3)}
   \put(23,5.8){\makebox(0,0){$\grishinnc | \grishinng$}}
   \put(24.8,-1.7){\oval(3,3)}
   \put(24.8,-1.7){\makebox(0,0){$\grishinna.\grishinng | \grishinne.\grishinnc$}}
   \put(24.8,-3.7){\makebox(0,0){\nnode{}{}}}
\end{picture}}
\end{center}
\caption{Contractions: binary dual residuated with Grishin}
\label{fig:contr:bindualgrish}
\end{figure}

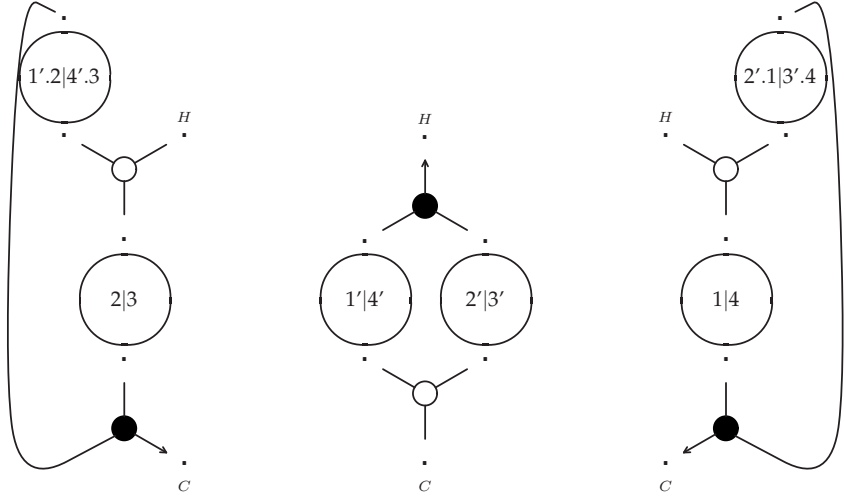
\begin{figure}
\begin{center}
\scalebox{0.8}{%
\begin{picture}(26,17)(0,0)
\thicklines
   \put(1,0.4){\mparrightcb{}{\nnode{}{C}}{\nnode{}{}}{}}
   \put(1,7.8641015){\mtensorc{}{\nnode{}{}}{\nnode{}{}}{\nnode{H}{}}}
%   \spline(0.7,-3.9)(-0.5,-4.4)(-1,3.8)(-0.5,12.0)(2.7,10.3) inv
   \spline(2.7,1.3)(-0.5,-0.4)(-1,3.8)(-0.5,16.0)(0.7,15.4)
%   \spline(2.7,1.3)(-0.5,-0.4)(-1,3.8)(-0.5,8.0)(0.7,7.5) orig
   \put(3,5.8){\oval(3,3)}
   \put(3,5.8){\makebox(0,0){$\grishinnf | \grishinnc$}}
   \put(1,13.2){\oval(3,3)}
   \put(1,13.2){\makebox(0,0){$\grishinna.\grishinnf | \grishinnh.\grishinnc$}}
   \put(1,15.2){\makebox(0,0){\nnode{}{}}}

   \put(11,7.8){\mparbottomc{}{\nnode{H}{}}{\nnode{}{}}{\nnode{}{}}}
   \put(11,0.4){\mtensorc{}{\nnode{}{C}}{\nnode{}{}}{\nnode{}{}}}
   \put(11,5.8){\oval(3,3)}
   \put(11,5.8){\makebox(0,0){$\grishinna | \grishinnh$}}
   \put(15,5.8){\oval(3,3)}
   \put(15,5.8){\makebox(0,0){$\grishinne | \grishinnd$}}

   \put(21,0.4){\mparleftcb{}{\nnode{}{C}}{\nnode{}{}}}
   \put(21,7.8641016){\mtensorc{}{\nnode{}{}}{\nnode{H}{}}{\nnode{}{}}}
%   \spline(23.3,1.3)(26.5,-0.4)(27,3.8)(26.5,8.0)(25.3,7.5) orig
    \spline(23.3,1.3)(26.5,-0.4)(27,3.8)(26.5,16.0)(25.3,15.4)
%   \spline(25.3,-3.9)(26.5,-4.4)(27,3.8)(26.5,12.0)(23.3,10.3) inv
   \put(23,5.8){\oval(3,3)}
   \put(23,5.8){\makebox(0,0){$\grishinnb | \grishinng$}}
   \put(24.8,13.2){\oval(3,3)}
   \put(24.8,13.2){\makebox(0,0){$\grishinne.\grishinnb |  \grishinnd.\grishinng$}}
   \put(24.8,15.2){\makebox(0,0){\nnode{}{}}}
\end{picture}}
\end{center}
\caption{Contractions: binary residuated with Grishin}
\label{fig:contr:bingrish}
\end{figure}

When we spell out all different possibilities for all different par
links, we end up with the schematic contractions shown in
Figures~\ref{fig:contr:bindualgrish} and \ref{fig:contr:bingrish}
for the binary residuated and dual residuated connectives. $A | B$
indicating that either Grishin rule $A$ or Grishin rule $B$ applies,
depending on the structure the tensor link finds itself in, and with
$A.B$ indicating the we apply Grishin rule $A$ followed by Grishin
rule $B$. We remark that this sequencing operation means moving links
first up then towards the $C$ formula in the structure.

%\begin{figure}
%\begin{center}
%\scalebox{0.8}{%
%\begin{picture}(26,10)(0,-1)
%\thicklines
%   \put(1,0.4){\mparrightcb{}{\nnode{}{C}}{}{}}
%   \put(1,3.8641016){\mtensorc{}{\nnode{}{}}{\nnode{}{}}{\nnode{H}{}}}
%   \spline(2.7,1.3)(-0.5,-0.4)(-1,3.8)(-0.5,8.0)(0.7,7.5)
%   \spline(0.7,0.1)(-0.5,-0.4)(-1,3.8)(-0.5,8.0)(2.7,6.3)
%   \spline(2.7,2.1)(1,0.4)(0,3.8)(1,7.2)(2.7,5.5)
%   \put(4.5,8){\oval(4,2.5)}
%   \put(4.5,-0.5){\oval(4,2.5)}

%   \put(11,0.4){\mtensorc{}{\nnode{}{C}}{}{}}
%   \put(11,3.8){\mparbottomc{}{\nnode{H}{}}{\nnode{}{}}{\nnode{}{}}}
%   \put(13,8){\oval(4,2.5)}
%   \put(13,-0.5){\oval(4,2.5)}

%   \put(21,0.4){\mparleftcb{}{\nnode{}{C}}{}}
%   \put(21,3.8641016){\mtensorc{}{\nnode{}{}}{\nnode{H}{}}{\nnode{}{}}}
%   \spline(23.3,1.3)(26.5,-0.4)(27,3.8)(26.5,8.0)(25.3,7.5)
%   \spline(25.3,0.1)(26.5,-0.4)(27,3.8)(26.5,8.0)(23.3,6.3)
%   \spline(23.3,2.1)(25,0.4)(26,3.8)(25,7.2)(23.3,5.5)
%   \put(21,3.8641016){\parleftcb{}{\nnode{H}{}}{}}
%   \put(21,0.4){\tensorc{}{\nnode{}{}}{\nnode{}{C}}{\nnode{}{}}}
%   \spline(25.3,0.1)(26.5,-0.4)(27,3.8)(26.5,8.0)(23.3,6.3)
%   \put(21.5,8){\oval(4,2.5)}
%   \put(21.5,-0.5){\oval(4,2.5)}
%   \put(22,14){\mparbottomc{}{A\otimes B}{A}{B}}
%   \put(16,14){\mparleftc{}{A\blolli B}{A}{B}}
%   \put(28,14){\mparrightc{}{A\lolli B}{A}{B}}
%\end{picture}}
%\end{center}
%\caption{Contractions: binary residuated}
%\label{fig:contr:binres}
%\end{figure}

A second important point to note is that these operations can be
\emph{nondeterministic}. For example if both subnets of a generalized
contraction contain links, we can use either possibility to move
toward a contraction redex.

However, the situation changes when we separate the Grishin I and Grishin IV interactions. In the Grishin I situation, only the
substructures containing just primed rules will remain. While this
removes the non-determinism for the (co-)impilications --- only the
`standard' contractions are valid in this case --- the product
formulas will still potentially generate multiple solutions. In the
Grishin IV situation, however, \emph{all} non-determinism disappears.
%\marginpar{remark here: Grishin IV = deterministic, Grishin I =
%  nondeterministic}

The generalized contractions suggest the following algorithm for
determining contractability in the Lambek-Grishin calculus: we use two
disjoint set data structures, one for residuated connected components
of tensor links
and one for dual residuated connected components of tensor links. 

Now to determine
contractability of the binary dual residuated connectives, it suffices
to know that both hypotheses and conclusions of the two substructures in the
figure are connected by a path of residuated tensor links. If they are,
we perform a set union operation on the hypothesis and conclusion
vertices in both disjoint set data structures.  For the residuated
connectives, we simply verify connectedness by a path of dual
residuated tensor links.

In the absence of either the Grishin class I or the Grishin class IV
structural rules, some of the substructures of the figure will be required
to be empty, but this will not influence the complexity.

The total cost of deciding whether an abstract proof structure with
$v$ vertices, $t$ tensor links and $p$ par links is
contractible is therefore summarized as follows.

\sk{.5}

\textbf{Initialisation:} $v$ \textsc{Make-Set} operations and $t$
\textsc{Union}s.

\textbf{Contraction:} at most $2p$ \textsc{Find-Set} operations followed by two
\textsc{Union}s.

\sk{.5}

We may still have to check all $p$ par links to find one
which is contractible, giving a total of $p^2$ `contraction
attemps'. This gives us an upper bound up the total complexity of $\Theta(p^2 \log v)$ \cite[p.\ 449]{algo}\footnote{This
  is the complexity when using just the path compression heuristic,
  the actual complexity is $O(m \alpha(m,n))$ where $\alpha$ is the
  `inverse' Ackerman function}, only a small increase over the naive
solution without structural rules.

\subsection{Binary and Unary Without Structural Conversions}

For a binary contraction a tensor and a par link have to be connected
at the two ports of the par links without the arrow. Given that there
are only three ports to every binary link, this means it is impossible
for two par links to both be candidates for reduction with the same
tensor link.

With the unary contractions, this situation changes. Since only one
port of the tensor and the par link have to be connected, a tensor link can
be a candidate for reduction with two par links.

\section{Applications}

I will now turn to some applications of the Lambek-Grishin calculus
LG. First by showing that then languages generated by LG grammars are
outside the context free languages.

\subsection{LG and Tree Adjoining Grammars}

\citeasnoun{kand88nl} shows that the non-associative Lambek calculus
NL generates only context free languages. Several additions to NL have
been proposed to increase the expressive power of the calculus. The
solution advocated by \citeasnoun{M95} is a combination of modes,
structural rules and control operators, whereas \citeasnoun{multiccg}
propose modes and combinators.

More recent research \cite{moortgat07sym,bernardi07conti} has looked
at syntactic and semantic applications of the Lambek-Grishin calculus
LG, which extends NL by adding dual residuated operators and
interactions between the residuated and dual residuated operators,
as shown in Figure~\ref{fig:grishin} and \ref{fig:grishinmc}.

It is the goal of this section to show that even LG with just the
Grishin IV interactions can generate languages which are not context
free. We will do this by giving an embedding translation of
lexicalized tree adjoining grammars (\ltag s).

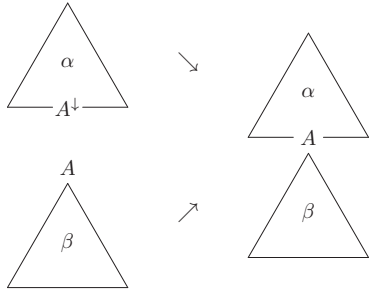
\begin{figure}
\begin{center}
\scalebox{0.8}{%
\begin{picture}(12,10)

\put(2,6){\makebox(0,0){$A^{\downarrow}$}}
\put(2,7.5){\makebox(0,0){$\alpha$}}
\drawline(1.5,6)(0,6)(2,9.4641016)(4,6)(2.5,6)

\drawline(0,0)(2,3.4641016)(4,0)(0,0)
\put(2,4){\makebox(0,0){$A$}}
\put(2,1.5){\makebox(0,0){$\beta$}}

\drawline(8,1)(10,4.4641016)(12,1)(8,1)
\put(10,5){\makebox(0,0){$A$}}
\put(10,2.5){\makebox(0,0){$\beta$}}
\drawline(9.5,5)(8,5)(10,8.4641016)(12,5)(10.5,5)
\put(10,6.5){\makebox(0,0){$\alpha$}}

%\put(6,5.5){\makebox(0,0){$\leadsto$}}
\put(6,7.5){\makebox(0,0){$\searrow$}}
\put(6,2.5){\makebox(0,0){$\nearrow$}}

\end{picture}}
\end{center}
\caption{The Substitution Operation}
\label{fig:subst}
\end{figure}

\begin{figure}
\begin{center}
\scalebox{0.8}{%
\begin{picture}(12,14)

\put(2,10){\makebox(0,0){$A^{*}$}}
\put(2,11.5){\makebox(0,0){$\alpha$}}
\drawline(1.5,10)(0,10)(2,13.4641016)(4,10)(2.5,10)
\put(2,14){\makebox(0,0){$A$}}

\drawline(0,0)(2,3.4641016)(4,0)(0,0)
\put(2,4){\makebox(0,0){$A$}}
\put(2,1.5){\makebox(0,0){$\beta''$}}
\drawline(1.5,4)(0,4)(2,7.4641016)(4,4)(2.5,4)
\put(2,5.5){\makebox(0,0){$\beta'$}}

\put(10,6.5){\makebox(0,0){$\alpha$}}
\drawline(9.5,5)(8,5)(10,8.4641016)(12,5)(10.5,5)
\put(10,9){\makebox(0,0){$A$}}

\drawline(8,1)(10,4.4641016)(12,1)(8,1)
\put(10,5){\makebox(0,0){$A$}}
\put(10,2.5){\makebox(0,0){$\beta''$}}

\drawline(9.5,9)(8,9)(10,12.4641016)(12,9)(10.5,9)
\put(10,10.5){\makebox(0,0){$\beta'$}}

%\put(6,6.5){\makebox(0,0){$\leadsto$}}
\put(6,4){\makebox(0,0){$\nearrow$}}
\put(6,11.5){\makebox(0,0){$\searrow$}}

\end{picture}}
\end{center}
\caption{The Adjunction Operation}
\label{fig:adj}
\end{figure}
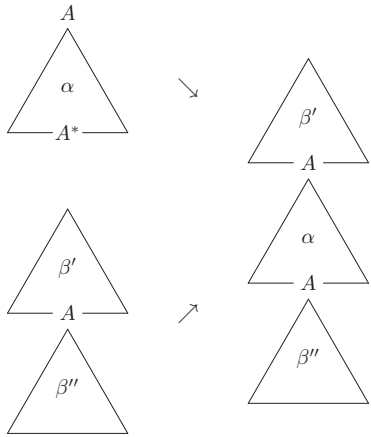

\ltag s are a widely used grammar formalism in computational
linguistics \cite{tags}. The basic objects are trees and there are two
operations on trees: \emph{substitution}, as shown in Figure~\ref{fig:subst}
replaces a leaf $A^{\downarrow}$ by a tree with root $A$, whereas
\emph{adjunction}, as shown in Figure~\ref{fig:adj}, replaces a node of
the original tree by a tree.

An embedding of \ltag s into multimodal categorial grammars has been
given in \cite[Chapter 10]{diss}. I will improve on this result
here. Most of the improvements are a consequence of using a subset of
\ltag s which I will call \ltagz. This allows me to have just a single
translation for adjunction points since all adjunction will take place
at formulas with a negative polarity. In addition the number of
structural rules is greatly reduced: instead of 12 different
structural rules, the rules Grishin 1 and Grishin 2 will suffice.

I will first define the \ltagz\ grammars, discuss some of the
differences with other definitions of \ltag s and then show that all
interesting language classes of \ltag s can be treated in \ltagz\ as
well.

\begin{definition} \label{def:ltag}
An \ltagz\ grammar is a tuple
$\langle T, N_S, N_A, I, A \rangle$ such that

\begin{itemize}
\item  $T$, $N_S$ and $N_A$ and three disjoint alphabets of terminals,
  substitution nonterminals and adjunction nonterminals respectively, we will use upper case letters $A , B, \ldots$ and of course the distinguised
start symbol $S$ to stand for members of $N_S$ whereas we will use upper
case letters $T, U, \ldots$ for members of $N_A$.

\item $I$ is a finite set of \emph{initial trees},
\item $A$ is a finite set of \emph{auxiliary trees}.
\end{itemize}

The trees in $I \cup A$ are called the \emph{elementary trees}.

Trees are subject to the following conditions:
\begin{itemize}
\item the root nodes of all initial trees are members of $N_S$,
\item the root nodes of all auxiliary trees are members of $N_A$,
\item every auxiliary tree has exactly one leaf which is a member of
  $N_A$ which we will call the \emph{foot node},
\item every elementary tree has exactly one leaf which is a member of
  $T$.
\item every adjunction node, which we will mark as $(T)$ in the tree,
  is on the path from the lexical leaf to the root of the tree.
\end{itemize}
\end{definition}

This definition differs on several points from the standard
definition of \ltag s as in for example \cite{joshi}. I will comment on
each of these points.

Firstly, the difference between substitution and adjunction
nonterminals is minor and is already implicit in the notation of
$A^{\downarrow}$ for substitution nodes and $A^{*}$ for foot nodes,
which like our choice of two different alphabets serves to remove any
possible confusion about whether a substitution or adjunction
operation should be applied to a node.

Some authors choose to mark null adjunction nodes explicitly. Given
that the translation of the adjunction nodes is slightly more
intricate than that of null-adjunction nodes, which we can just
ignore, I have chosen opposition strategy of marking the (non-null)
adjunction nodes explicitly.

Lexicalization is a fairly common restriction on \ltag s, our only
additional restriction to it is that we require a \emph{unique}
terminal leaf.

The final and most important restriction is the requirement that every
adjunction takes place on the path from the lexical leaf to the root
of the tree. This is a real restriction, but one that simplifies our
embedding result. 

The definition of \ltagz\ is close to the definition of \emph{normal}
\ltag s used by \citeasnoun{conv} to show correspondence between \ltag
s
and combinatory categorial grammars, with the following differences:
first, we don't need the requirement that all internal nodes have
either the obligatory adjunction or the null adjunction constraint,
second, the adjunction nodes are required to be on the path from the
root to the lexical leaf instead of the foot node.

\begin{definition} Given an \ltagz\ grammar $g$ a \emph{derivation
    tree} $d$ is a binary branching tree such that:

\begin{itemize}
\item every leaf of $d$ is an elementary tree of $g$,
\item every branch combines its two daughter trees using either the
  adjunction or the substitution operation,
\item the root node is a tree which has the distinguished symbol $S$
  as its root and only terminals as its leaves.
\end{itemize}
\end{definition}

We will show that there are \ltagz\ grammars that
can handle all the interesting phenomena that \ltag\ grammars can.

\begin{lemma}\label{lem:ltaggen} The are \ltagz\ grammars generating

\begin{itemize}
\item the copy language $\{
  ww \; | \; w \in \{a,b\}^+ \}$,
\item counting dependencies $\{ a^n b^n c^n \},
  n > 0$ and
\item crossed dependencies $\{ a^n b^m c^n d^m \},
  n > 0$
\end{itemize}
\end{lemma}

\paragraph{Proof} Figure~\ref{fig:copy}, Figure~\ref{fig:counting} and
Figure~\ref{fig:crossed} shows the \ltagz grammars
generating the copy language, counting depencies and crossed
dependencies respectively. Showing we generate all and only the
strings in the different languages is easy once we keep the following
invariants in mind.

\begin{itemize}
\item For the copy language, the $(T)$ adjunction point always has the
  entire copy of $w$ as its descendants. Every adjunction adds either
  an \textit{a} or a \textit{b} to the end of the first string
  as well as to the end of the copy while creating a new
  adjunction point covering the new copy.
\item For the counting dependencies, the $(T)$ adjunction point always
  has the $a^n b^n$ part of the string as its descendants. Every
  adjunction will add an \textit{a} and a \textit{b} to both sides of
  the $a^n b^n$ part, while creating a new adjunction point containing
  the bigger $a^n b^n$ sequence. In addition, a \textit{c} is added
  after the final \textit{b}. 
\item For the crossed dependencies, first $(T)$ will have
  all \textit{c}s as its descendants, then $(U)$ will have all
  \textit{c}s and \textit{d}s as its descendants. All $(T)$ adjunctions generate
  both an \textit{a} and a \textit{c} keeping all \textit{c}s under
  the $(T)$. When we generate the final \textit{a} and \textit{c}
  terminals, we start adjoining \textit{b}s and \textit{d}s at the
  $(U)$ adjunction point, the \textit{b}s appearing before the new
  ajunction point and the \textit{d}s at the end inside it.  \qed
\end{itemize}

{\setlength{\unitlength}{3mm}
\begin{figure}
\begin{center}
\begin{picture}(20,17)(0,3)
\put(1,12){\lubranch{$A$}{\terminal{a}}}
\put(1,3){\lubranch{$B$}{\terminal{b}}}

\put(6,15){\lbbranch{$S$}{$A$}{}}
\put(9,12){\lubranch{$(T)$}{\terminal{a}}}

\put(15,6){\lbbranch{$T$}{$B$}{}}
\put(17,3){\lbbranch{$(T)$}{$T$}{\terminal{b}}}

\put(6,6){\lbbranch{$S$}{$B$}{}}
\put(9,3){\lubranch{$(T)$}{\terminal{b}}}

\put(15,15){\lbbranch{$T$}{$A$}{}}
\put(17,12){\lbbranch{$(T)$}{$T$}{\terminal{a}}}

\end{picture}
\end{center}

\caption{the copy language $\{ ww \; | \; w \in \{a,b\}^+ \}$}
\label{fig:copy}
\end{figure}
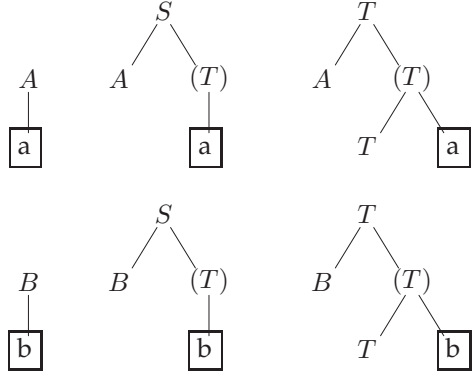
}

{\setlength{\unitlength}{3mm}
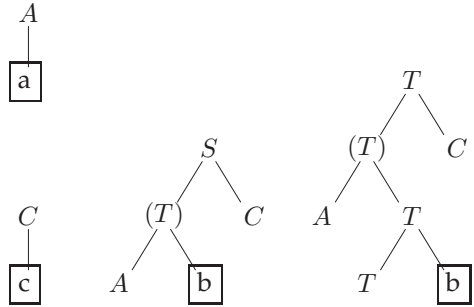
\begin{figure}
\begin{center}
\begin{picture}(20,13.5)(0,3)
\put(1,12){\lubranch{$A$}{\terminal{a}}}
\put(1,3){\lubranch{$C$}{\terminal{c}}}

\put(15,6){\lbbranch{$(T)$}{$A$}{}}
\put(17,3){\lbbranch{$T$}{$T$}{\terminal{b}}}
\put(17,9){\lbbranch{$T$}{}{$C$}{}}

\put(8,6){\lbbranch{$S$}{}{$C$}}
\put(6,3){\lbbranch{$(T)$}{$A$}{\terminal{b}}}

\end{picture}
\end{center}

\caption{counting dependencies $\{ a^n b^n c^n \},
  n > 0$}
\label{fig:counting}
\end{figure}
}

{\setlength{\unitlength}{3mm}
\begin{figure}
\begin{center}
\begin{picture}(30,17)(0,3)
\put(1,12){\lubranch{$A$}{\terminal{a}}}
\put(1,3){\lubranch{$B$}{\terminal{b}}}

\put(6,15){\lbbranch{$S$}{$A$}{}}
\put(9,12){\lubranch{$(T)$}{\terminal{c}}}

\put(15,15){\lbbranch{$T$}{$A$}{}}
\put(17,12){\lbbranch{$(T)$}{$T$}{\terminal{c}}}

\put(6,6){\lbbranch{$S$}{$A$}{}}
\put(9,3){\lubranch{$(U)$}{\terminal{c}}}

\put(15,6){\lbbranch{$T$}{$A$}{}}
\put(17,3){\lbbranch{$(U)$}{$T$}{\terminal{c}}}

\put(24,6){\lbbranch{$U$}{$B$}{}}
\put(26,3){\lbbranch{$(U)$}{$U$}{\terminal{d}}}

\end{picture}
\end{center}

\caption{crossed dependencies $\{ a^n b^m c^n d^m \},
  n > 0$}
\label{fig:crossed}
\end{figure}
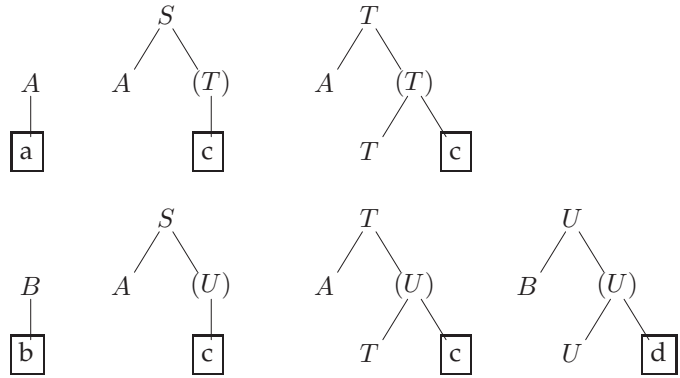
}

%\begin{corollary} \ltagz\ is a mildly context sensitive grammar formalism.
%\end{corollary}

%This is an immediate consequence the polynomial parsing for
%\ltagz\ (which can be done using any LTAG parsing algorithm) and
%Lemma~\ref{lem:ltaggen}.

\begin{definition} Let $g$ be an \ltagz\ grammar, we will define the
  corresponding LG grammar $g'$ as follows. The atomic formulas $A$ of
  $g'$ will be $N_A \cup N_S$ to which we add a new formula $i$. We
  will translate all members of $N_A \cup N_S$ by the corresponding
  member of $A$, except for the foot nodes $T$, which will be translated as
  $T' := (T \blolli i) \otimes i$.

  We proceed by recursive descent of the \ltagz\ tree from the root
  $R$ downwards towards the unique terminal leaf $t$.

\begin{itemize}
\item We start at the root $R$ and use $R$ as the
  current formula $f$.

\item We translate an adjunction point $(T)$ by assigning $f := (F
  \looparrowleft f) \looparrowright ((T \blolli i) \otimes i)$.
  
\item We translate a binary branch by assigning $f := A \lolli f$ if the terminal leaf is
  a descendant of the right node or $f := f \blolli A$ if it is a
  descendant of the left node. $A$ is a pure product formula
  representing the structure of the descendants of the other node;
  typically $A$ is just an atomic formula.

\item  When we arrive at the terminal leaf $t$, we add
  $\textit{lex}(t) = R$ to the lexicon.
\end{itemize}
\end{definition}

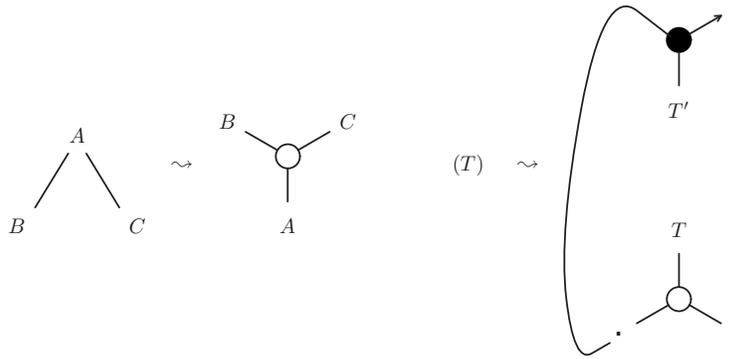
\begin{figure}
\begin{center}
\scalebox{0.8}{%
\begin{picture}(26,16)(0,-1)
\thicklines
   \put(1,4){\lbbranch{$A$}{$B$}{$C$}}
   \put(6.5,6){\makebox(0,0){$\leadsto$}}
   \put(8,4){\mtensorc{}{A}{B}{C}}
   \put(16,6){\makebox(0,0){$(T)$}}
   \put(18,6){\makebox(0,0){$\leadsto$}}
   \put(21,7.8641015){\parrightcb{}{\nnode{}{}}{T'}{}}
   \put(21,0.4){\tensorc{}{T}{\nnode{}{}}{\nnode{}{}}}
   \spline(20.7,0.1)(19.5,-0.6)(19,3.8)(20.5,12.0)(22.7,10.3)
\end{picture}}
\end{center}
\caption{Translating \ltagz\ to LG}
\label{fig:ltagtolg}
\end{figure}

This translation is perhaps easiest to visualize in the form of trees
as shown in Figure~\ref{fig:ltagtolg}. From this figure it should be
clear that the adjunction operation is going to correspond to two $T$
axiom connections followed by the Grishin 1 and 2 rules and a
contraction, as shown schematically in Figure~\ref{fig:contr:bindualgrish}.

 Some properties to remark about
this translation: first of all, internal nodes which are not
adjunction points, that is have the null adjunction constraint,
disappear. Given that they are just ornamental in the \ltag\ grammar,
this is just a cosmetic change. 

Secondly, given that the following sequent is derivable in
LG even without interaction principles

$$ (((T \blolli i) \otimes i) \looparrowleft A) \looparrowright T \vdash A$$

\noindent we can handle the possibility that no adjunction takes place
at an adjunction node very naturally.

Another point, made clear by Figure~\ref{fig:ltagtolg} is that our
\ltagz\ trees are upside-down! This is easy to remedy using the
symmetries of LG, but I feel the current solution using the residuated
instead of the dual residuated connectives as the `main' connectives
is to be preferred.

A final point concerns the use of $(A \blolli i) \otimes i$
formulas. This is necessary to prevent multiple auxiliary trees to be
adjoined at the same time, connecting the root and foot nodes together
to form a sort of derived auxiliary tree. In all of the given example
grammars this would lead to overgeneration: we explicitly want to
assign the root and foot nodes a null adjunction constraint. In the LG
grammar we exploit the derivability relation for this. Because  $A
\nvdash (A \blolli i) \otimes i$ we cannot perform the
$[\textit{R}\blolli]$ contraction when we connect the root and foot
node of two auxiliary trees. There are other possible solutions: one is
to use different formulas $T$ and $U$ for the foot and root nodes as
well as for the corresponding atomic formulas in the adjunction
point. This would correspond to an obligatory adjunction constraint
similary to the obligatory adjunction solution proposed by
\citeasnoun{conv}, but it would result in added lexical ambiguity to
allow for cases where there is no adjunction at an adjunction
point. Another solution would be to use the unary modes to implement
the same restriction and replace $(A \blolli i) \otimes i$ by
$\Diamond \Box A$.

As an example, the following LG lexicons correspond to the three
\ltagz\ grammars we presented before.

The copy grammar is shown below.

\sk{.5}
\begin{tabular}{ll}
\textit{lex}(a) & $A$ \\
\textit{lex}(a) & $(T \looparrowleft (A \lolli S)) \looparrowright ((T \blolli i) \otimes i)  $ \\
\textit{lex}(a) & $((T\blolli i)\otimes i) \lolli (T \looparrowleft
(A \lolli T)) \looparrowright ((T \blolli i) \otimes i))  $ \\
\textit{lex}(b) & $B$ \\
\textit{lex}(b) & $(T \looparrowleft (B \lolli S)) \looparrowright ((T \blolli i) \otimes i)  $ \\
\textit{lex}(b) & $((T\blolli i)\otimes i) \lolli (T \looparrowleft
(B \lolli T)) \looparrowright ((T \blolli i) \otimes i))  $ \\
\end{tabular}
\sk{.5}

In the following, we will choose to abbreviate the formulas $(T
\blolli i) \otimes i$ by $T'$ to improve the readability. The
abbreviated version of the grammar above looks as follows.

\sk{.5}
\begin{tabular}{ll}
\textit{lex}(a) & $A$ \\
\textit{lex}(a) & $(T \looparrowleft (A \lolli S)) \looparrowright T'  $ \\
\textit{lex}(a) & $T' \lolli (T \looparrowleft
(A\lolli T)) \looparrowright T')  $ \\
\textit{lex}(b) & $B$ \\
\textit{lex}(b) & $B (T \looparrowleft (B \lolli S)) \looparrowright T' $ \\
\textit{lex}(b) & $T' \lolli (T \looparrowleft
(B\lolli T)) \looparrowright T')  $ \\
\end{tabular}
\sk{.5}

The grammar for counting dependencies looks as follows.

\sk{.5}
\begin{tabular}{ll}
\textit{lex}(a) & $A$ \\
\textit{lex}(b) & $A \lolli ((T \looparrowleft (S\blolli C)) \looparrowright T'))  $ \\
\textit{lex}(b) & $T' \lolli (A \lolli (T \looparrowleft
(T\blolli C)) \looparrowright T'  $ \\
\textit{lex}(c) & $C$ \\
\end{tabular}
\sk{.5}

Finally, the grammar for crossed dependencies is shown below.

\sk{.5}
\begin{tabular}{ll}
\textit{lex}(a) & $A$ \\
\textit{lex}(b) & $B$ \\
\textit{lex}(c) & $(T \looparrowleft (A\lolli S)) \looparrowright T'  $ \\
\textit{lex}(c) & $T' \lolli (T \looparrowleft
(A\lolli T)) \looparrowright T')  $ \\
\textit{lex}(c) & $(U \looparrowleft (A \lolli S)) \looparrowright U'  $ \\
\textit{lex}(c) & $T' \lolli (U \looparrowleft
(A\lolli T)) \looparrowright U')  $ \\
\textit{lex}(d) & $U' \lolli (U \looparrowleft
(B \lolli U)) \looparrowright U')  $ \\
\end{tabular}
\sk{.5}

\begin{lemma} \label{lem:rootfoot} Let $g'$ be the translation of an \ltagz\ grammar $g$ into
  LG and let $\mathcal S$ be the proof structure corresponding to an auxiliary
  tree of $g$. Any proof net $\mathcal P$ of $g'$ which has $\mathcal S$ as a
  substructure has the root and foot node of $\mathcal S$ connected to
  the two atomic formulas of one adjunction point of $\mathcal P$.
\end{lemma}

\paragraph{Proof} Look at the root node $r$ and the foot node $f$ of a
proof structure $\mathcal S$. Given that $r$ and $f$ are both members
of $N_A$, there are few possibilities for identifying $r$ with other
atomic formulas. If we identify $r$ with a foot node, we will end up
with a non-contractible tree, given that $A \nvdash (A \blolli i)
\otimes i$. Connecting it to a substitution leaf is impossible, given
that $N_S \cap N_A = \emptyset$. So the only possibility is to perform
an axiom link with an adjunction node. 

Now look at the corresponding foot node. If we identify it with the
root of another auxiliary tree, we get a non-contractible tree
again. The root of an initial tree is excluded, given that we have an
atom from a different alphabet. Attaching it to a \emph{different}
adjunction node is excluded as well, since we won't be able to perform
either $[\textit{L}\looparrowright]$ contraction. So the only
remaining option is to attach it to the other atomic formula of the
same adjunction point as the root node. \qed

\begin{lemma} For every \ltagz\ grammar $g$ which generates langague
  $L$, the LG grammar $g'$ generates the same language.
\end{lemma}

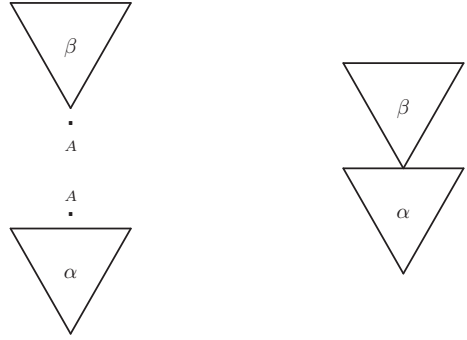
\begin{figure}
\begin{center}
\begin{tabular}{cc}
\scalebox{0.8}{%
\begin{picture}(10,12)(0,1)
\thicklines
   \put(5,4.5){\makebox(0,0){$\nnode{A}{}$}}
   \put(5,7.5){\makebox(0,0){$\nnode{}{A}$}}
   \drawline(3,11.5)(7,11.5)(5,8)(3,11.5)
   \drawline(3,4)(7,4)(5,0.5)(3,4)
   \put(5,2.5){\makebox(0,0){$\alpha$}}
   \put(5,10){\makebox(0,0){$\beta$}}
\end{picture}}
&
\scalebox{0.8}{%
\begin{picture}(10,12.5)(0,1)
\thicklines
   \drawline(3,9.5)(7,9.5)(5,6)(3,9.5)
   \drawline(3,6)(7,6)(5,2.5)(3,6)
   \put(5,4.5){\makebox(0,0){$\alpha$}}
   \put(5,8){\makebox(0,0){$\beta$}}
\end{picture}}
\end{tabular}
\end{center}
\caption{The substitution operation on abstract proof structures}
\label{fig:substaps}
\end{figure}

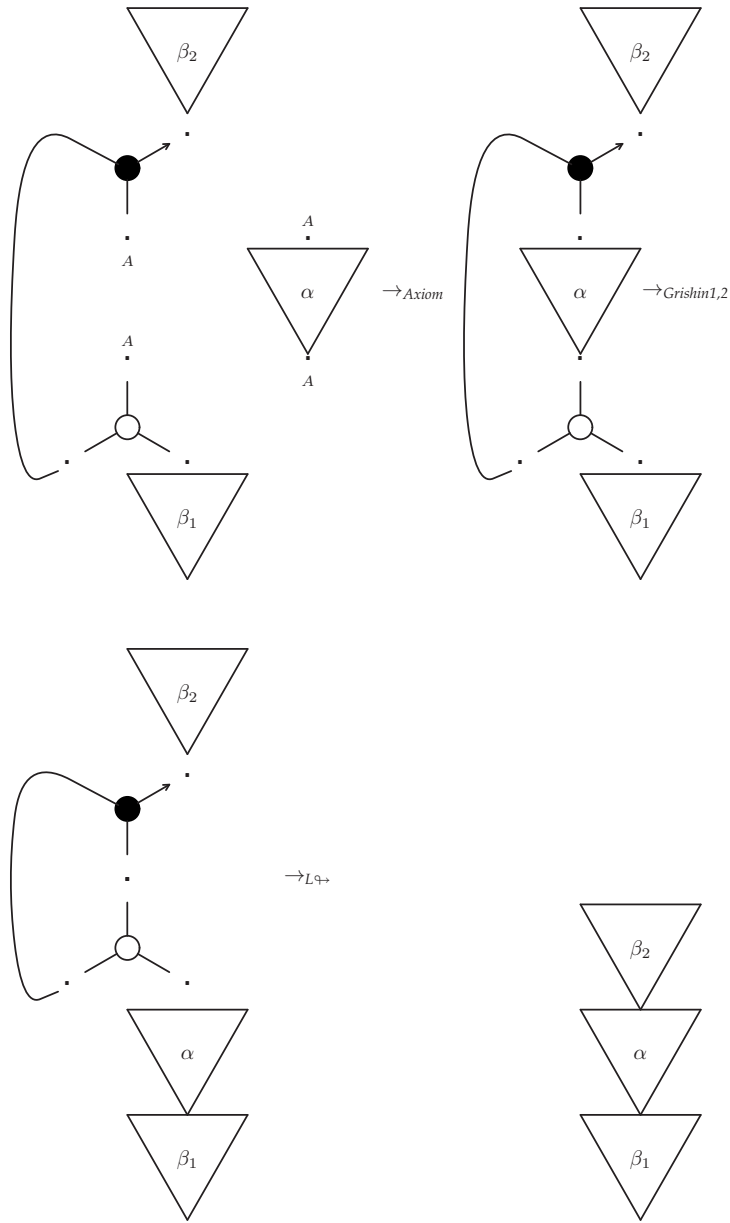
\begin{figure}
\begin{center}
\begin{tabular}{cc}
\scalebox{0.8}{%
\begin{picture}(14,18)(0,-2)
\thicklines
   \put(1,7.8641015){\parrightcb{}{\nnode{}{}}{\nnode{}{A}}{}}
   \put(1,0.4){\tensorc{}{\nnode{A}{}}{\nnode{}{}}{\nnode{}{}}}
   \spline(0.7,0.1)(-0.5,-0.4)(-1,3.8)(-0.5,12.0)(2.7,10.3)

   \put(9,7.8641015){\makebox(0,0){$\nnode{A}{}$}}
   \put(9,3.8641016){\makebox(0,0){$\nnode{}{A}$}}
   \drawline(7,7.5)(11,7.5)(9,4)(7,7.5)
   \drawline(3,15.5)(7,15.5)(5,12)(3,15.5)
   \drawline(3,0)(7,0)(5,-3.5)(3,0)
   \put(5,-1.5){\makebox(0,0){$\beta_1$}}
   \put(5,14){\makebox(0,0){$\beta_2$}}
   \put(9,6){\makebox(0,0){$\alpha$}}
   \put(12.5,6){\makebox(0,0){$\rightarrow_{\textit{Axiom}}$}}   
\end{picture}} &
\scalebox{0.8}{%
\begin{picture}(10,18)(0,-2)
\thicklines
   \put(1,7.8641015){\parrightcb{}{\nnode{}{}}{\nnode{}{}}{}}
   \put(1,0.4){\tensorc{}{\nnode{}{}}{\nnode{}{}}{\nnode{}{}}}
   \spline(0.7,0.1)(-0.5,-0.4)(-1,3.8)(-0.5,12.0)(2.7,10.3)
   \drawline(1,7.5)(5,7.5)(3,4)(1,7.5)
   \drawline(3,15.5)(7,15.5)(5,12)(3,15.5)
   \drawline(3,0)(7,0)(5,-3.5)(3,0)
   \put(5,-1.5){\makebox(0,0){$\beta_1$}}
   \put(5,14){\makebox(0,0){$\beta_2$}}
   \put(3,6){\makebox(0,0){$\alpha$}}
   \put(6.5,6){\makebox(0,0){$\rightarrow_{\textit{Grishin1,2}}$}}   
\end{picture}} \\
\scalebox{0.8}{%
\begin{picture}(14,21)(0,-2)
\thicklines
   \put(1,7.8641015){\parrightcb{}{\nnode{}{}}{\nnode{}{}}{}}
   \put(1,4.4){\tensorc{}{\nnode{}{}}{\nnode{}{}}{\nnode{}{}}}
   \put(9,7.8641015){\makebox(0,0){$\rightarrow_{\textit{L}\looparrowright}$}}   
   \spline(0.7,4.1)(-0.5,3.6)(-1,7.8)(-0.5,12.0)(2.7,10.3)
   \drawline(3,3.5)(7,3.5)(5,0)(3,3.5)
   \drawline(3,15.5)(7,15.5)(5,12)(3,15.5)
   \drawline(3,0)(7,0)(5,-3.5)(3,0)
   \put(5,-1.5){\makebox(0,0){$\beta_1$}}
   \put(5,14){\makebox(0,0){$\beta_2$}}
   \put(5,2){\makebox(0,0){$\alpha$}}
\end{picture}} &
\scalebox{0.8}{%
\begin{picture}(10,21)(0,-2)
\thicklines
   \drawline(3,3.5)(7,3.5)(5,0)(3,3.5)
   \drawline(3,7)(7,7)(5,3.5)(3,7)
   \drawline(3,0)(7,0)(5,-3.5)(3,0)
   \put(5,-1.5){\makebox(0,0){$\beta_1$}}
   \put(5,5.5){\makebox(0,0){$\beta_2$}}
   \put(5,2){\makebox(0,0){$\alpha$}}
\end{picture}}
\end{tabular}
\end{center}
\caption{The adjunction operation on abstract proof structures}
\label{fig:adjaps}
\end{figure}

\paragraph{Proof} (Sketch) Let $d$ be an \ltagz\ derivation tree ending in tree
$t$ using a grammar $g$. Let $g'$ the corresponding LG grammar. We
show that $g'$ derives the same tree $t$. 

For every substitution we
simply identify the two corresponding nodes in the LG proof
structures. This will correspond to an axiom in the resulting proof
net and produces a tree isomorphic to the result of applying the
substitution operation in $g$. Figure~\ref{fig:substaps} shows how the
(abstract) proof structures can be combined.

For every adjunction we identify the root and foot nodes with the two
nodes of the adjunction point. We can apply the $[\textit{L}\otimes]$
and the $[\textit{R}\lolli]$ contractions straigt away. After all
axiom connections are performed it is time for the generalized
$[\textit{L}\looparrowright]$ contractions, working from the inside
out. Every generalized contraction will produce a subtree which is
isomorphic to the result of applying the adjunction operation in
$g$. Figure~\ref{fig:adjaps} shows the different operations on the
abstract proof structure.

For the other direction, we show that if $g'$ produces a derivation
ending in tensor tree $t$, then $g$ produces this same tree $t$ as
well.

Lemma~\ref{lem:rootfoot} shows that axiom connections corresponding to
adjuctions come in pairs and we need to apply rules Grishin 1
and 2 in order to contract the $[\textit{L}\looparrowright]$ link in
the case of adjunction at a node. Having no adjunctions at an
adjunction point corresponds to connecting and contracting the $T$ and
$T'$ substructures. All other axiom connections correspond to
substitutions in the grammar \qed

By giving an embedding translation of \ltag s, we've shown that LG
with the Grishin class IV interactions handle more complicated
phenomena than those we can treat using context free
grammars. \textit{How much} more is still
unclear. 

\possessivecite{moortgat07sym} treatment of generalized quantifiers
appears to move us beyond simple \ltag\ grammars. Generalized
quantifiers are generally handles using multi-component tree adjoining
grammars. Generalizing the above translation to MCTAGs seems an
interesting possibility, whereas it would also be a candidate for
giving an \emph{upper bound} on the descriptive complexity of LG.

The type $(s \looparrowleft s) \looparrowright np$ assigned
to generalized quantifiers looks similar to the $(t \looparrowleft a)
\looparrowright t$ translation of insertion points. Both are instances
of the $q(A,B,C)$ operator which would make NL+q another candidate for
the desciptive complexity of LG.

\subsection{Scrambling}

\editout{
\section{Definability}

Defining connectives by using combinations of other
connectives. Eg. 

\begin{tabular}{cc}
$A^{\perp}\equiv A\lolli\perp$ & $\rule{0pt}{1ex}^{\perp}\! A \equiv
\perp\blolli A$ \\
$A^{\one}\equiv A\looparrowright \one$ & $\rule{0pt}{1ex}^{\one}\! A \equiv
\one\looparrowleft A$ \\
$\Diamond A \equiv A\otimes \one$ & $\Box A \equiv A\blolli \one$ \\

\end{tabular}

NB these are \emph{behavioral} equivalences, there is no
interderivability between any of the proposed formulas in a
translation.

\section{Fragments}

TLG (unary/binary residuals)

TLG+Galois (unary/binary residuals+unary Galois)

Lambek Galois (Galois only, implicit associative product)

Bilinear Lambek Calculus (binary res+dual)

Display Logic (all except binary gc/dgc)

Abrusci/Ruet (binary res+dual, negation defined or tensor-only)}

\section{Conclusions}

I've shown how to extend the proof net calculus for $\nldr$ to
display logic, adding several families of connectives while dropping
only the contraint that a tensor link has a unique conclusion. I've
shown basic soundness and completeness results and discussed the
complexity of the contraction criterion for several sublogics.

Finally, I've shown how to embed \ltag s into LG, giving a lower bound
on the descriptive complexity of LG and making the logic a candidate
for a mildly context sensitive grammar.

\appendix
\section{Translation Key}

\begin{table}
\begin{tabular}{|c|c|c|c|c|}
\hline Here       & LL       & DL   & TLG    & BLL \\ \hline

\multicolumn{5}{|c|}{Structural --- Nullary} \\ \hline
$\epsilon$ &          & $\Phi$      &        & $1$ \\ \hline
\multicolumn{5}{|c|}{Structural --- Unary} \\ \hline
$\langle.\rangle$ &   & $\circ$     & $\langle.\rangle$ & \\
$\lfloor.\rfloor$ &   & $\flat$     & $\flat$ & \\
$\lceil.\rceil$   &   & $\sharp$    & $\sharp$ & \\ \hline
\multicolumn{5}{|c|}{Structural --- Binary} \\ \hline
$\circ$    & $,$      & $;$         & $\circ$ & \\
$<$        &          & $<$         &         & \\
$>$        &          & $>$         &         & \\
$\lozenge$ &          &             &         & \\
$\triangleleft$ &          &             &         & \\
$\triangleright$ &          &             &         & \\ \hline
\end{tabular}
\caption{Translation Key --- Structural Connectives}
\label{tab:strtransl}
\end{table}

\begin{table}
\begin{tabular}{|c|c|c|c|c|}
\hline Here       & LL       & DL   & TLG    & BLL \\ \hline
\multicolumn{5}{|c|}{Logical --- Nullary} \\ \hline
$\one$     & $\one$   & $\one$      &        & $\one$ \\
$\perp$    & $\perp$  & $\zero$     &        & $\zero$ \\ \hline
\multicolumn{5}{|c|}{Logical --- Unary} \\ \hline
$\Diamond$ && $\Diamond$ & $\Diamond$ & \\
$\Box$ && $\Box$ & $\Box^{\downarrow}$ & \\
$.\rule{0pt}{1ex}^{\one}$ && $.\rule{0pt}{1ex}^{\one}$ & $.\rule{0pt}{1ex}^{\one}$ & $.\rule{0pt}{1ex}^{l}$ \\
$\rule{0pt}{1ex}^{\one}.$ && $\rule{0pt}{1ex}^{\one}.$ & $\rule{0pt}{1ex}^{\one}.$ & $.\rule{0pt}{1ex}^{r}$ \\
$.\rule{0pt}{1ex}^{\perp}$ & $.\rule{0pt}{1ex}^{\perp}$ & $.\rule{0pt}{1ex}^{\zero}$ & $.\rule{0pt}{1ex}^{\zero}$ & \\
$\rule{0pt}{1ex}^{\perp}.$ && $\rule{0pt}{1ex}^{\zero}.$ & $\rule{0pt}{1ex}^{\zero}.$ & \\ \hline
\multicolumn{5}{|c|}{Logical --- Binary} \\ \hline
$\otimes$ & $\otimes$ & $\otimes$ & $\bullet$ & $\otimes$ \\
$\lolli$ & $\lolli$ & $\rightarrow$ & $\backslash$ & $\backslash$ \\
$\blolli$ & & $\leftarrow$ & $/$ & $/$ \\
$\pr$ & $\pr$ & $\oplus$ & & $\oplus$ \\
$\looparrowright$ & & $>\!\!\!-$ & & \\
$\looparrowleft$ & & $-\!\!\!<$ & & \\
$\downarrow$ &&&& \\
$\swarrow$ &&&& \\
$\searrow$ &&&& \\
$\uparrow$ &&&& \\
$\nwarrow$ &&&& \\
$\nearrow$ &&&& \\ \hline
\end{tabular}
\caption{Translation Key --- Logical Connectives}
\label{tab:logtransl}
\end{table}

\bibliography{moot}

\end{document}